\documentclass[10pt, twocolumn, twoside]{article}
\usepackage[round]{natbib}
\usepackage{graphicx}

\usepackage{amssymb}
\usepackage{latexsym}
\usepackage[top=2.7cm,bottom=2.7cm, left=2.2cm, right=2.2cm]{geometry}
\usepackage{framed,multirow}
\usepackage{amsmath}
\usepackage{amssymb}
\usepackage{pdflscape} 
\usepackage{rotating}
\usepackage{pdfpages}
\usepackage{ragged2e}
\usepackage{booktabs}
\usepackage{cuted}
\usepackage{array}
\usepackage{caption}
\usepackage{enumitem}
\setlist[itemize]{leftmargin=*}
\setlist[itemize]{itemsep=2pt}

\usepackage{multirow}
\usepackage{url}

\usepackage{xspace}
\usepackage{placeins}
\usepackage{hyperref}
\usepackage{hyphenat}
\usepackage{adjustbox}

\AtBeginDocument{%
  \def\thefnote{\myfnsymbol{fnote}}}
\makeatletter
\def\myfnsymbol#1{\expandafter\@myfnsymbol\csname c@#1\endcsname}
\def\@myfnsymbol#1{\ifcase #1\or $\dagger$\or $\dagger\dagger$\else \@ctrerr\fi}
\def\fntext[#1]#2{\g@addto@macro\@fnotes{%
   \refstepcounter{fnote}\elsLabel{#1}%
   \def\thefootnote{\thefnote}
   \global\setcounter{footnote}{\c@fnote}%
   \footnotetext{#2}}}
\makeatother

\begin{document}


\newgeometry{top=2cm, bottom=2cm, left=3cm, right=3cm}
\begin{titlepage}
  \centering
  \rule{\textwidth}{1.5pt}\par
  \vspace{1em}
  {\Large \textbf{Advances in Automated Fetal Brain MRI Segmentation and Biometry: Insights from the FeTA 2024 Challenge}\par}
  \vspace{1em}
  \rule{\textwidth}{1.5pt}\par
  \vspace{1em}
  
  \vspace{.5cm}
  {Vladyslav Zalevskyi$^{1,2,*}$, 
Thomas Sanchez$^{1,2}$, 
Misha Kaandorp$^{5,43}$, 
Margaux Roulet$^{1,2}$,
Diego Fajardo-Rojas$^{3}$, 
Liu Li$^{7}$, 
Jana Hutter$^{3,4}$, 
Hongwei Bran Li$^{10,11}$, 
Matthew Barkovich$^{9}$,
Hui Ji$^{5,6}$, 
Luca Wilhelmi$^{5}$, 
Aline Dändliker$^{5}$, 
Céline Steger$^{5,6}$, 
Mériam Koob$^{1}$, 
Yvan Gomez$^{36,37}$,
Anton Jakovčić$^{44}$, 
Melita Klaić$^{44}$, 
Ana Adžić$^{44}$, 
Pavel Marković$^{44}$, 
Gracia Grabarić$^{44}$,
Milan Rados$^{44}$, 
Jordina Aviles Verdera$^{3,4}$, 
Gregor Kasprian$^{46}$, 
Gregor Dovjak$^{46}$,
Raphael Gaubert-Rachmühl$^{5}$, 
Maurice Aschwanden$^{5}$, 
Qi Zeng$^{12}$, 
Davood Karimi$^{12}$,
Denis Peruzzo$^{13}$, 
Tommaso Ciceri$^{13}$, 
Giorgio Longari$^{14}$, 
Rachika E. Hamadache$^{15}$,
Amina Bouzid$^{15}$, 
Xavier Lladó$^{15}$, 
Simone Chiarella$^{16}$, 
Gerard Martí-Juan$^{17}$,
Miguel Ángel González Ballester$^{17,38}$, 
Marco Castellaro$^{18}$, 
Marco Pinamonti$^{18}$, 
Valentina Visani$^{18}$,
Robin Cremese$^{19}$, 
Keïn Sam$^{19}$,
Fleur Gaudfernau$^{20}$, 
Param Ahir$^{21}$, 
Mehul Parikh$^{21}$,
Maximilian Zenk$^{22,39}$, 
Michael Baumgartner$^{22,39}$, 
Klaus Maier-Hein$^{22,39,40,41,42}$, 
Li Tianhong$^{23}$,
Yang Hong$^{23}$, 
Zhao Longfei$^{23}$, 
Domen Preloznik$^{24}$, 
Žiga Špiclin$^{24}$,
Jae Won Choi$^{25}$, 
Muyang Li$^{26}$, 
Jia Fu$^{26}$, 
Guotai Wang$^{26}$, 
Jingwen Jiang$^{27}$,
Lyuyang Tong$^{27}$, 
Bo Du$^{27}$, 
Andrea Gondova$^{28,47}$, 
Sungmin You$^{28,47}$, 
Kiho  Im$^{28,47,48}$,
Abdul Qayyum$^{7}$, 
Moona Mazher$^{29}$, 
Steven A Niederer$^{7}$, 
Andras Jakab$^{5,32}$,
Roxane Licandro$^{30,31}$, 
Kelly Payette$^{3,4,5}$\textsuperscript{\textdagger}, 
Meritxell Bach Cuadra$^{2,1}$\textsuperscript{\textdagger}\\\par}

\vspace{.7cm}
{
\footnotesize $^1$ \textit{Department of Radiology, Lausanne University Hospital and University of Lausanne, Lausanne, Switzerland}\\
\footnotesize $^2$ \textit{CIBM Center for Biomedical Imaging, Lausanne, Switzerland}\\
\footnotesize $^3$ \textit{Department of Early Life Imaging, School of Biomedical Engineering \& Imaging Sciences,}\\ \footnotesize\textit{King’s College London, London, UK}\\
\footnotesize $^4$ \textit{Smart Imaging Lab, University Hospital Erlangen, Erlangen, Germany}\\
\footnotesize $^5$ \textit{Center for MR-Research, University Children’s Hospital Zurich, University of Zurich, Zurich, Switzerland}\\
\footnotesize $^6$ \textit{Neuroscience Center Zurich, University of Zurich, Zurich, Switzerland}\\
\footnotesize $^7$ \textit{National Heart \& Lung Institute, Imperial College London, London, UK}\\
\footnotesize $^9$ \textit{University of California, San Francisco; UCSF Benioff Children’s Hospital, San Francisco, California, USA}\\
\footnotesize $^{10}$ \textit{Department of Quantitative Biomedicine, University of Zurich, Zurich, Switzerland}\\
\footnotesize $^{11}$ \textit{Department of Informatics, Technical University of Munich, Munich, Germany}\\
\footnotesize $^{12}$ \textit{Boston Children’s Hospital, Harvard Medical School, Boston, Massachusetts, USA}\\
\footnotesize $^{13}$ \textit{Neuroimaging Unit, Scientific Institute IRCCS E. Medea, Bosisio Parini, Italy}\\
\footnotesize $^{14}$ \textit{Department of Informatics, Systems and Communication, University of Milano Bicocca, Milan, Italy}\\
\footnotesize $^{15}$ \textit{Research Institute of Computer Vision and Robotics (ViCOROB), Universitat de Girona, Girona, Spain}\\
\footnotesize $^{16}$ \textit{Università di Bologna, Bologna, Italy}\\
\footnotesize $^{17}$ \textit{BCN MedTech, Department of Engineering, Universitat Pompeu Fabra, Barcelona, Spain}\\
\footnotesize $^{18}$ \textit{Department of Information Engineering, University of Padova, Padova, Italy}\\
\footnotesize $^{19}$ \textit{Institut Pasteur, Université Paris Cité, CNRS UMR 3571, Decision and Bayesian Computation, Paris, France}\\
\footnotesize $^{20}$ \textit{Inria, HeKA, PariSantéCampus, Paris, France}\\
\footnotesize $^{21}$ \textit{L. D. College of Engineering, Gujarat, India}\\
\footnotesize $^{22}$ \textit{Medical Faculty Heidelberg, Heidelberg University; Pattern Analysis and Learning Group, Department of Radiation Oncology, Heidelberg University Hospital, Heidelberg, Germany}\\
\footnotesize $^{23}$ \textit{Canon Medical Systems (China) Co., Ltd, China}\\
\footnotesize $^{24}$ \textit{Faculty of Electrical Engineering, University of Ljubljana, Ljubljana, Slovenia}\\
\footnotesize $^{25}$ \textit{Department of Radiology, Seoul National University Hospital, Seoul, South Korea}\\
\footnotesize $^{26}$ \textit{School of Mechanical and Electrical Engineering, University of Electronic Science and Technology of China, Chengdu, China}\\
\footnotesize $^{27}$ \textit{School of Computer Science, Wuhan University, Wuhan, China}\\
\footnotesize $^{28}$ \textit{Fetal Neonatal Neuroimaging and Developmental Science Center, Boston Children’s Hospital, Harvard Medical School, Boston, Massachusetts, USA}\\
\footnotesize $^{29}$ \textit{Hawkes Institute, Department of Computer Science, University College London, London, UK}\\
\footnotesize $^{30}$ \textit{Laboratory for Computational Neuroimaging, Athinoula A. Martinos Center for Biomedical Imaging, Massachusetts General Hospital/Harvard Medical School, Charlestown, Massachusetts, USA}\\
\footnotesize $^{31}$ \textit{Department of Biomedical Imaging and Image-guided Therapy, Computational Imaging Research Lab (CIR), Early Life Image Analysis Group, Medical University of Vienna, Vienna, Austria}\\
\footnotesize $^{32}$ \textit{University Research Priority Project Adaptive Brain Circuits in Development and Learning (AdaBD), University of Zurich, Zurich, Switzerland}\\
\footnotesize $^{36}$ \textit{Department Woman-Mother-Child, CHUV, Lausanne, Switzerland}\\
\footnotesize $^{37}$ \textit{BCNatal Fetal Medicine Research Center (Hospital Clínic and Hospital Sant Joan de Déu), Universitat de Barcelona, Barcelona, Spain}\\
\footnotesize $^{38}$ \textit{ICREA, Barcelona, Spain}\\
\footnotesize $^{39}$ \textit{German Cancer Research Center (DKFZ) Heidelberg, Division of Medical Image Computing, Heidelberg, Germany}\\
\footnotesize $^{40}$ \textit{Pattern Analysis and Learning Group, Department of Radiation Oncology, Heidelberg University Hospital, Heidelberg, Germany}\\
\footnotesize $^{41}$ \textit{Helmholtz Imaging, German Cancer Research Center (DKFZ), Heidelberg, Germany}\\
\footnotesize $^{42}$ \textit{Faculty of Mathematics and Computer Science, Heidelberg University, Heidelberg, Germany}\\
\footnotesize $^{43}$ \textit{University of Zurich, Zurich, Switzerland}\\
\footnotesize $^{44}$ \textit{Croatian Institute for Brain Research, School of Medicine, University of Zagreb, Zagreb, Croatia}\\
\footnotesize $^{46}$ \textit{Department of Biomedical Imaging and Image-Guided Therapy, Division of Neuroradiology and Musculoskeletal Radiology, Medical University of Vienna, Vienna, Austria}\\
\footnotesize $^{47}$ \textit{Division of Newborn Medicine, Boston Children’s Hospital, Harvard Medical School, Boston, Massachusetts, USA}\\
\footnotesize $^{48}$ \textit{Department of Radiology, Boston Children’s Hospital, Harvard Medical School, Boston, Massachusetts, USA}\\
\footnotesize $^{49}$ \textit{Fetal Neonatal Neuroimaging and Developmental Science Center, Boston Children’s Hospital, Harvard Medical School, Boston, Massachusetts, USA}\\
\vspace{.2cm}
\textsuperscript{\textdagger} Equal contributions\\
$^*$ Corresponding author: \texttt{vladyslav.zalevskyi@unil.ch}
}

\vspace{.7cm}
\begin{minipage}{0.85\textwidth}

\small \textbf{Abstract.} Accurate segmentation and biometric analysis are essential for studying the developing fetal brain \textit{in utero}. The Fetal Brain Tissue Annotation (FeTA) Challenge 2024 builds upon previous editions to further advance the clinical relevance and robustness of automated fetal brain MRI analysis. This year’s challenge introduced biometry prediction as a new task complementing the usual segmentation task. The segmentation task also included a new low-field (0.55T) MRI testing set and used Euler characteristic difference (ED) as a topology-aware metric for ranking, extending the traditional overlap or distance-based measures.

A total of 16 teams submitted segmentation methods for evaluation. Segmentation performance across top teams was highly consistent across both standard and low-field MRI data. Longitudinal analysis over past FeTA editions revealed minimal improvement in accuracy over time, suggesting a potential performance plateau, particularly as results now approach or surpass reported levels of inter-rater variability. However, the introduction of the ED metric revealed topological differences that were not captured by conventional metrics, underscoring its value in assessing segmentation quality. Notably, the curated low-field MRI dataset achieved the highest segmentation performance, illustrating the potential of affordable imaging systems when combined with high-quality preprocessing and reconstruction.

A total of 7 teams submitted automated biometry methods for evaluation. While promising, this task exposed a critical limitation: most submitted methods failed to outperform a simple baseline that predicted measurements based solely on gestational age, without using image data. Performance varied widely across biometric measurements and between teams, indicating both current challenges and opportunities for improvement in this area. These findings highlight the need for better integration of volumetric context and stronger modeling strategies needed for the clinical adoption of automated fetal biometry estimation.

In addition, we analyzed different dimensions of domain shifts within our data and observed that image quality was the most influential factor affecting model generalization, with Dice score differences of up to 0.10 between low- and high-quality scans. The choice of super-resolution reconstruction pipeline also had a substantial impact on segmentation performance. Other factors—such as gestational age, pathology, and acquisition site—also contributed to performance variability, but their effects were comparatively smaller.

Overall, FeTA 2024 provides a rigorous, multi-faceted benchmark for evaluating multi-class segmentation and biometry estimation in fetal brain MRI. It emphasizes the need for data-centric approaches, improved topological modeling, and greater dataset diversity to develop clinically reliable and generalizable AI tools for fetal neuroimaging.

\vspace{.7cm}
Domain Shift --- Challenge results

  \end{minipage}

\end{titlepage}
\restoregeometry
\if

\author{Vladyslav Zalevskyi$^{1,2}$\thanks{Corresponding author: vladyslav.zalevskyi@unil.ch}, 
Thomas Sanchez$^{1,2}$, 
Misha Kaandorp$^{5,43}$, 
Margaux Roulet$^{1,2}$,\\
Diego Fajardo-Rojas$^{3}$, 
Liu Li$^{7}$, 
Jana Hutter$^{3,4}$, 
Hongwei Bran Li$^{10,11}$, 
Matthew Barkovich$^{9}$,\\
Hui Ji$^{5,6}$, 
Luca Wilhelmi$^{5}$, 
Aline Dändliker$^{5}$, 
Céline Steger$^{5,6}$, 
Mériam Koob$^{1}$, 
Yvan Gomez$^{36,37}$,\\
Anton Jakovčić$^{44}$, 
Melita Klaić$^{44}$, 
Ana Adžić$^{44}$, 
Pavel Marković$^{44}$, 
Gracia Grabarić$^{44}$,\\
Milan Rados$^{44}$, 
Jordina Aviles Verdera$^{3,4}$, 
Gregor Kasprian$^{46}$, 
Gregor Dovjak$^{46}$,\\ 
Raphael Gaubert-Rachmühl$^{5}$, 
Maurice Aschwanden$^{5}$, 
Qi Zeng$^{12}$, 
Davood Karimi$^{12}$,\\ 
Denis Peruzzo$^{13}$, 
Tommaso Ciceri$^{13}$, 
Giorgio Longari$^{14}$, 
Rachika E. Hamadache$^{15}$,\\ 
Amina Bouzid$^{15}$, 
Xavier Lladó$^{15}$, 
Simone Chiarella$^{16}$, 
Gerard Martí-Juan$^{17}$,\\ 
Miguel Ángel González Ballester$^{17,38}$, 
Marco Castellaro$^{18}$, 
Marco Pinamonti$^{18}$, 
Valentina Visani$^{18}$,\\ 
Robin Cremese$^{19}$, 
Keïn Sam$^{19}$,
Fleur Gaudfernau$^{20}$, 
Param Ahir$^{21}$, 
Mehul Parikh$^{21}$,\\ 
Maximilian Zenk$^{22,39}$, 
Michael Baumgartner$^{22,39}$, 
Klaus Maier-Hein$^{22,39,40,41,42}$, 
Li Tianhong$^{23}$,\\ 
Yang Hong$^{23}$, 
Zhao Longfei$^{23}$, 
Domen Preloznik$^{24}$, 
Žiga Špiclin$^{24}$,\\ 
Jae Won Choi$^{25}$, 
Muyang Li$^{26}$, 
Jia Fu$^{26}$, 
Guotai Wang$^{26}$, 
Jingwen Jiang$^{27}$,\\ 
Lyuyang Tong$^{27}$, 
Bo Du$^{27}$, 
Andrea Gondova$^{28,47}$, 
Sungmin You$^{28,47}$, 
Kiho Im$^{28,47,48}$,\\ 
Abdul Qayyum$^{7}$, 
Moona Mazher$^{29}$, 
Steven A Niederer$^{7}$, 
Andras Jakab$^{5,32}$,\\ 
Roxane Licandro$^{30,31}$, 
Kelly Payette$^{3,4,5}$\textsuperscript{\textdagger}, 
Meritxell Bach Cuadra$^{2,1}$\textsuperscript{\textdagger}\\
\small $^1$ \textit{Department of Radiology, Lausanne University Hospital and University of}\\ 
\small \textit{Lausanne, Lausanne, Switzerland}\\
\small $^2$ \textit{CIBM Center for Biomedical Imaging, Lausanne, Switzerland}\\
\small $^3$ \textit{Department of Early Life Imaging, School of Biomedical Engineering \& Imaging Sciences, King’s}\\ 
\small \textit{ College London, London, UK}\\
\small $^4$ \textit{Smart Imaging Lab, University Hospital Erlangen, Erlangen, Germany}\\
\small $^5$ \textit{Center for MR-Research, University Children’s Hospital Zurich,}\\ 
\small \textit{ University of Zurich, Zurich, Switzerland}\\
\small $^6$ \textit{Neuroscience Center Zurich, University of Zurich, Zurich, Switzerland}\\
\small $^7$ \textit{National Heart \& Lung Institute, Imperial College London, London, UK}\\
\small $^9$ \textit{University of California, San Francisco; UCSF Benioff Children’s Hospital,}\\ 
\small \textit{ San Francisco, California, USA}\\
\small $^{10}$ \textit{Department of Quantitative Biomedicine, University of Zurich, Zurich, Switzerland}\\
\small $^{11}$ \textit{Department of Informatics, Technical University of Munich, Munich, Germany}\\
\small $^{12}$ \textit{Boston Children’s Hospital, Harvard Medical School, Boston, Massachusetts, USA}\\
\small $^{13}$ \textit{Neuroimaging Unit, Scientific Institute IRCCS E. Medea, Bosisio Parini, Italy}\\
\small $^{14}$ \textit{Department of Informatics, Systems and Communication, University of}\\ 
\small \textit{ Milano Bicocca, Milan, Italy}\\
\small $^{15}$ \textit{Research Institute of Computer Vision and Robotics (ViCOROB),}\\ 
\small \textit{ Universitat de Girona, Girona, Spain}\\
\small $^{16}$ \textit{Università di Bologna, Bologna, Italy}\\
\small $^{17}$ \textit{BCN MedTech, Department of Engineering, Universitat Pompeu Fabra, Barcelona, Spain}\\
\small $^{18}$ \textit{Department of Information Engineering, University of Padova, Padova, Italy}\\
\small $^{19}$ \textit{Institut Pasteur, Université Paris Cité, CNRS UMR 3571,}\\ 
\small \textit{ Decision and Bayesian Computation, Paris, France}\\
\small $^{20}$ \textit{Inria, HeKA, PariSantéCampus, Paris, France}\\
\small $^{21}$ \textit{L. D. College of Engineering, Gujarat, India}\\
\small $^{22}$ \textit{Medical Faculty Heidelberg, Heidelberg University; Pattern Analysis}\\ 
\small \textit{ and Learning Group, Department of Radiation Oncology, Heidelberg University}\\ 
\small \textit{ Hospital, Heidelberg, Germany}\\
\small $^{23}$ \textit{Canon Medical Systems (China) Co., Ltd, China}\\
\small $^{24}$ \textit{Faculty of Electrical Engineering, University of Ljubljana, Ljubljana, Slovenia}\\
\small $^{25}$ \textit{Department of Radiology, Seoul National University Hospital, Seoul, South Korea}\\
\small $^{26}$ \textit{School of Mechanical and Electrical Engineering, University}\\ 
\small \textit{ of Electronic Science and Technology of China, Chengdu, China}\\
\small $^{27}$ \textit{School of Computer Science, Wuhan University, Wuhan, China}\\
\small $^{28}$ \textit{Fetal Neonatal Neuroimaging and Developmental Science Center,}\\ 
\small \textit{ Boston Children’s Hospital, Harvard Medical School, Boston, Massachusetts, USA}\\
\small $^{29}$ \textit{Hawkes Institute, Department of Computer Science, University College London, London, UK}\\
\small $^{30}$ \textit{Laboratory for Computational Neuroimaging, Athinoula A. Martinos Center}\\ 
\small \textit{ for Biomedical Imaging, Massachusetts General Hospital/Harvard Medical School,}\\ 
\small \textit{ Charlestown, Massachusetts, USA}\\
\small $^{31}$ \textit{Department of Biomedical Imaging and Image-guided Therapy,}\\ 
\small \textit{ Computational Imaging Research Lab (CIR), Early Life Image Analysis Group,}\\ 
\small \textit{ Medical University of Vienna, Vienna, Austria}\\
\small $^{32}$ \textit{University Research Priority Project Adaptive Brain Circuits}\\ 
\small \textit{ in Development and Learning (AdaBD), University of Zurich, Zurich, Switzerland}\\
\small $^{36}$ \textit{Department Woman-Mother-Child, CHUV, Lausanne, Switzerland}\\
\small $^{37}$ \textit{BCNatal Fetal Medicine Research Center (Hospital Clínic}\\ 
\small \textit{ and Hospital Sant Joan de Déu), Universitat de Barcelona, Barcelona, Spain}\\
\small $^{38}$ \textit{ICREA, Barcelona, Spain}\\
\small $^{39}$ \textit{German Cancer Research Center (DKFZ) Heidelberg, Division of Medical}\\ 
\small \textit{ Image Computing, Heidelberg, Germany}\\
\small $^{40}$ \textit{Pattern Analysis and Learning Group, Department of Radiation Oncology,}\\ 
\small \textit{ Heidelberg University Hospital, Heidelberg, Germany}\\
\small $^{41}$ \textit{Helmholtz Imaging, German Cancer Research Center (DKFZ), Heidelberg, Germany}\\
\small $^{42}$ \textit{Faculty of Mathematics and Computer Science, Heidelberg University, Heidelberg, Germany}\\
\small $^{43}$ \textit{University of Zurich, Zurich, Switzerland}\\
\small $^{44}$ \textit{Croatian Institute for Brain Research, School of Medicine, University of Zagreb, Zagreb, Croatia}\\
\small $^{46}$ \textit{Department of Biomedical Imaging and Image-Guided Therapy, Division}\\ 
\small \textit{ of Neuroradiology and Musculoskeletal Radiology, Medical University of Vienna, Vienna, Austria}\\
\small $^{47}$ \textit{Division of Newborn Medicine, Boston Children’s Hospital,}\\ 
\small \textit{ Harvard Medical School, Boston, Massachusetts, USA}\\
\small $^{48}$ \textit{Department of Radiology, Boston Children’s Hospital,}\\ 
\small \textit{ Harvard Medical School, Boston, Massachusetts, USA}\\
}

\date{}

\section{Introduction}
\label{1intro}
The fetal brain undergoes rapid and complex development throughout gestation, influenced by both genetic and environmental factors. Understanding this dynamic process is critical in both clinical and research domains, as neurodevelopmental disruptions are linked to congenital anomalies and long-term cognitive or physiological impairments \citep{griffiths2017use, ciceri2024fetalatlasses, VandenBergh2018}. In vivo imaging biomarkers derived from ultrasonography (US) or magnetic resonance imaging (MRI) provide non-invasive and quantifiable metrics to monitor prenatal brain development. Deviations from normative patterns in these biomarkers have been associated with a range of pathologies, including corpus callosum \citep{Marathu2024, Lamon2024} and posterior fossa malformations \citep{Dovjak2020, Mahalingam2021}, ventriculomegaly \citep{chen2024machine}, and have been shown to correlate with neurodevelopmental outcomes in conditions such as congenital heart disease \citep{Sadhwani2022}, intrauterine growth restriction \citep{EgaaUgrinovic2015, Meijerink2023}, and preterm birth \citep{Story2021, Hall2024}.

Fetal brain MRI has emerged as an important non-invasive tool for studying neurodevelopment in utero and diagnosing congenital disorders, complementing ultrasonography~\citep{griffiths2017use, alamo2010fetal}. Accurate and automatic segmentation of fetal brain tissues in MRI is critical for quantitative analysis and biomarker extraction, including tissue volumetry, cortical morphometry \citep{payette2023feta_2021rseults}, and biometric measurements \citep{she2023automatic}. Manual segmentation, however, remains labor-intensive, error-prone, and susceptible to inter-observer variability, underscoring the necessity of reliable automated techniques.

While clinical US and 2D MRI are the standard techniques for assessing fetal development \citep{tilea2009cerebral}, the use of super-resolution reconstruction (SRR) techniques to generate 3D fetal brain reconstructions has emerged as a powerful advancement. SRR methods fuse multiple 2D MRI slices (often motion-corrupted) into a single, enhanced 3D motion-corrected volume, significantly improving brain analysis \citep{GAFNER2020109369, Avisdris2021, Matthew2024}. Recent studies have shown that biometric measurements derived from 3D SRR volumes correlate strongly with those from ultrasound, while offering greater rater confidence than using 2D MRI series \citep{Lamon2024, Khawam2021, GAFNER2020109369, sanchez2024biometry, Kyriakopoulou2016, Ciceri2023}.

The Fetal Tissue Annotation (FeTA) challenges, held in 2021 \citep{payette2023feta_2021rseults} and 2022 \citep{payette2024feta_2022rseults}, have significantly advanced fetal brain MRI analysis by providing public datasets and standardized evaluation protocols for brain tissue segmentation. The \textbf{FeTA 2024 challenge} builds on previous editions, retaining the core \textbf{brain tissue segmentation} task and introducing a new clinically relevant objective: \textbf{biometry extraction}, alongside several other key innovations.

Firstly, FeTA 2024 introduces a new low-field (LF, 0.55T) MRI testing dataset. LF MRI offers a low-cost alternative to 1.5–3T systems, making it especially valuable in resource-limited settings~\citep{arnold2023low, marques2019low}. This affordability supports research in low- and middle-income countries with large pediatric populations, where access to high-field MRI is limited, hindering studies on brain development under normal and adverse conditions~\citep{Murali2023, AvilesVerdera2023}.

Secondly, we introduced the Euler characteristic difference as an additional ranking metric for segmentation~\citep{taha2015metrics}. Unlike overlap- or distance-based metrics, it captures topological correctness, offering a complementary view of performance~\citep{maier2024metrics}. This is especially relevant for downstream tasks like cortical surface extraction or morphometric analyses (e.g., sulcal folding, cortical maturation, or structural abnormality assessment)~\citep{Yehuda2023, Clouchoux2011}.

In FeTA 2024, we promote the development of generalizable, fully automated methods for fetal brain analysis, enabling the extraction of key imaging biomarkers via multi-class tissue segmentation and biometry across diverse acquisition and reconstruction settings. This paper offers \textbf{a comprehensive overview of the challenge}, covering its organization, submitted algorithms, performance assessment, benchmarking, and evaluation using the BIAS reporting framework, which emphasizes transparency, reproducibility, and fairness~\citep{BIAS}.
We also analyze performance trends over time, tracking improvements in state-of-the-art segmentation accuracy across FeTA editions. Finally, we assess \textbf{data quality} in both training and testing sets to examine its impact on the generalization of submitted methods. Combined with other domain shifts—such as gestational age, pathology, super-resolution reconstruction, and acquisition site—our work provides a deep \textbf{ overview of how domain shifts affect deep learning models for fetal brain analysis} and informs strategies to mitigate their impact.

\section{Methods}

\subsection{Challenge organization}

\paragraph{Context} The FeTA 2024 challenge was held as a thematic event within the Perinatal, Preterm, and Pediatric Image Analysis (PIPPI) workshop\footnote{\url{https://pippiworkshop.github.io/}}, part of the Medical Image Computing and Computer-Assisted Intervention (MICCAI) 2024 conference. 
The challenge was run through a custom platform, available at \url{https://fetachallenge.github.io/}, which provided participants with all the necessary information on the organization, time frame, and submission instructions.

\paragraph{Data, participation and submission} 
Challenge participation required submission of \textbf{fully automated} segmentation and/or biometry algorithms. A training set of 3D super-resolution fetal brain MRI from two institutions was provided; no validation set was released, and test data remained private for evaluation. Participants could use publicly available external datasets and pre-trained models, provided these were public and fully documented in the algorithm description, as well as use both 2D and 3D models.

Participants submitted their algorithms as Docker containers with a command-line interface for test data evaluation\footnote{Instructions: \url{https://github.com/fetachallenge/fetachallengesubmission}.}. Any programming language was allowed, provided the input/output followed the evaluation utility specifications. Each team was allowed one submission, except in cases of technical errors (e.g., Docker issues), which could be corrected upon notification by the organizers. Evaluation on test data was performed by the organizers using publicly available code\footnote{Available at \url{https://fetachallenge.github.io/pages/Evaluation}.}.
To promote transparency and reproducibility, FeTA 2024 encouraged participants to share their code publicly. A Docker Hub page (\url{https://hub.docker.com/repositories/fetachallenge2024}) was created to host containers from teams who agreed to release their Docker images. 

\paragraph{Timetable, rewards and results paper}
The challenge followed a predefined schedule: training data was released on May 21, 2024; registration opened after challenge acceptance. The Docker submission deadline was extended to August 4, 2024, and algorithm descriptions were due by August 12. On August 23, the top five teams were invited to prepare 2-minute pitch presentations for the challenge day and, along with all participants, were invited to present posters at the dedicated conference session.
The challenge took place in person on October 6, 2024, during MICCAI. Results were announced live and later published on the challenge website, along with top teams' presentations (with their consent).
The top three teams in each task received certificates and small gifts, including a 3D-printed fetal brain keychain for in-person attendees. The highest-ranking team in each task also received a box of artisanal Swiss-made cookies.
Organizers could participate but were not eligible for awards.

All teams with valid submissions and interest in the publication were included in this results paper, with up to three members per team listed as co-authors. Teams were free to publish their algorithms and results independently after the challenge, without embargo, provided they cited both the data publication \citep{Payette2021_naturefetadata} and this summary paper.

\paragraph{Data usage terms and conflicts of interest}
The training data from the University Children's Hospital Zürich (\textbf{Kispi}) and General Hospital Vienna/Medical University of Vienna were provided with specific licensing conditions. Kispi data, hosted on the Synapse platform\footnote{\url{https://www.synapse.org/Synapse:syn25649159/wiki/610007}}, \textbf{is for non-commercial use only}. \textbf{Vienna data} is governed by a custom Data Transfer Agreement, \textbf{allowing use for challenge purposes only}. Participants could modify the data, including generating synthetic data through augmentation, as long as modifications were documented and synthetic data could be provided to the organizers upon request.
None of the organizers participated in this year’s challenge or have conflicts of interest to disclose. The challenge awards were funded by the institutional budget (Kispi), and none of the participants were involved in funding. Only organizers at Kispi had full access to the testing dataset, as they managed data transfer agreements with all providers.

\begin{figure*}[t]
    \centering
    \includegraphics[width=0.7\linewidth]{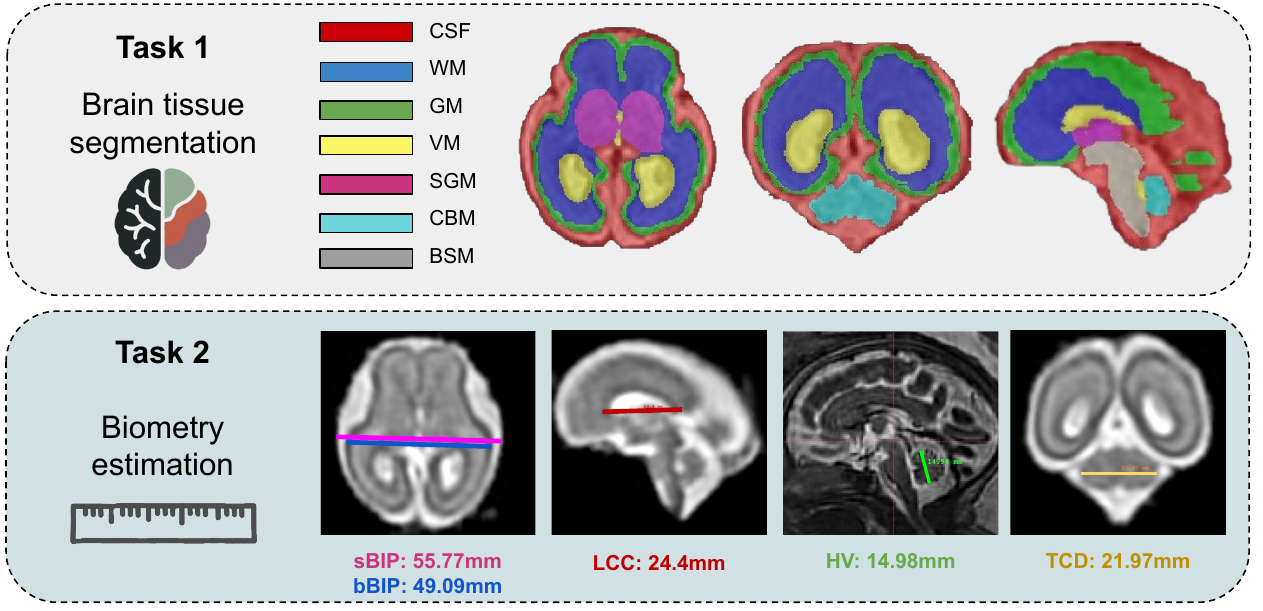}
    \caption{FeTA 2024 Challenge Tasks. Task 1 involves segmentation of fetal brain tissues into seven classes, while Task 2 focuses on estimating five biometric measurements, both illustrated in the figure.}
    \label{fig:1tasks}
\end{figure*}

\subsection{Challenge tasks}
\looseness=-1
The FeTA challenge presents two primary tasks (see Figure \ref{fig:1tasks}). Participants could choose to compete in either or both tasks.

\paragraph{\textbf{Task 1. Fetal brain tissue segmentation}} This task aims to develop algorithms that automatically delineate different tissues in SRR fetal brain MRI. The 3D semantic segmentation involves classifying each voxel into one of seven predefined classes: Background, External CSF, Grey Matter (GM), White Matter (WM), Ventricles including cavum (VM), Cerebellum (CBM), Deep Grey Matter (SGM), and Brainstem (BSM). Reference annotation procedures and inter-rater variability analyses for all datasets (except the new LF set) are detailed in \citet{Payette2021_naturefetadata} and \citet{payette2024feta_2022rseults}. The LF dataset followed the same annotation protocol, with seven annotators (AJ, CS, RG, VZ, YG, MA, MR) each segmenting a specific label map. These were merged into a single reference annotation, reviewed, and corrected by two fetal MRI experts (KP, AJ).

\paragraph{\textbf{Task 2. Biometric measurements prediction}} The goal of this task is to develop algorithms that automatically and accurately estimate key fetal brain biometry from MRI. The selected measurements—length of the corpus callosum (LCC), height of the vermis (HV), brain biparietal diameter (bBIP), skull biparietal diameter (sBIP), and transverse cerebellar diameter (TCD)—were chosen to \textbf{minimize annotation burden} while providing \textbf{complementary anatomical and diagnostic value}. Four raters contributed: YG (5 years’ fetal MRI experience), MKo (16 years), and junior raters RG and MA (reviewed by AJ, 12 years)\footnote{Biometry annotation protocol is described on our website: \url{https://fetachallenge.github.io/pages/Data_description}}. Not all measurements were available for all cases: in the test set, 15 cases lacked LCC, one lacked HV, and one lacked TCD due to annotator uncertainty. In the training set, 102 of 120 cases had complete annotations—10 Kispi cases were excluded for poor quality; 5 Kispi and 3 Vienna cases had partial annotations. While the main goal was to predict biometry values, the training set also included 3D \textbf{landmark annotations}—single-voxel labels marking anatomical structures used to derive each measurement. Clinicians identified these landmarks during annotation, and the actual biometry values were computed via organizer-provided scripts. Both the landmarks and scripts were shared, allowing participants to either regress biometry directly or predict landmarks, followed by automated biometry measurement.

\begin{figure*}[ht]
    \centering
    \includegraphics[width=.85\linewidth]{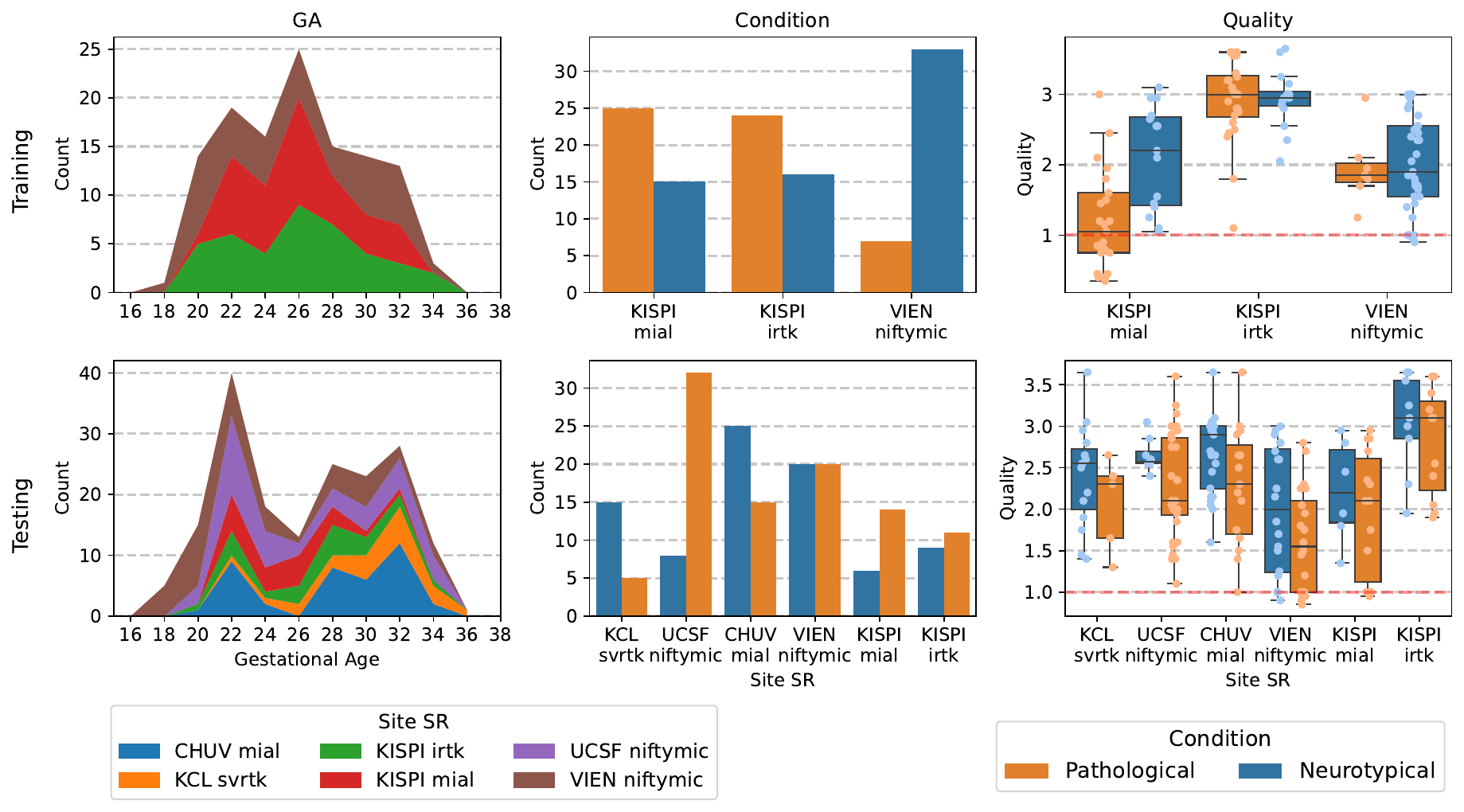}
   \caption{FeTA 2024 data distribution by GA (weeks), condition, and image quality (0 = lowest, 4 = highest; 1 = minimum acceptable), stratified by Site and SR method for training (top) and testing (bottom) sets. In the image quality plots, the red dotted line marks the threshold score (1.0); images with a score below this value are classified as poor quality.}
    \label{fig:gapatholsitesplit}
\end{figure*}

\subsection{Challenge data sets}
Subject selection aimed to ensure a representative cohort spanning 18–35 weeks of gestation, including both neurotypical and pathological cases (e.g., spina bifida, ventriculomegaly, corpus callosum malformations) to reflect clinical practice. UCSF and CHUV data were acquired during clinical fetal MRI scans following ultrasound referral, performed by trained medical staff. Data from KCL, Kispi, and Vienna were collected using research protocols. All cohorts had approval by the local ethics committee for use in the challenge after anonymization\footnote{\textbf{KISPI:} Ethical Committee of the Canton of Zurich, Switzerland (Decision numbers: 2017 00885, 2016 01019, 2017 00167). \textbf{CHUV}:  Ethics Committee of the Canton de Vaud, Switzerland (CER-VD 2021 00124).  \textbf{Vienna:} Approved by the ethics review board and data clearing department at the Medical University of Vienna. \textbf{UCSF:} institutional review board (IRB 16 20619). \textbf{KCL}: Ethics
Committee Dulwich (Ethics code 19 LO 0852).}.

Each case included a 3D fetal brain MRI reconstruction, manual brain tissue segmentation, and biometry annotations. Metadata included gestational age (GA) and a binary label indicating neurotypical or pathological status. To preserve anonymity, gender was excluded and GA was randomly offset by ±3 days.


The challenge dataset comprises of 120 training and 180 test cases. The test set was split into \textit{in-domain} (from the same institutions and protocols as training data) and \textit{out-of-domain} cases. To ensure balance, both subsets were similar in size to the training set. Demographic characteristics, including GA and pathology distribution, were matched across training and testing cohorts (see Figure \ref{fig:gapatholsitesplit}). For this year’s challenge, manual quality control was performed on all training and testing cases following the protocol by \citet{sanchez2024assessing}, ensuring comparable data quality between training and testing sets (Figure~\ref{fig:gapatholsitesplit}).


The FeTA 2024 training and testing datasets are identical to those used in FeTA 2022, with two additions: a new low-field out-of-domain test set from King’s College London (KCL) and manual biometry annotations.

All cases were acquired using T2-weighted single-shot fast spin-echo sequences\footnote{Also known as HASTE (Siemens), SSTSE (Philips), or SSFSE (GE), depending on the scanner manufacturer.}, the standard for structural fetal MRI due to their high signal-to-noise ratio and reduced sensitivity to fetal motion. To further mitigate motion artifacts, multiple stacks were acquired in various orientations (axial, sagittal, coronal, and off-plane). Manual selection of 2D stacks was done at each site and then combined into a single high-resolution, isotropic 3D image via super-resolution reconstruction. The resulting 3D volumes were zero-padded to 256x256x256 and reoriented to a standard radiological plane. A summary of acquisition parameters, demographic characteristics, and reconstruction methods in all sites is provided in Table~\ref{tab:datasets}. Additional details about the new KCL dataset are provided below. For further detailed information on the FeTA 2022 acquisitions, please refer to \citet{payette2024feta_2022rseults}.

\begin{table*}
\centering
\caption{FeTA 2024 datasets properties. $N_n$ - number of neurotypical subjects, $N_p$ - number of pathological subjects. "+" indicates the minimum TE value}
\resizebox{.95\linewidth}{!}{%
\begin{tabular}{c|cclclcccc}
\toprule
\textbf{Used} & \textbf{Testing} & \multirow{2}{*}{\textbf{Institution}} & \multicolumn{1}{c}{\multirow{2}{*}{\textbf{Scanner}}} & \multirow{2}{*}{\textbf{N}} & \multicolumn{1}{c}{\textbf{SR}} & \textbf{SR res.} & \textbf{TR/TE} & \textbf{GA} & \multirow{2}{*}{$\mathbf{N_n/N_p}$} \\
\textbf{for} & \textbf{domain} & & & & \multicolumn{1}{c}{\textbf{method}} & (mm$^3$) & (ms) & (weeks) &\\
\midrule
\multirow[c]{2}{*}{\rotatebox[origin=c]{90}{\small\textbf{Training}}} & \multirow[c]{2}{*}{\rotatebox[origin=c]{90}{\begin{tabular}{c} In \\ domain \end{tabular}}} & KISPI & \begin{tabular}{l} GE Signa Discovery \\ MR450/MR750 \\ (1.5T/3T respectively)* \end{tabular} & 80 & \hspace{-.3cm}\begin{tabular}{l} MIALSRTK (40) \\ IRTK-simple (40) \end{tabular} & (0.5)$^3$ & \begin{tabular}{c} 2000-3500/\\ 120+ \end{tabular} & 20-34.4 & 49/31\\ 
&& Vienna & \begin{tabular}{l}Philips Ingenia/Intera (1.5T)\\ Philips Achieva (3T)* \\ \end{tabular} & 40 & NiftyMIC & (1.0)$^3$ & \begin{tabular}{c} 6000-22000/\\80-140 \end{tabular} & 19.3-34.4 & 33/7\\
\midrule
\multirow[c]{5}{*}{\rotatebox[origin=c]{90}{\hspace{-2.5cm}\small\textbf{Testing}}} & \multirow[c]{2}{*}{\rotatebox[origin=c]{90}{\begin{tabular}{c} In \\ domain \end{tabular}}} & KISPI & \begin{tabular}{l} GE Signa Discovery \\ MR450/MR750 \\ (1.5T/3T respectively)* \end{tabular} & 40 & \hspace{-.3cm}\begin{tabular}{l} MIALSRTK (20) \\ IRTK-simple (20) \end{tabular} & (0.5)$^3$ & \begin{tabular}{c} 2000-3500/\\ 120+ \end{tabular} & 21.3-34.6 & 15/25\\ 
&& Vienna & \begin{tabular}{l}Philips Ingenia/Intera (1.5T)\\ Philips Achieva (3T)* \\ \end{tabular} & 40 & NiftyMIC & (1.0)$^3$ & \begin{tabular}{c} 6000-22000/\\80-140 \end{tabular} & 18.1-35.5 & 20/20\\
\cmidrule(l){2-10}
& \multirow[c]{3}{*}{\rotatebox[origin=c]{90}{\hspace{-1cm}\begin{tabular}{c} Out of \\ domain \end{tabular}}} & CHUV & \begin{tabular}{l} Siemens MAGNETOM \\ Aera (1.5T) \end{tabular} & 40 &  MIALSRTK & (1.125)$^3$ & 1200/90 & 21.0-35.0 & 25/15\\ 
&& UCSF & \begin{tabular}{l} GE Signa Discovery\\ MR750/MR750W (3T) \\ \end{tabular} & 40 & NiftyMIC & (0.8)$^3$ & \begin{tabular}{c} 200-3500/\\100+ \end{tabular} & 20.0-35.1 & 8/32\\
&& KCL & \begin{tabular}{l} Siemens MAGNETOM\\ Free.Max (0.55T) \\ \end{tabular} & 20 & SVRTK & (0.8)$^3$ & 2500/106 & 21.0-35.0 & 15/5\\
\bottomrule
\end{tabular}}
\vspace{-.3cm}
\begin{flushleft}
\noindent\tiny{*The training dataset contained data from both 1.5T and 3T scanners. However, which cases belonged to which scanner were not provided to the participants as it was part of the data anonymization process. Therefore, the breakdown of number of cases per scanner is not provided here.}
\end{flushleft}
\label{tab:datasets}
\end{table*}

Data from \textbf{KCL} was collected using a 0.55T low-field MRI scanner (Siemens MAGNETOM Free.Max) with a HASTE sequence as part of a prospective single-center study and fully anonymized following local procedures (Ethics Committee Dulwich 19 LO 0852). The acquired stacks had a resolution of 1.5mm x 1.5mm x 4.5mm, which were then reconstructed into a high-resolution volume of 0.8mm x 0.8mm x 0.8mm using SVRTK \citep{Uus2020}. Key acquisition parameters include a flip angle of 180°, a field-of-view of 450 × 450mm², and a base resolution of 304x304 pixels, yielding a voxel size of 1.5 × 1.5  × 4.5mm³. The acquisition time ranged from 64 to 122 seconds. Data collection took place at St Thomas’ Hospital in London, United Kingdom, without the use of maternal or fetal sedation. All acquisitions were performed using the contour L coil and the integrated spine coil while the mother was in a supine position.  This dataset is used only in the testing set. 


\subsection{Evaluation Metrics}
We provide a short recall of the ranking metrics. The detailed mathematical formulation is available in supplementary materials A1.
\paragraph{Task 1. Segmentation}
Performance of segmentation algorithms is comprehensively assessed through complementary metrics of spatial overlap, volume, shape, and topological correctness:
\begin{itemize}
    \item \textbf{Dice Similarity Coefficient (Dice; $\uparrow$)\footnote{$\uparrow$ means that a higher score is better and $\downarrow$ that a lower score is better}:} measures voxel-wise correspondence between the predicted and ground truth (GT) segmentations.
    \item \textbf{Volume Similarity (VS; $\uparrow$):} measures the similarity of the volumes between the predicted and GT segmentations.
    \item \textbf{Hausdorff Distance (HD95; $\downarrow$):} quantifies the distance between contours of the predicted and GT segmentations with robustness to outliers.
    \item \textbf{Euler Characteristics Difference (ED; $\downarrow$)}: evaluates the topological similarity between the predicted and GT segmentations. 
    
\end{itemize}
\label{par:euler}
As ED is included in the ranking for the first time, we describe it further. It is based on the Euler characteristic (EC):
   \[
   EC = \text{BN}_0 - \text{BN}_1 + \text{BN}_2
   \]
   
   where  Betti Number \(\text{BN}_0\) represents the number of connected components (i.e., regions),  \(\text{BN}_1\) represents the number of loops or holes and \(\text{BN}_2\) represents the number of voids or cavities. The ED difference is then computed as $|EC_{pred} - EC_{GT}|$. Smaller differences indicate better topological alignment. The Betti number values of GT are: for all brain tissue labels, \( \text{BN}_1 = 0 \) and \( \text{BN}_2 = 0 \).  For the eCSF, WM, ventricles, cerebellum, dGM, and brainstem, \( \text{BN}_0 = 1 \), while for GM, \( \text{BN}_0 = 2 \). 
   
\paragraph{Task 2: Biometry Estimation}
The primary metric for evaluating biometry estimation algorithms is \textbf{mean average percentage error (MAPE; $\downarrow$)} , which quantifies the error in the estimated biometric measurements relative to the actual measurements:
\[
MAPE =  \frac{1}{N} \sum_{i=1}^N \frac{|  y_i - \hat{y}_i |}{\text{$y_i$}} \times 100,
\]
where \(y_i\) and \(\hat{y}_i\) are GT and predicted measurements respectively, and \(N\) is the total number of measurements. This metric accounts for variable sizes of the target structures and is used to assess the accuracy of the estimated biometric measurements.


\subsection{Ranking}
\label{ranking}
Submissions are ranked based on metrics computed for each brain tissue label (or biometric measurement) in the predicted maps of the fetal brain volumes. For segmentation, the final rank is the average of all 4 metrics: Dice, HD95, VS, and ED. For biometry, the final rank is based on MAPE. For metrics where higher values are better (Dice, VS), the algorithm with the highest value ranks best. For metrics where lower values are better (like HD95 and ED for segmentation and MAPE for biometry tasks), the algorithm with the lowest value ranks best. The individual label rankings are summed, and the algorithm with the highest combined rank is considered the best.

In cases of missing results (e.g., if an algorithm fails to detect a label or if the entire label map is empty), the worst possible values will be assigned to the algorithm. For example, if a label is missing in the label map, it will receive a Dice and VS of 0. For  HD95, EC, and MAPE, the missing values are set to double the maximum value of other algorithms for that sub-ranking. This ensures that algorithms with missing results are ranked last for that specific task/brain tissue.

\subsubsection{Biometry baselines}
In the ranking of Task 2, two additional baseline models representing lower and upper performance limits were incorporated as separate submissions. These entries, intended solely for benchmarking purposes, were not considered in the formal determination of the challenge competition ranking.

\paragraph{Lower bound: Gestational age regression model}
This \textit{model}, referred to as \textbf{[GA]} in the result's Table \ref{tab:biom_ranks}, is a simple univariate linear regression baseline. For each biometric measurement \( y \), the model predicts its value \( \hat{y}\) using the gestational age (\(\text{GA}\)) as the sole explanatory variable, mathematically:  
\[
\hat{y} = \beta_0 + \beta_1 \cdot \text{GA},
\]
\looseness=-1
where \( \beta_0 \) is the intercept and \( \beta_1 \) is the regression coefficient learned from the training data. This baseline does not rely on the image and aims at quantifying how strongly the GA can account for the size of a given structure.

\paragraph{Upper bound: Inter-rater variability}
The upper bound is set by averaging inter-rater variability, further denoted as \textbf{[inter-rater]}. This reflects the best-expected accuracy, accounting for measurement errors and uncertainties between manual raters. For each biometric measurement in the test dataset, annotations from two independent observers are used by comparing one observer’s measurement to the other’s, with the result averaged across all test cases.

\subsection{Statistical analysis}
The non-parametric \textbf{Wilcoxon signed-rank} test was used to assess performance differences between algorithms, as the Shapiro-Wilk test indicated non-normal distribution. To evaluate performance differences across subsets (e.g., neurotypical vs. pathological cohorts), we applied the \textbf{Mann-Whitney U test} (Wilcoxon rank-sum test).  For all tests, statistical significance was set at $p<0.05$. For multiple comparisons, such as between sites or labels, we applied \textbf{Bonferroni correction}.

\subsection{Further analysis}
FeTA 2024, as the third edition of the challenge, provides an opportunity to assess progress and unsolved challenges. We report two additional analyses: \textbf{(i)} the evolution of top-performing segmentation models over the last three editions, and \textbf{(ii)} the impact of different domain shift sources on model performance.

\subsubsection{Insights from three years of competition: progress or plateau?}
To assess progress in fetal brain tissue segmentation, we analyze the evolution of top-performing algorithms over time. Specifically, we compare the performance of the highest-ranked teams from the FeTA challenges in 2021, 2022, and 2024, evaluating segmentation accuracy across the dataset splits available in each respective year.

To extend the longitudinal comparison, we perform a retrospective evaluation of the 2022 winning method on the KCL dataset, first introduced as a test set in 2024. This is enabled by the 2022 winning team’s release of their Docker container\footnote{\url{https://hub.docker.com/r/fetachallenge22/feta-imperial-tum-2022-nnunet}}, allowing us to assess the generalization of a previously state-of-the-art solution to new, unseen data, and to identify both progress and persistent limitations.

\begin{figure*}
    \centering
    \includegraphics[width=.8\linewidth]{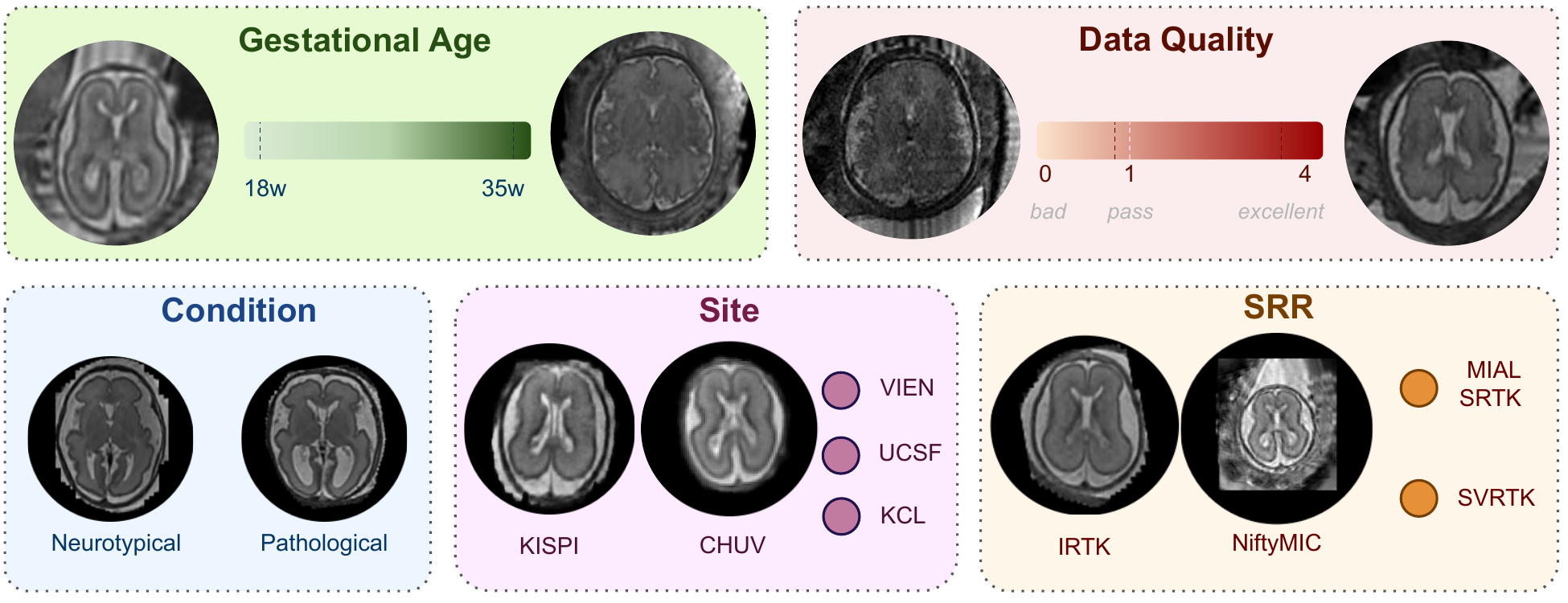}
    \caption{Illustration of the sources of domain shifts in fetal brain MRI datasets of FeTA 2024. Demonstrated across gestational age (18 vs. 35 weeks), data quality (0.9 vs. 3.64), clinical condition (neurotypical vs. pathological), acquisition site (KISPI vs. CHUV), and SRR methods (IRTK vs. NiftyMIC). In each comparison, only the indicated domain is varied, while all other domains remain constant. Additional domains within each source, not shown here, are represented by circles.}
    \label{fig:domain_shifts}
    \vspace{-.5cm}
\end{figure*}

\subsubsection{Quantifying domain shifts}
Domain shifts remain a key obstacle in fetal brain MRI analysis, often undermining model generalizability. These shifts arise from variations in subject demographics, imaging protocols, scanner types, and reconstruction methods \citep{Dockès2021_domshift}. In fetal imaging, GA notably affects brain morphology and contrast, while pathologies such as ventriculomegaly, for example, can significantly alter anatomical structure. Beyond biological and acquisition-related variability, low contrast or motion artifacts can degrade reconstruction quality, adversely affecting segmentation and biometry.\\

\noindent\textit{Is image quality a domain in itself?} To assess whether image quality impacts model generalization, we manually rated the quality of all 180 test volumes using the protocol from \citet{sanchez2024assessing} and explored the interaction of data quality with the performance of the submitted algorithms across the test data.

\paragraph{Comparing the impact of domain shift factors}
To assess how domain shifts influence segmentation performance, we examine six key sources of variability: image quality, GA, condition (neurotypical or pathological), acquisition site, testing domain (seen vs. unseen during training), and the SRR method. These factors are summarized in Figure~\ref{fig:domain_shifts}. To evaluate the influence of domain shift factors on segmentation performance, we trained a random forest regressor for each metric of interest (Dice, HD95, Volume Similarity, Euler Difference), using six dataset-level variables as input features. Target values were defined as the average metric scores across the top 3 teams. To estimate feature importance, we applied SHapley Additive exPlanations (SHAP)~\citep{NIPS2017_7062_SHAP}, which quantify the contribution of each feature by computing its average marginal effect across all possible feature combinations. This approach provides a unified and interpretable measure of how each factor affects performance.

\section{Results}
\subsection{Algorithm description}
We received 176 access requests for the KISPI training cohort hosted on Synapse during the challenge active period (May–July 2024). However, not all of these requests were related to the FeTA challenge, as the dataset is also available for broader research purposes. For the Vienna dataset, 53 data access applications were submitted, but only 30 applicants completed the data transfer agreement process and successfully received the data.

For the segmentation task, we received 16 valid submissions, all evaluated on the full test set. One team declined participation in this paper and was excluded from the analysis; results from the remaining 15 teams are presented. For the biometry task, 7 teams submitted results. One team (\texttt{falcons}) failed to generate valid outputs for all test cases and was penalized accordingly, as described in the Section~\ref{ranking}. Notably, all biometry participants also submitted segmentation entries, leveraging segmentation outputs either as a preprocessing step or direct input for biometry estimation. A detailed description of each algorithm is provided in the supplementary materials (appendix A2) and summarized in Tables~\ref{tab:segm_alg_descr} and~\ref{tab:biom_agl_descr}. 

\subsubsection{Common data and model augmentation strategies}
Participants adopted a variety of approaches, with the majority utilizing 3D architectures—14 out of 16 for the segmentation task and 6 out of 7 for the biometry task. Across both tasks, two strategies were commonly used: data augmentation and model ensembling.

\textbf{Data augmentation} was universally applied, with all segmentation (16 teams) and biometry (7 teams) models using it. Standard transformations like flipping, rotation, scaling, and intensity shifts were common, while advanced methods, such as SynthSeg \citep{billot2023synthseg} or global intensity non-linear augmentations (GIN) \citep{ouyang2022causality}, were used by 3 teams. Some teams also simulated domain-specific artifacts, including fetal motion and bias field.

\textbf{Ensembling} was a key approach in segmentation, used by 14 out of 16 teams. This included combining models trained on different cross-validation splits (4 teams) or using varied architectures, training setups, data orientations, or augmentation schemes (8 teams). Some also integrated pre- or post-processing models, like denoising autoencoders or skull-stripping (2 teams). Ensembling was less common in the biometry task, with only 2 out of 7 teams employing it, as most models built biometry predictions in a single pipeline on top of segmentation outputs.

\subsubsection{Segmentation models}
Among the 16 submissions, the most common architectures were \textbf{nnU-Net} \cite{isensee2018nnu} (9 teams) and \textbf{U-Net} \citep{ronneberger2015u,ciek2016} (6 teams), often used as baselines. Many teams enhanced these models with \textbf{attention mechanisms} \citep{vaswani2017attention}, \textbf{residual connections} \citep{he2016identity}, or \textbf{ensembling}. Others explored alternatives
such as \textbf{Swin Transformers} \citep{liu2021swin}, custom U-Net variants, or hybrid CNN–Transformer designs. Most models were developed in \textbf{PyTorch} (12 teams), with parameter counts ranging from 5M to 140M (median: 31M, mean: 44.8M).
\vfill\null
\newpage

Use of \textbf{external data} was limited to 5 teams, primarily leveraging dHCP data \citep{Hughes2016, Edwards2022}, fetal brain atlases \citep{Gholipour2017_atlass, kcl_dhcp_atals}, or foundation models pretrained on large-scale image datasets \citep{mednextpaper}. 

\subsubsection{Biometry models}
All biometry models leveraged segmentation outputs, either as pre-processing, auxiliary, or core input. Two teams employed nnU-Net or U-Net variants for direct regression, while others used custom CNNs (1/7) or more complex architectures integrating attention mechanisms or hybrid designs (4/7). Prediction strategies varied across teams: two teams directly regressed biometry values; three teams predicted 3D landmark coordinates; and two teams generated 3D landmark heatmaps. In the latter two approaches, biometry values were subsequently computed using scripts provided by the organizers. Most teams used 3D models (6/7), implemented primarily in \textbf{PyTorch} (6/7), with one using TensorFlow. Three teams leveraged \textbf{external data}, such as dHCP and fetal brain atlases, or employed foundation models pre-trained on large-scale datasets.

\begin{landscape}
\begin{table}[t]
\centering
\vfill

\caption{Summary of the algorithms submitted for the fetal brain tissue segmentation task.}
\label{tab:segm_alg_descr}

\scriptsize
\resizebox{0.95\columnwidth}{!}{%
\begin{tabular}{m{2cm}m{2cm}m{2cm}m{0.5cm}m{5cm}m{2cm}m{2.5cm}m{3cm}m{5cm}}

\toprule
\textbf{Team name}       & \textbf{Model Architecture}                                   & \textbf{Deep Learning Framework}  & \textbf{Dim} & \textbf{Data Augmentation}                                                                                                                                                              & \textbf{Cross-Validation}  & \textbf{External Data}                                                                                         & \textbf{Ensembling}                                                & \textbf{Original Aspects}                                                                                                                                                                                                                                                                                                                                                                           \\ \midrule
\texttt{cemrg}           & Hybrid Cross Attention Swin Transformer and CNN      & PyTorch, nnUNet          & 3D             & Horizontal Flipping, Vertical flipping, scaling, normalization                                                                                                                 & 5-fold            & No                                                                                                    & No                                                        & The Cross Attention Transformer (CAT) block design.                                                                                                                                                                                                                                                                                                                                        \\
\texttt{CeSNE-DiGAIR}    & 3D UNet                                              & MONAI                    & 3D             & Deformable (SyN) registrations between couples of neurotypicall and pathological scans from the preprocessed training dataset. Skull-stripping with BOUNTI                     & Not specified     & No                                                                                                    & Use of models for post-processing                         & Denoising autoencoder for segmentation accuracy enhancement.                                                                                                                                                                                                                                                                                                                               \\
\texttt{falcons}         & 2D Attention Gated U-Net                             & TensorFlow               & 2D             & Rotation, width/height shift, vertical/horizontal flip, zooming, brightness, gaussian noise, gaussian blurring                                                                 & Not specified     & 70 images from dHCP)                                                                                  & Models with different architecture and (or) training data & Series of preprocessing steps including brain extraction, alignment, and non-uniform intensity correction. Ensembling of models trained on different orientations (axial, sagital, coronal)                                                                                                                                                                                                \\
\texttt{feta\_sigma}     & UxLSTMEnc, UNet                                      & nnUNet                   & 3D             & Rotation, Scaling, Translation, Gaussian Noise, Mirror Transform.                                                                                                              & 5-fold            & No                                                                                                    & Models with different architecture and (or) training data & Use of UxLSTM and ensembling with nnUnet, Background masking.                                                                                                                                                                                                                                                                                                                              \\
\texttt{hilab}           & nnU-Net                                              & PyTorch, nnUNet          & 3D             & Default nnU-Net augmentations, histogram equalization, differentiated probabilities for sample selection in random copy-paste augmentations, replication of challenging cases. & 5-fold            & No                                                                                                    & Models with different architecture and (or) training data & Applying histogram equalisation to 3D images, differentiated probabilities for sample selection in random copy-paste augmentations, strategically replicating challenging cases in the training data. Enesemble of 5 models with different hyperparameters and pre-processing settings.                                                                                                    \\
\texttt{jwcrad}          & Residual-USE-Net                                     & PyTorch, MONAI           & 3D             & Rotation, scaling, translation, intensity shift, low resolution simulation.                                                                                                    & 5-fold            & No                                                                                                    & Model trained on differnt CV splits                       & Custom auxiliary loss function based on transformation consistency.                                                                                                                                                                                                                                                                                                                        \\
\texttt{LIT}             & Attention UNet, nnUNet ResidualEncoderUNet           & PyTorch, nnUNet          & 3D             & Rotation, Scaling, Gaussian Noise, Gaussian Blur, Brightness Alteration, Contrast Adjustment, Low Resolution Simulation, Gamma Adjustment, Mirroring                           & 6-fold            & No                                                                                                    & Model trained on differnt CV splits                       & Custom brain mask detection with Attention Unet                                                                                                                                                                                                                                                                                                                                            \\
\texttt{lmrcmc}          & nnUNet, SegVol                                       & nnUNet, MONAI            & 3D             & nnUNet: default, SegVol: flip, ScaleIntensity, ShiftIntensity, GibbsNoise, BiasField, KSpaceSpikeNoise and Affine augmentation; with SLAug.                                    & Not specified     & No                                                                                                    & Models with different architecture and (or) training data & Ensemble of U-Net and a foundation model, use of the SegVol model in fetal brain segmentation.                                                                                                                                                                                                                                                                                             \\
\texttt{mic-dkfz}        & U-Net (nnUNet), U-Net with Residual encoder          & nnUNet                   & 3D             & randomized; blur, gaussian noise, spatial (rotation, scaling, flipping), brightness, contrast, low-resolution simulation, gamma, sharpening, blank rectangle                   & 5-fold            & Yes (pre-training in MultiTalent)                                                                     & Models with different architecture and (or) training data & \begin{tabular}[c]{@{}l@{}}Pretraining with MultiTalent on a \\collection of publicly available datasets.\\Ensemble of 3 nn-Unet configurations \\ with different data augmentations .\end{tabular} \\
\texttt{paramahir\_2023} & 3D UNet (segmentation), custom UNet-based (biometry) & MONAI                    & 3D             & Random Flipping, Random Rotation, Random Intensity Shifts                                                                                                                      & Not specified     & No                                                                                                    & No                                                        & Combination of segmentation and biometry prediction in a unified pipeline.                                                                                                                                                                                                                                                                                                                 \\
\texttt{pasteurdbc}      & MedNeXt\_L and nnUNet                                & nnUNet                   & 3D             & RandomScaling, RandomRoatation, RandomAdjustContrast, RandFlip                                                                                                                 & 5-fold            & Multi-modal multi-organ medical image datasets used in the pre-trained MedNeXt\_L foundational model & Models with different architecture and (or) training data & Used additional datasets with CT and brain MRI images for model pre-training                                                                                                                                                                                                                                                                                                               \\
\texttt{qd\_neuroincyte} & Swin UNETR                                           & MONAI                    & 3D             & Random sliding window, flipping, 1\% gaussian noise, rigid rotation of ± 25° around all axes, random shifting ± 5 mm along all axes.                                           & Not specified     & No                                                                                                    & Use of models for post-processing                         & Brain masking for vienna                                                                                                                                                                                                                                                                                                                                                                   \\
\texttt{unipd-sum-aug}   & 2D Swin-UMamba                                       & PyTorch, nnUNetv2, Monai & 2D             & TorchIO transforms and GIN techniques, pair-wise co-registration, affine and rigid transforms using the Advanced Nomalization Tools.                                           & 5-fold            & Model pre-trained on ImageNet                                                                         & Model trained on differnt CV splits                       & Pretrained on imageNet repository and using GIN.                                                                                                                                                                                                                                                                                                                                           \\
\texttt{UPFetal24}       & nnU-Net ResEncL                                      & nnUNetv2                 & 3D             & Default nnU-Net augmentations; differentiated by specific data augmentations for each of the three models.                                                                     & 5-fold for config & dHCP and fetal atlasses                                                                               & Models with different architecture and (or) training data & Data augmentation strategies and ensembling of models                                                                                                                                                                                                                                                                                                                                      \\
\texttt{ViCOROB}         & nnUNet                                               & nnUNet, PyTorch          & 3D             & Random bias field, motion artifacts, low-resolution simulation, SynthSeg-inspired T2w image synthesizer                                                                        & 3-fold            & No                                                                                                    & Model trained on differnt CV splits                       & SynthSeg-inspired T2w image synthesizer, Sharpness-Aware Minimization (SAM) optimizer                                                                                                                                                                                                                                                                                                      \\ \bottomrule
\end{tabular}
}
\end{table}
\end{landscape}

\begin{strip}
\centering
\captionof{table}{Summary of the algorithms submitted for the biometry estimation task.}
\label{tab:biom_agl_descr}
\resizebox{\linewidth}{!}{%
\begin{tabular}{lllm{7cm}m{5cm}m{5cm}m{3cm}m{3.5cm}m{2cm}}
\toprule
\textbf{Team name}                & \textbf{Architecture}          & \textbf{Dimensionality} & \textbf{Original Aspects}                                                                                                 & \textbf{External datasets}                                                                                       & \textbf{Framework/languange}                                                                                                        \\ \midrule
\texttt{qd\_neuroincyte} & SwinUnetr             & 3D             & Relies on segmentation. Predict landmark heat maps only using the segmentation maps and then calculate biometry. & No additional data was used                                                                             & Pytorch 2.2.2                                                                                                              \\
\texttt{CeSNE-DiGAIR}             & CNN                   & 3D             & Relies on segmentation. Predict the keypoints given the segmentation.                                            & No additional data was used                                                                             & PyTorch Version 2.4.0                                                                                                      \\
\texttt{jwcrad}                   & Residual-USE-Net      & 3D             & Relies on segmentation. Uses the segmentation maps to localize and preprocess the input images by masking and cropping the original 3D image. Predict landmark heat maps using the preprocessed images and then calculate biometry. & No additional data was used                                                                             & PyTorch 2.2.2                                                                                                              \\
\texttt{pasteurdbc}               & MedNeXt\_L nnUNet     & 3D             & Use of a pre-trained foundational model.                                                                                                                & Yes (for the pre-trained MedNeXt\_L foundational model, multi-modal multi-organ medical image datasets) &                                                                                                                            \\
\texttt{falcons}                  & Attention Gated U-Net & 2D             & Relies on segmentation. Predict the biometry values directly                                                     & Yes (+70 images from dHCP)                                                                              & Tensorflow(2.10.0) FMRIB Software Library(FSL 6.0), CIVET(2.1.0), Advanced Normalization Tools(ANTs), Scikit-learn (1.5.1) \\

\texttt{feta\_sigma}          & nnUNet, UxLSTMEnc                  & 3D             & Ensemble network of nnUnet and UxLSTMEnc.                               & No additional data was used                                                                             & PyTorch                                                                          \\
\texttt{paramahir\_2023}          & UNet                  & 3D             & Relies on segmentation. Predict the biometry values by regressing the U-Net features.                               & No additional data was used                                                                             & PyTorch 2.3 -                                                                          \\\bottomrule

\end{tabular}%
}

\vspace{1cm}
\end{strip}

\subsection{FeTA 2024 results}
\label{sec:f24results}

\subsubsection{Brain tissue segmentation ranking}
\paragraph{Segmentation performance overview}
Figure~\ref{fig:4_segm_sites_metrics} highlights performance across sites and metrics, revealing a general \textbf{performance plateau} among top methods. For most teams, average Dice scores stabilized around \textbf{0.8-0.82}, HD95 around \textbf{2.8-2.1}, and VS around \textbf{0.9-0.92}, while the ED showed wider variability (ranging from \textbf{20} to \textbf{40}), highlighting its sensitivity to topological inaccuracies.

\begin{figure*}[ht]
    \centering
    \includegraphics[width=1\linewidth]{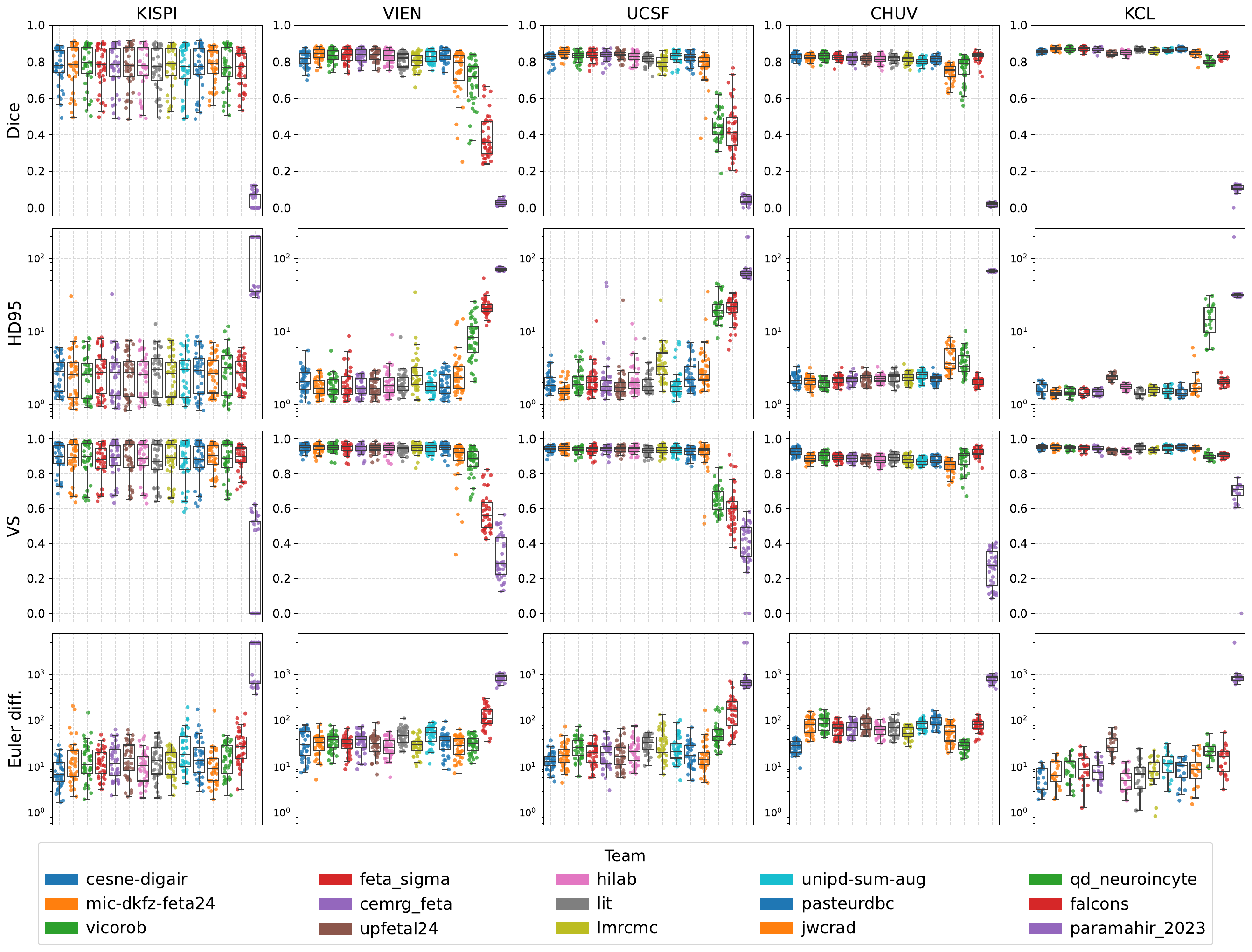}
    \caption{Segmentation performance by site and evaluation metric. In each subplot, teams are ranked from left to right based on their average performance across all labels for the given metric (best to worst). Team colors are consistent across plots and correspond to the legend.}
    \label{fig:4_segm_sites_metrics}
\end{figure*}
\paragraph{Site-specific trends}
Despite being introduced in this edition as a new low-field, out-of-domain dataset, KCL showed the best segmentation performance. In contrast, KISPI yielded the lowest performance, even though it was part of the previous editions’ training and testing data. Across metrics, UCSF and KISPI displayed higher interquartile ranges, particularly for Dice, HD95, and VS, reflecting greater variability across methods. Some teams (\texttt{falcons}, \texttt{qd\_neuroincyte}) experienced performance drops on sites that use NiftyMIC SRR, like UCSF or VIEN, with Dice scores dropping to 0.38–0.44 compared to 0.76–0.83 on other sites.

\paragraph{Label-specific trends}
SGM, GM, and BS were consistently the most challenging labels to segment across all teams, as shown by lower performance metrics in Supplementary Materials A4. Among the top three models, Dice scores dropped from an average (across all labels) of 0.82 to 0.80 for SGM, 0.79 for BS, and 0.74 for GM. HD95 increased from 2.24 to 3.6 for BS and 3.0 for SGM, while VS declined from 0.92 to 0.86 for SGM and 0.88 for BS. GM also showed a marked increase in ED, from 33.14 to 137, reflecting a significant loss in topological accuracy. 


\paragraph{Ranking summary}
Table~\ref{tab:ranking_segm} presents the aggregated average metrics and rankings per team \footnote{The rankings originally published on the website and announced during the MICCAI challenge differ from those presented in this paper due to a change in the ED estimation method. Specifically, we updated the way the ground-truth Euler characteristic is determined. In the original rankings, it was computed based on the manual segmentations. In the current results, it is calculated from manually defined topological properties (see \ref{par:euler}). Manual segmentations were created via interpolation and because many structures were not segmented on every slice, the resulting ground-truth segmentations contained numerous topological errors (e.g., holes, disconnected components). As a result, they did not reliably represent the expected topological properties of the anatomical structures. To address this, we now use manually specified topological values to calculate the ground-truth EC.}. Qualitative examples of the segmentations are provided in Supplementary Appendix A9. Notable rank discrepancies across metrics highlight their complementary nature. Figure~\ref{fig:segm_ranking} provides a more granular view, showing single-metric rankings across different sites and tissue labels. Dice score rankings remained relatively consistent across submissions and anatomical regions, while ED rankings showed greater variability, both across tissues and sites, reinforcing the importance of using multiple metrics to capture distinct aspects of segmentation quality.

Figure~\ref{fig:topex2} further illustrates the added value of topological metrics. In a comparative example, \texttt{mic-dkfz-feta24} achieves similar Dice and lower HD95 scores but poorer ED and VS, suggesting that voxel-level agreement alone may not suffice for tasks requiring topologically accurate surfaces, such as morphological analysis.

\vfill\null
\newpage
\paragraph{Per-tissue and condition analysis}
Extended performance results split by site, tissue label, and pathology status are available in supplementary materials (sections A3,  A4, and A5, respectively).

\begin{table*}
\centering

\includegraphics[width=.9\linewidth]{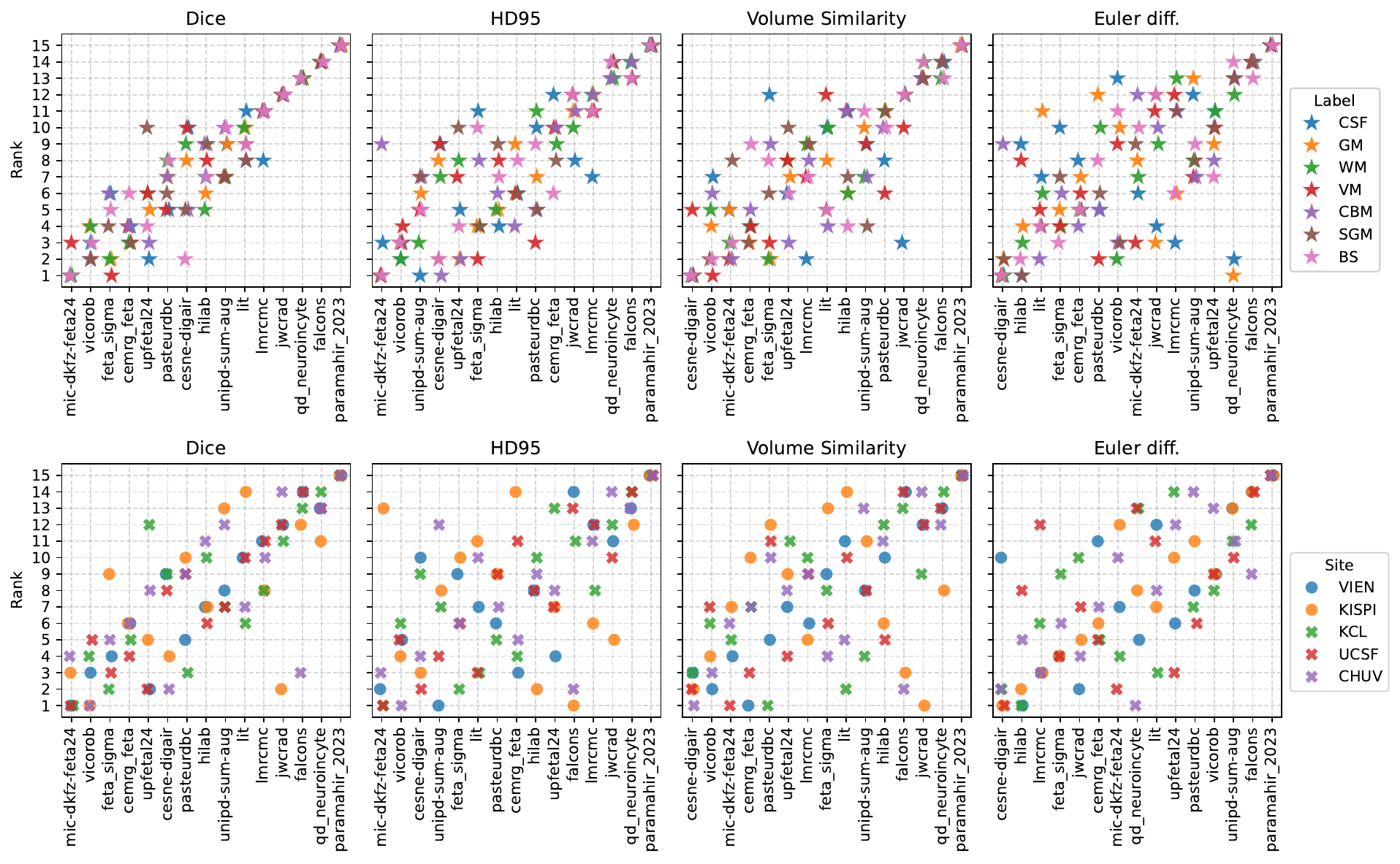}
\vspace{-.2cm}
\captionof{figure}{Detailed submissions rankings. Top: Across labels. Bottom: Across sites (circles indicated in-domain splits and crosses out-of-domain splits). Teams are sorted in each subplot by their overall ranking in the segmentation task, from the best to the worst.}
\label{fig:segm_ranking}
\vspace{.2cm}
\caption{Segmentation ranking and average metrics}
\label{tab:ranking_segm}
\resizebox{.8\linewidth}{!}{%
\begin{tabular}{lcccccccccc}
\toprule
\multirow{2}{*}{\textbf{Team}} & \multicolumn{2}{c}{\textbf{Dice}} & \multicolumn{2}{c}{\textbf{HD95}} & \multicolumn{2}{c}{\textbf{VS}} & \multicolumn{2}{c}{\textbf{ED}} & \multirow{2}{1.2cm}{\centering\textbf{Mean rank}} & \multirow{2}{1.2cm}{\centering \textbf{Final rank}} \\
\cmidrule(rl){2-3} \cmidrule(rl){4-5}\cmidrule(rl){6-7}\cmidrule(rl){8-9}
& Rank & Value & Rank & Value& Rank & Value& Rank & Value &  &  \\
\midrule
\texttt{cesne-digair}    & 8                              & 0.816         & 3                                & 2.317           & 1                            & 0.929       & 1                                     & 20.921                           & 3.25                          & \textbf{1}                           \\
\texttt{mic-dkfz-feta24} & 1                              & 0.828         & 2                                & 2.224           & 3                            & 0.918       & 8                                     & 37.206                           & 3.50                          & \textbf{2}                           \\
\texttt{vicorob}         & 2                              & 0.825         & 1                                & 2.187           & 2                            & 0.920       & 11                                    & 41.293                           & 4.00                          & \textbf{3}                       \\
\texttt{feta\_sigma}     & 3                              & 0.822         & 7                                & 2.430           & 5                            & 0.914       & 4                                     & 31.710                           & 4.75                          & \textbf{4}                           \\
\texttt{cemrg\_feta}     & 4                              & 0.822         & 10                               & 2.836           & 4                            & 0.916       & 7                                     & 34.382                           & 6.25                          & \textbf{5}                           \\
\texttt{upfetal24}       & 5                              & 0.820         & 6                                & 2.412           & 6                            & 0.913       & 9                                     & 39.967                           & 6.50                          & \textbf{6}                           \\
\texttt{hilab}           & 7                              & 0.816         & 8                                & 2.434           & 9                            & 0.911       & 3                                     & 30.123                           & 6.75                          & \textbf{7}      \\
\texttt{lit}             & 10                             & 0.808         & 5                                & 2.391           & 8                            & 0.911       & 10                                    & 40.085                           & 8.25                          & \textbf{8}                         \\
\texttt{lmrcmc}          & 11                             & 0.805         & 11                               & 3.179           & 7                            & 0.913       & 5                                     & 32.751                           & 8.50                          & \textbf{9}                       \\
\texttt{unipd-sum-aug}   & 9                              & 0.811         & 4                                & 2.332           & 10                           & 0.909       & 13                                    & 46.668                           & 9.00                          & \textbf{10}                       \\
\texttt{pasteurdbc}      & 6                              & 0.817         & 9                                & 2.474           & 11                           & 0.909       & 12                                    & 41.521                           & 9.50                          & \textbf{11}                      \\
\texttt{jwcrad}          & 12                             & 0.769         & 12                               & 3.569           & 12                           & 0.886       & 2                                     & 29.744                           & 9.50                          & \textbf{11}                      \\
\texttt{qd\_neuroincyte} & 13                             & 0.681         & 13                               & 10.441          & 13                           & 0.827       & 6                                     & 34.295                           & 11.25                         & \textbf{13}                      \\
\texttt{falcons}         & 14                             & 0.628         & 14                               & 11.040          & 14                           & 0.765       & 14                                    & 100.729                          & 14.00                         & \textbf{14}                      \\
\texttt{paramahir\_2023} & 15                             & 0.040         & 15                               & 80.757          & 15                           & 0.337       & 15                                    & 1416.515                         & 15.00                         & \textbf{15}                \\
\bottomrule
\end{tabular}}
\end{table*}

\begin{figure*}[ht]
\centering
    \vspace{-.5cm}
    \includegraphics[width=.7\linewidth]{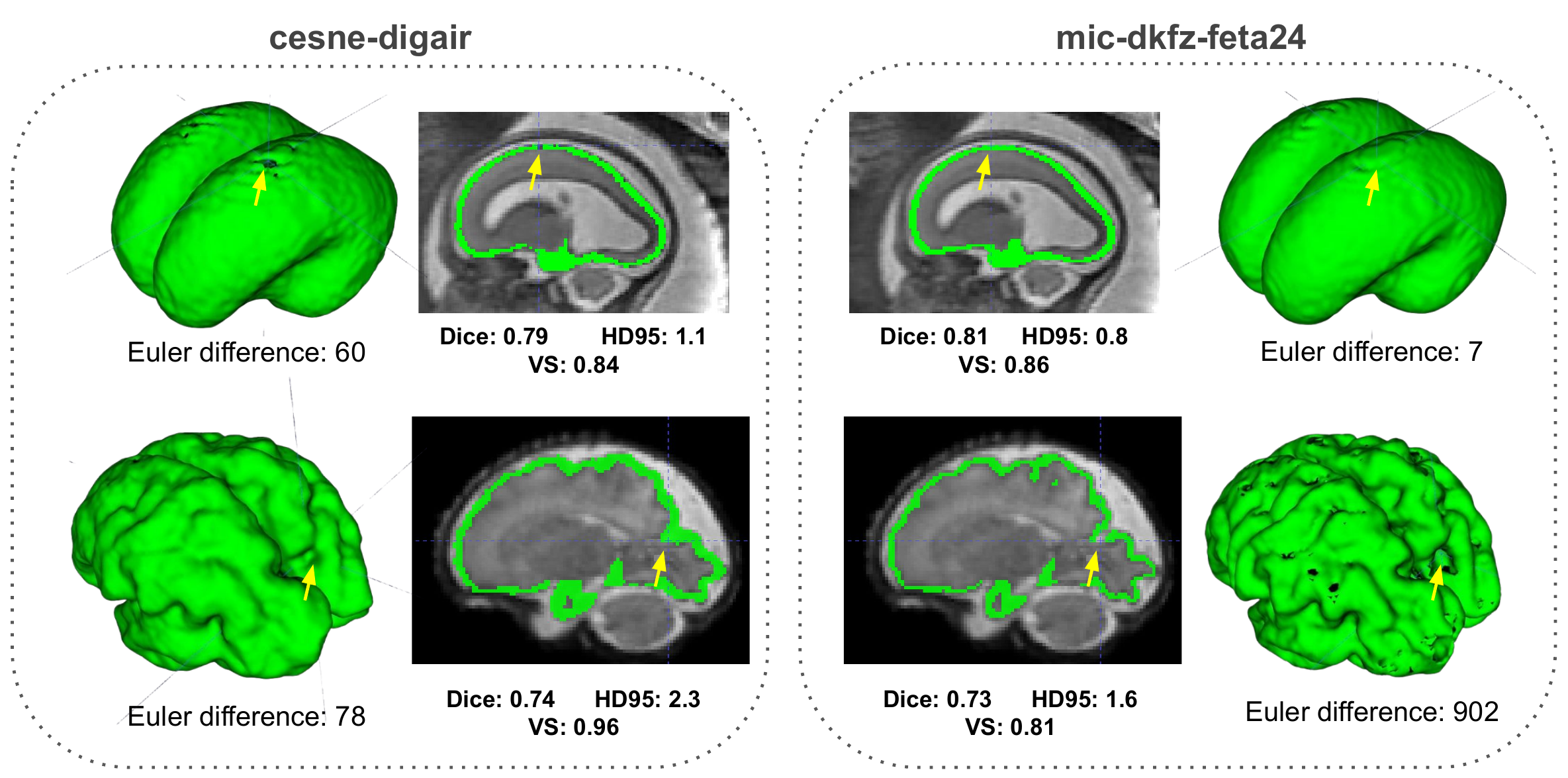}
    
    \captionof{figure}{Segmentation results and reconstructed cortical GM surfaces for two representative fetal cases from \texttt{cesne-digair} and \texttt{mic-dkfz-feta24}, visualized using ITK-SNAP \citep{py06nimg}. The top row (fetus at 22 weeks GA) illustrates that higher Dice scores sometimes correspond to smaller topological errors. However, the bottom row (fetus at 30 weeks GA) demonstrates significant topological issues (e.g., holes, fragmentation) in the \texttt{mic-dkfz-feta24} surface, despite comparable Dice and HD95 values. This underscores the need for additional topological and structural metrics, such as ED, to comprehensively evaluate segmentation quality, as metrics like Dice or HD95 alone are insufficient to capture topological accuracy.}\label{fig:topex2}
    \vspace{1cm}
\end{figure*}

\vfill\null
\newpage

\subsubsection{Biometry ranking}
\paragraph{Performance across sites and measurement} 
Figure~\ref{fig:biom_boxplot} summarizes model performance per site and biometric measurement, with detailed values available in the supplementary materials section A6. 

VIEN was the most challenging site, where no method outperformed the \textbf{[GA]} baseline (MAPE: 0.106$\pm$0.112), including the best-performing teams \texttt{cesne-digair} and \texttt{jwcrad}, which reached similar error levels. In contrast, KISPI emerged as the least challenging, with all three top teams exceeding the baseline. Across KCL, UCSF, and CHUV, only two teams per site (out of the top 3: \texttt{jwcrad}, \texttt{feta\_sigma}, \texttt{cesne\_digair}) achieved better-than-baseline performance. Measurement-wise, LCC, HV, and TCD were consistently more difficult, with HV and LCC showing the highest MAPE across all teams and raters. In contrast, sBIP and bBIP were among the best estimated. Notably, only \texttt{jwcrad} surpassed the baseline across all measurements, while a few others, including \texttt{feta\_sigma} and \texttt{pasteurdbc}, did so on selected metrics.
\vfill\null
\newpage

Although multiple teams performed comparably on individual metrics, the clear winner in the ranking (see Table~\ref{tab:biom_ranks}) was \texttt{jwcrad}, demonstrating consistent superiority across both site and measurement variations. 

\paragraph{Robustness in pathological vs. neurotypical condition} To assess model generalizability, we compared biometry performance between neurotypical and pathological brains (see Figure \ref{fig:biom_path}). 
While most measurements did not reveal statistically significant differences between groups, bBIP showed better accuracy in the healthy cohort, particularly at VIEN. Conversely, UCSF results suggested slightly better performance for pathological subjects.

In summary, the best-performing method, \texttt{jwcrad}, came within 9\% of expert agreement for some measurements (e.g., TCD). However, for others like bBIP, its results differed from the expert range by as much as 60\%. This reveals important gaps where automated biometry methods still fall short, especially in pathological cases.

\vfill\null
\newpage

\begin{strip}
\centering
\captionof{table}{Metrics and ranking for the biometry estimation task sorted by the final MAPE. [GA] and [inter-rater] entries do not represent participating models, thus their rank is marked as \textit{*}}
\label{tab:biom_ranks}
\resizebox{.9\linewidth}{!}{
\begin{tabular}{lcccccccccccc}
\toprule
\textbf{Team} 
& \multicolumn{2}{c}{\textbf{LCC}} 
& \multicolumn{2}{c}{\textbf{HV}} 
& \multicolumn{2}{c}{\textbf{bBIP}} 
& \multicolumn{2}{c}{\textbf{sBIP}} 
& \multicolumn{2}{c}{\textbf{TCD}} 
& \multirow{2}{1.2cm}{\centering \textbf{Final MAPE}} & \multirow{2}{1.2cm}{\centering \textbf{Final rank}}\\
\cmidrule(lr){2-3} \cmidrule(lr){4-5} 
\cmidrule(lr){6-7} \cmidrule(lr){8-9} 
\cmidrule(lr){10-11} 
 & \textbf{MAPE} & \textbf{Rank}
 & \textbf{MAPE} & \textbf{Rank}
 & \textbf{MAPE} & \textbf{Rank}
 & \textbf{MAPE} & \textbf{Rank}
 & \textbf{MAPE} & \textbf{Rank}
 & & \\
\midrule
\texttt{\textit{[inter-rater]}} & \textit{9.59} & \textit{*} & \textit{8.04} & \textit{*} & \textit{3.28} & \textit{*} & \textit{1.49} & \textit{*} & \textit{4.89} & \textit{*} & \textit{5.38} & \textit{*} \\
\texttt{jwcrad} & \textbf{11.15} & \textbf{1} & 10.32 & 2 & 5.43 & 2 & 4.78 & 3 & 7.21 & 2 & \textbf{7.72} & \textbf{1} \\
\texttt{\textit{[GA]}} & \textit{12.75} & \textit{3} & \textit{11.26} & \textit{3} & \textit{6.82} & \textit{5} & \textit{6.47} & \textit{5} & \textit{10.80} & \textit{3} & \textit{9.56} & \textit{*} \\
\texttt{cesne-digair} & 17.72 & 4 & \textbf{9.82} & \textbf{1} & \textbf{4.02} & \textbf{1} & 4.74 & 2 & 12.34 & 4 & 9.59 & 2 \\
\texttt{feta\_sigma} & 12.59 & 2 & 11.55 & 4 & 5.74 & 3 & 5.54 & 4 & 13.66 & 5 & 9.76 & 3 \\
\texttt{pasteurdbc} & 20.47 & 5 & 43.48 & 7 & 6.51 & 4 & \textbf{3.74} & \textbf{1} & \textbf{5.43} & \textbf{1} & 15.83 & 4 \\
\texttt{paramahir\_2023} & 28.48 & 6 & 29.35 & 5 & 26.13 & 7 & 25.46 & 6 & 30.78 & 6 & 28.03 & 5 \\
\texttt{falcons} & 34.88 & 8 & 46.25 & 8 & 24.62 & 6 & 28.13 & 7 & 36.72 & 7 & 34.09 & 6 \\
\texttt{qd\_neuroincyte} & 32.78 & 7 & 42.84 & 6 & 38.41 & 8 & 37.83 & 8 & 47.92 & 8 & 40.07 & 7 \\
\bottomrule
\end{tabular}}

    \includegraphics[width=.8\linewidth]{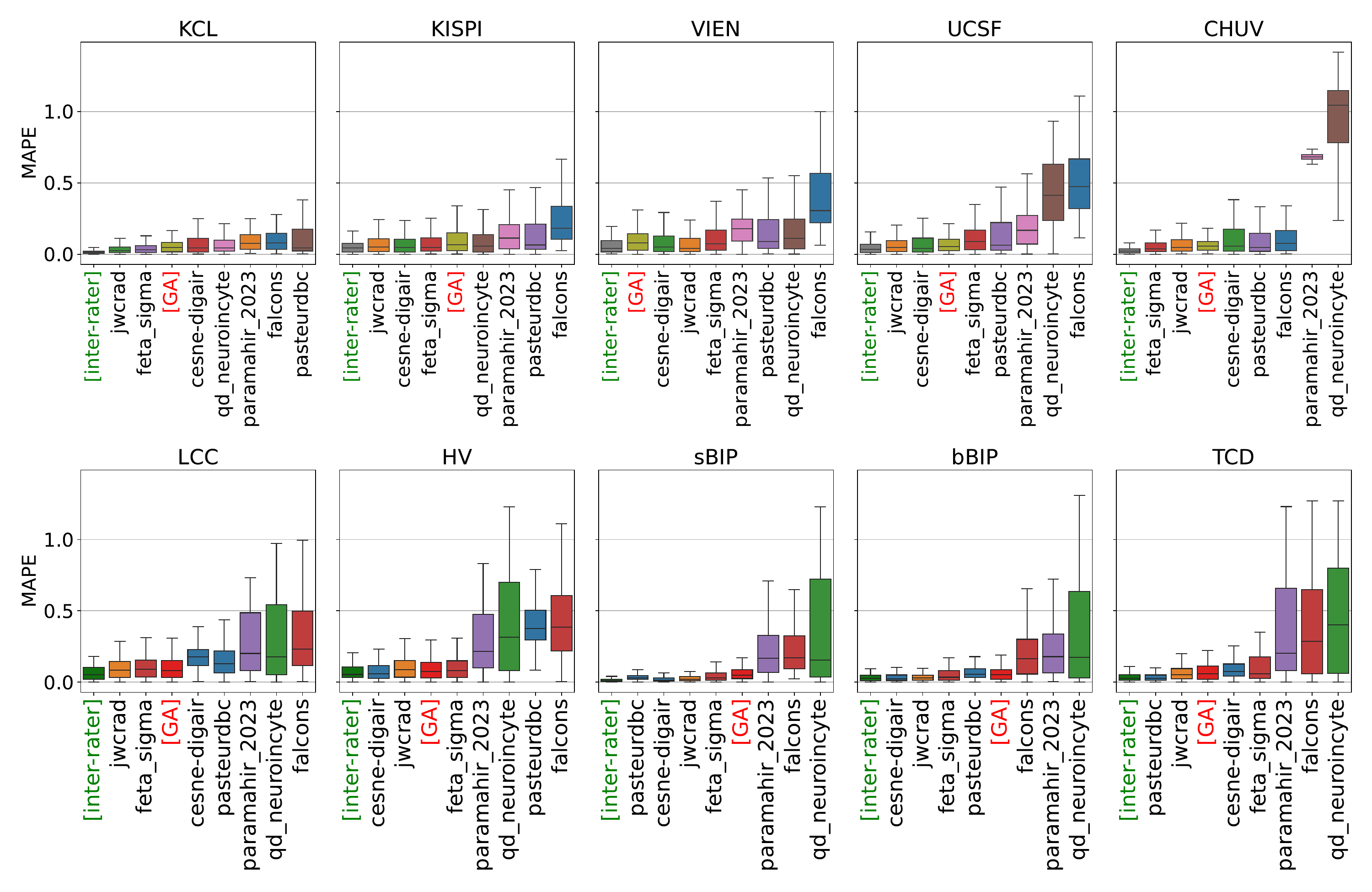}
    \vspace{-.3cm}
    \captionof{figure}{Biometry results per site (top) and label (bottom) for all teams participating in the Task 2 together with the GA baseline model ([GA]) and the inter-rater variability [inter-rater]. Teams are sorted in ascending order for each subplot independently, based on their mean MAPE for a given site or label.}
    
    \label{fig:biom_boxplot}

    \includegraphics[width=.8\linewidth]{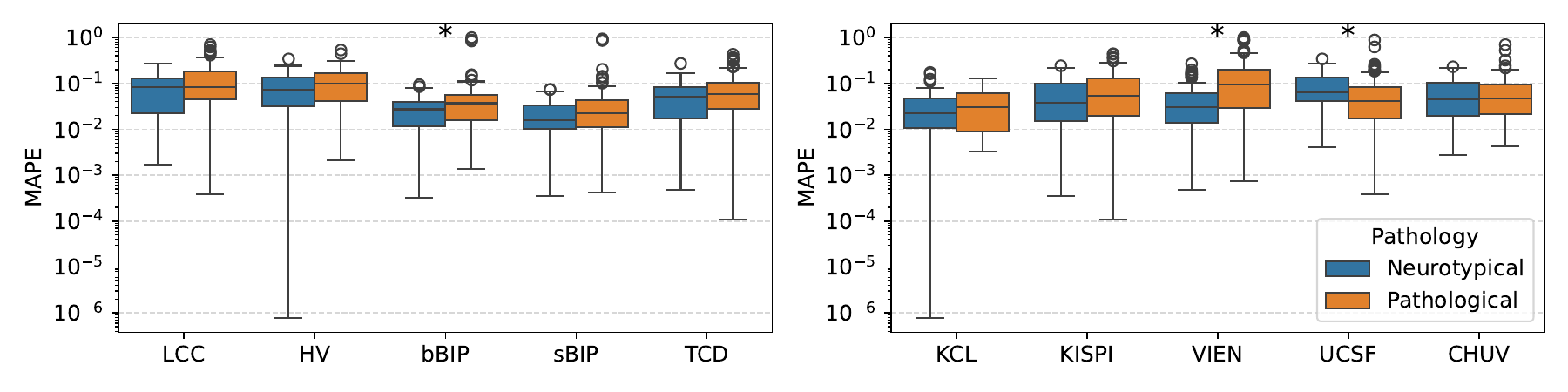}
    \vspace{-.3cm}
    \captionof{figure}{Biometry results for healthy and pathological subjects across labels and sites for the winning team \texttt{jwcrad}. Asterisks above the boxplot indicate statistically significant differences between the two groups (p $< $0.05, Mann-Whitney test).}
    \label{fig:biom_path}
    \vspace{1cm}
\end{strip}

\begin{strip}
\centering
\captionof{table}{Mean and standard deviation (mean$\pm$std) for different metrics across years and splits over all labels. Bold and * highlight the years that have statistically significant improvement in the values compared to the previous year. ED was not estimated in FeTA 2021}
\label{tab:long_years}
\resizebox{0.6\linewidth}{!}{
\begin{tabular}{llcccc}
\toprule
\textbf{Year} &  \textbf{Site} & \textbf{Dice}  & \textbf{HD95} & \textbf{VS} & \textbf{ED} \\
\midrule
2021 & KISPI & 0.79$\pm$\scriptsize{0.16} & 2.81$\pm$\scriptsize{3.43} & 0.89$\pm$\scriptsize{0.16} & not estimated\\

\midrule
\multirow[c]{5}{*}{2022} & CHUV & 0.81$\pm$\scriptsize{0.09} & 2.33$\pm$\scriptsize{1.68} & 0.88$\pm$\scriptsize{0.10} & 77.43$\pm$\scriptsize{168.91} \\
 & KCL & 0.87$\pm$\scriptsize{0.05} & 1.46$\pm$\scriptsize{0.57} & 0.95$\pm$\scriptsize{0.05} & 28.61$\pm$\scriptsize{53.89} \\
 & KISPI & 0.77$\pm$\scriptsize{0.18} & 3.17$\pm$\scriptsize{4.16} & 0.87$\pm$\scriptsize{0.18} & 18.98$\pm$\scriptsize{54.56} \\
 & UCSF & 0.84$\pm$\scriptsize{0.06} & 2.02$\pm$\scriptsize{1.44} & 0.95$\pm$\scriptsize{0.05} & 18.73$\pm$\scriptsize{44.52} \\
 & VIEN & 0.84$\pm$\scriptsize{0.08} & 1.87$\pm$\scriptsize{1.46} & 0.95$\pm$\scriptsize{0.06} & 32.20$\pm$\scriptsize{67.05} \\
\midrule
\multirow[c]{5}{*}{2024} & CHUV & 0.83$\pm$\scriptsize{0.06} & 2.23$\pm$\scriptsize{1.40} & \textbf{0.93$\pm$\scriptsize{0.06}*} & \textbf{29.10$\pm$\scriptsize{51.56}*}\\
 & KCL & 0.86$\pm$\scriptsize{0.05} & 1.69$\pm$\scriptsize{0.52} & 0.95$\pm$\scriptsize{0.04} & \textbf{6.26$\pm$\scriptsize{13.21}*} \\
 & KISPI & 0.78$\pm$\scriptsize{0.15} & 2.95$\pm$\scriptsize{2.86}& 0.89$\pm$\scriptsize{0.14} & 9.21$\pm$\scriptsize{17.84} \\
 & UCSF & 0.82$\pm$\scriptsize{0.07} & 2.13$\pm$\scriptsize{1.31} & 0.94$\pm$\scriptsize{0.05} & 14.57$\pm$\scriptsize{25.90} \\
 & VIEN & 0.81$\pm$\scriptsize{0.09} & 2.27$\pm$\scriptsize{1.69} & 0.95$\pm$\scriptsize{0.05} & 38.13$\pm$\scriptsize{93.48} \\
 \bottomrule
\end{tabular}

}
\end{strip}

\subsection{Segmentation performance across challenge editions (2021, 2022 and 2024)}
Over the three editions of the FeTA challenge, the segmentation task has expanded both in terms of dataset size (from 40 to 180 test cases) and site diversity (from 1 to 5 imaging centers). To evaluate progress over time, we compared segmentation performance across the years 2021, 2022, and 2024, focusing on common testing sites. Table~\ref{tab:long_years} summarizes aggregated metrics for the top-performing teams each year (\texttt{cesne-digair} for 2024,  \texttt{FIT\_1} for 2022 and \texttt{NVAUTO} for 2021), and Figure~\ref{fig:kiviat_long} provides a visual overview of mean scores per label and site across years, with markers for statistically significant differences.

\begin{figure*}[t]
    \centering
    \vspace{-.3cm}
    \includegraphics[width=.9\linewidth]{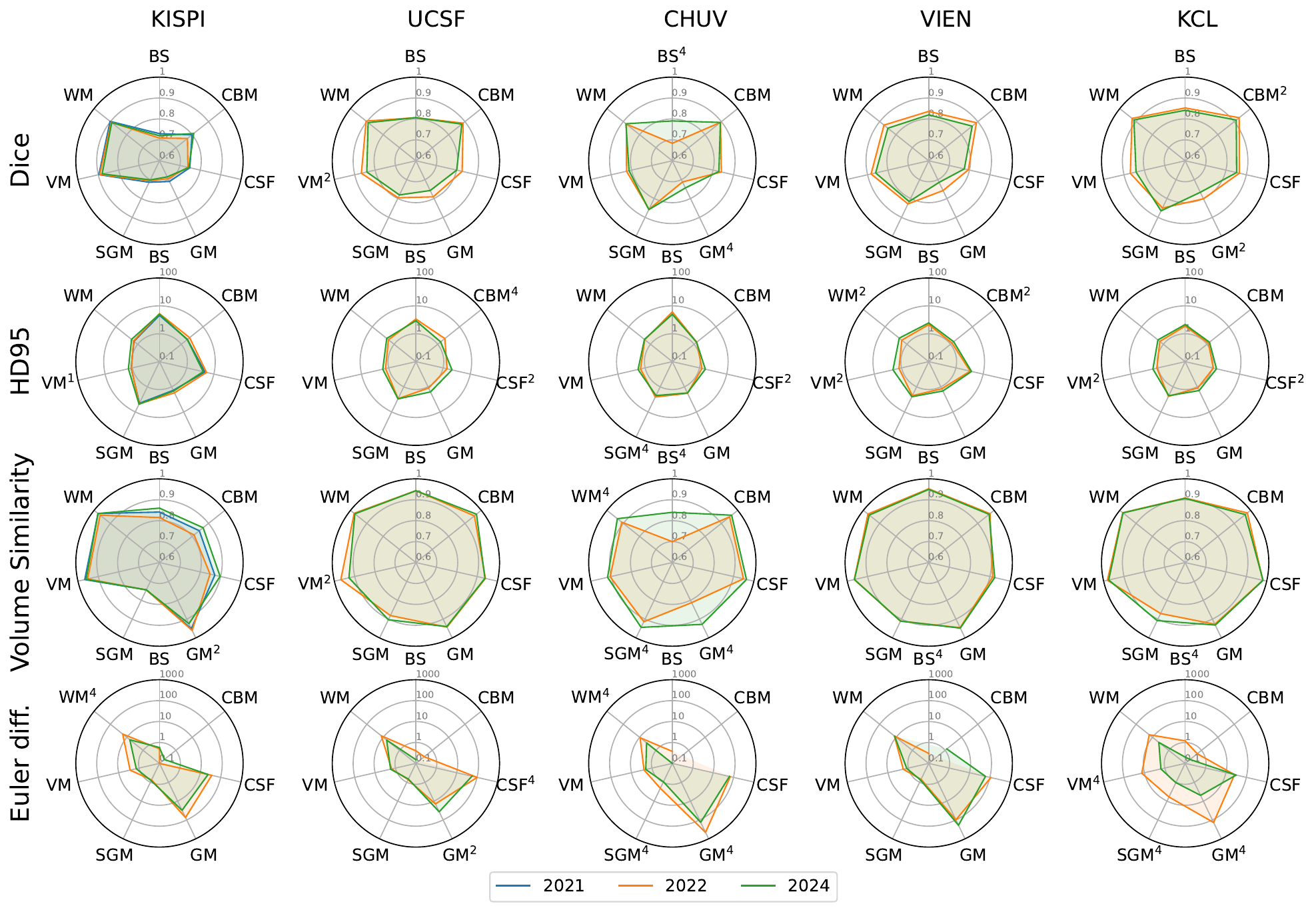}
    \captionof{figure}{ Segmentation performance improvement over the three editions of the FeTA Challenge. The superscript number above each label indicates whether the performance for a particular year and site-label-metric combination was statistically significantly better compared to all other available years. A superscript "2" indicates that the results for 2022 were the best, while a superscript "4" indicates that the results for 2024 were the best.}
    \vspace{-.3cm}
    \label{fig:kiviat_long}
\end{figure*}

\noindent\textit{KISPI split (2021–2024).} This is the only site included in all three editions. No statistically significant improvement over the years was observed across the tracked metrics: Dice (0.79$\pm$0.16 $\rightarrow$ 0.77$\pm$0.18 $\rightarrow$ 0.78$\pm$0.15), HD95 (2.81$\pm$3.43 $\rightarrow$ 3.17$\pm$4.16 $\rightarrow$ 2.95$\pm$2.86), and VS (0.89$\pm$0.16 $\rightarrow$ 0.87$\pm$0.18 $\rightarrow$ 0.89$\pm$0.14). The only statistically significant change occurred in the VS metric for the GM label between 2021 and 2022 (0.96 $\rightarrow$ 0.94), but no consistent improvement was found in subsequent years or for other labels.

\noindent\textit{Other sites (2022–2024).} For sites such as CHUV, UCSF, and VIEN—which were included in both 2022 and 2024—no consistent improvement was observed across metrics or tissue labels. While some metrics showed statistically significant changes, these were isolated and not consistent across sites, making it difficult to interpret them as evidence of overall progress in segmentation performance. Notably, ED improved substantially for CHUV and KCL. However, this improvement may be partially influenced by the inclusion of ED in the 2024 ranking, which favored algorithms with better topological performance. At CHUV, both VS and ED were significantly better than in 2022, but the other two metrics did not show similar trends.

Overall, although methods have become more sophisticated and the data more diverse, performance has not consistently improved across editions.

\subsection{Domain shifts evaluation}

\subsubsection{Impact of image quality on performance}
The impact of image quality on model performance as determined by computing the conditional mean across quality ratings ($\mathbb{E}[f(x) | \text{Quality}]$) is shown in the right-most column in Figure~\ref{fig:cond_means}. We see a clear effect of image quality on Dice, with a generally increasing Dice with the increasing image quality, amounting to a change from 0.75 Dice on average for the lowest quality data (with scores close to 1) and an average quality close to 0.85 for the highest quality data. Results using HD95 and VS generally align with the ones from Dice, except for GA and quality. The relationship is, however, not as clear for ED, although best quality images tend to yield the smallest ED. 

A more detailed analysis of the correlation between quality and the difference scores in the supplementary materials A7 showed a generally high Pearson correlation between quality and Dice ($r$ = 0.5-0.7) for all sites except KISPI-mial ($r$=0.4) and USCF-nmic ($r$=0.06 -- no correlation). The same trends, although weaker, were observed for HD95 and VS, except CHUV-mial and HD95, which had virtually no correlation ($r$=0.05). Results for ED showed no clear pattern, and larger correlations ($r$=0.3-0.4) were not statistically significant.

\subsubsection{Relative contribution of domain-shift sources}

Figure~\ref{fig:cond_means} displays the conditional means across different factors. The analysis revealed a pronounced site-SR effect: for example, the KISPI-mial site produced notably lower Dice scores, whereas the CHUV-mial site was associated with higher ED values. In addition, gestational age (GA) significantly affected both Dice and ED scores. Similar trends can be found in the supplementary materials (appendix A7), although HD95 scores appear to be less influenced by GA.

\vfill\null
\newpage

Figure~\ref{fig:shap} presents a SHAP analysis for all metrics using only image-level descriptors—namely, image quality, subject condition, and gestational age. (We excluded Site-SR from this analysis because its dependence on the other variables could lead to misleading SHAP values under the assumption of feature independence~\citep{mase2019explaining}.) Overall, the SHAP analysis summarizes how these factors influence the Dice and ED scores: image quality generally has the largest impact, followed by GA (except for HD95). Although the pathological status of a subject generally has a small effect, we observed that severely pathological cases often have lower GA, which might introduce confounding. The plot’s color coding further confirms that higher image quality and GA are associated with increased Dice scores—for example, poor quality data may result in about –0.1 Dice, compared to an average of +0.05 Dice for good quality data.

\vfill\null
\newpage
\begin{strip}
    \centering
    \includegraphics[width=1\linewidth]{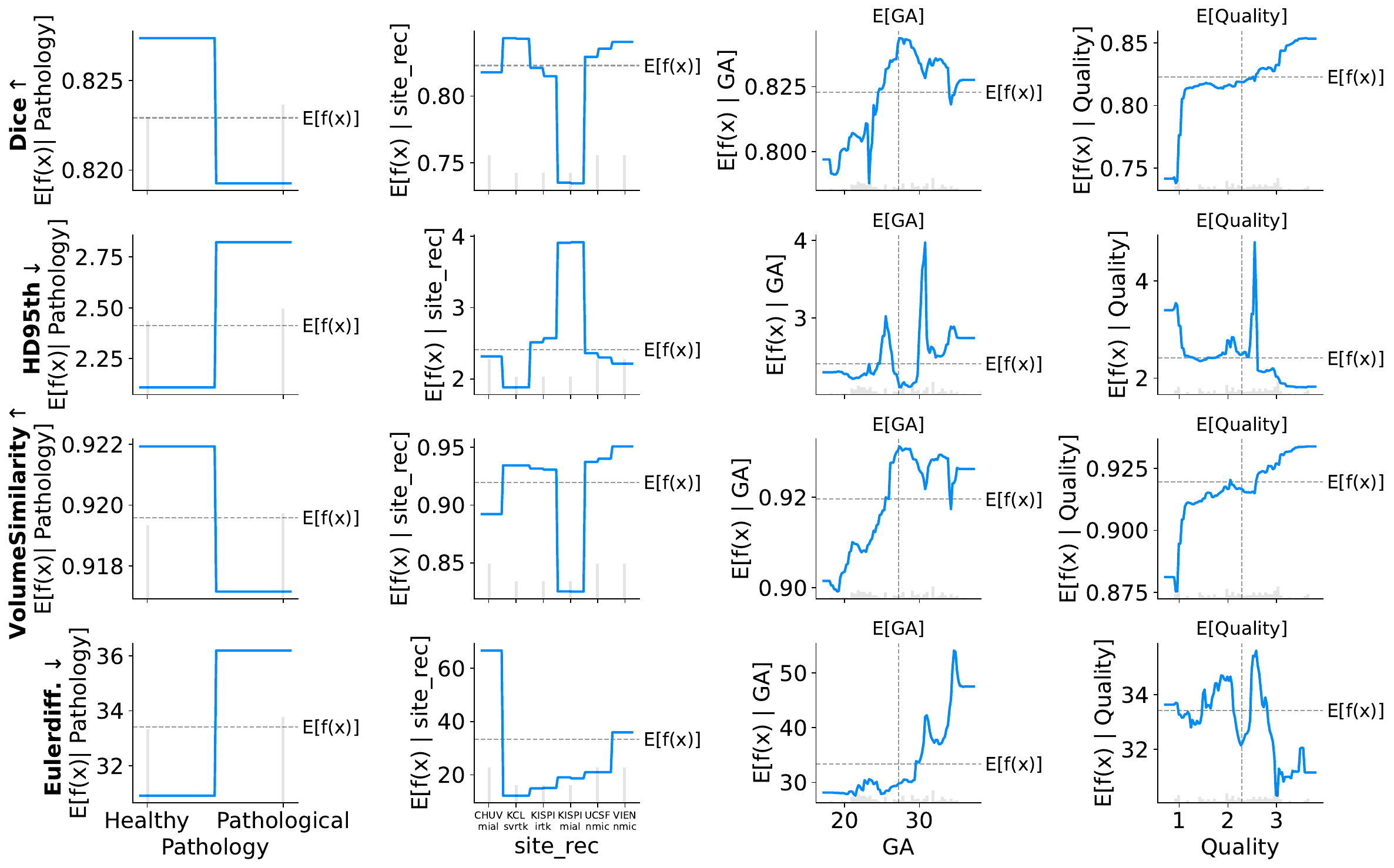}
    \captionof{figure}{\textbf{Conditional mean} plots for Dice, HD95, VS and ED, for different domain shift factors: pathology, site and SR, GA and image quality. The conditional mean shows how a given metric deviates from the global expected performance ($\mathbb{E}[f(x)]$) when a specific variable is used for conditioning.}
    \label{fig:cond_means}

    \includegraphics[width=\linewidth]{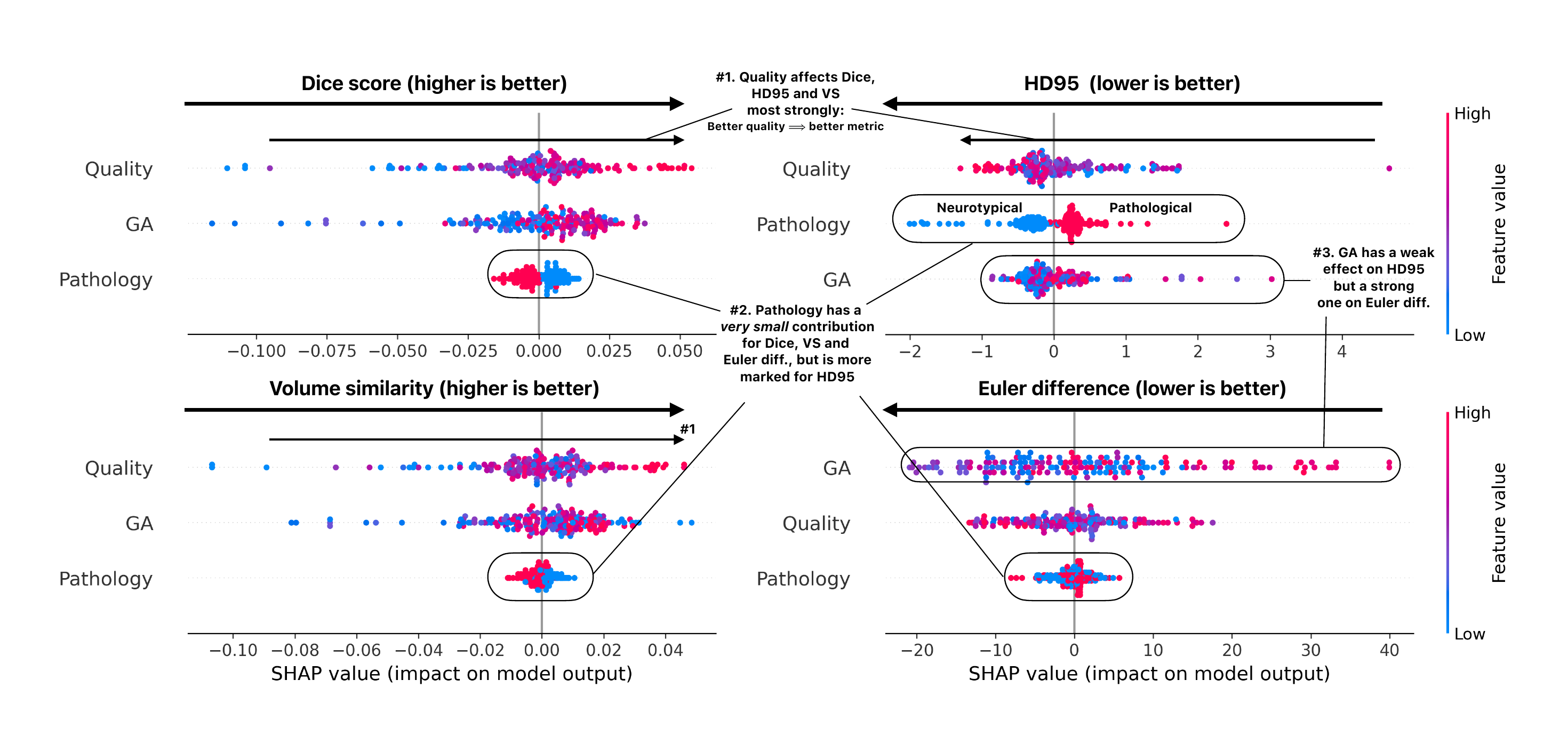}
    \captionof{figure}{\textbf{SHAP value distribution} across the data for segmentation metrics. Compared to conditional means (Fig. \ref{fig:cond_means}), SHAP values are the attribution of the impact on Dice or ED of different factors. Blue dots correspond to lower values of a variable (i.e. low GA, low quality), and red ones to higher values.}
    \label{fig:shap}
\end{strip}


\section{Discussion}

\subsection{FeTA 2024 results and ranking}
The multi-site, multi-task design of this challenge offered a unique opportunity to evaluate the progress and robustness of fetal brain image analysis algorithms. We summarize the main observations below. 
\paragraph{Submitted methods overview} Analyzing solutions of the best 3 teams for the segmentation task, we see that all of them used 3D models, specifically U-Nets or nnU-Nets. External data did not play a major role, with two of the top three teams relying solely on the provided dataset.  Notably, the first-place team, \texttt{cesne-digair}, uniquely incorporated a denoising autoencoder \citep{larrazabal2020post} to enhance segmentation accuracy. This approach significantly boosted their performance, leading to a 50\% improvement in the ED metric compared to the second-best team. This gain was particularly important for topological correctness, as no other metric showed such a large performance gap. All top-three teams applied extensive data augmentation, including techniques like SynthSeg \citep{billot2023synthseg}, combinations of standard augmentations, and deformable registration between pathological and healthy subjects to simulate greater anatomical diversity, as well as model ensembling.

The top three biometry teams each approached estimation differently—using keypoint regression, landmark heatmap segmentation, and other methods—but all relied on the results of the tissues segmentation and 3D models. Their model architectures varied (CNN, U-Net, and Transformer), indicating that no standard baseline has yet emerged for this task. The diversity in methods suggests that the field is still exploring the most effective strategies for biometry estimation.

\paragraph{Topology metric is a valuable add-on} In the brain tissue segmentation task, the introduction of the topology-aware metric provided meaningful complementary insights beyond traditional overlap-based measures. Despite the architectural diversity and growing methodological sophistication of the submitted approaches, the performance differences among the top teams were minimal, with Dice scores showing tight clustering, suggesting that gains in segmentation accuracy may be reaching a plateau. Differences in team rankings across evaluation metrics (Dice, HD95, VS, ED) highlight the need to consider complementary metrics beyond voxel-wise overlap.  Introducing ED as a ranking metric provided a more nuanced assessment of the segmentation quality. This is reflected in Table~\ref{tab:long_years}, where we see a marked improvement in ED. While teams did not specifically optimize their models for topological consistency, the new ranking scheme allowed us to discriminate between methods that otherwise had very similar performances (Table~\ref{tab:ranking_segm}). Evaluation noise, where performance variations across testing sets is larger than the difference across top-performing methods, is a well-known problem in medical imaging challenges~\citep{varoquaux2022machine}, and the introduction of an additional ranking metric allowed for selection methods with desirable properties. Further validation on clinical tasks leveraging surface extraction\citep{Clouchoux2011,Yehuda2023} would be needed to truly see the potential of encouraging topological consistency in the FeTA challenge.

\paragraph{Low-field MRI tissue segmentation quality is encouraging} The newly introduced 0.55T data from KCL provided an unexpected insight: it consistently achieved the highest segmentation accuracy across all sites. However, it is important to note that, to reduce domain shifts, we retrospectively selected high-quality reconstructions using a version of SVRTK specifically tailored for low-field MRI data\citep{Uus2020,uus2024scanner}. This careful case selection, paired with a very recent SR pipeline, might have positively biased the performance for this cohort. As such, performance on more challenging low-field cases remains to be fully assessed.
Nevertheless, these results are encouraging for two reasons: progress in SR pipelines~\citep{uus2024scanner,xu2023nesvor} means that more challenging cases will be successfully reconstructed, and that image quality will generally increase~\citep{sanchez2024assessing,uus2023retrospective}. Low-field MRI systems hold significant promise for expanding access to prenatal imaging, particularly in low- and middle-income countries. When combined with good image quality and advances in automatic fetal exam planning~\citep{neves2024fully}, they could meaningfully enhance prenatal care in resource-limited settings.

\paragraph{Automated biometry needs strong baselines to ensure meaningful progress} To reflect current clinical practice and bridge the gap between routine 2D fetal brain assessments and emerging 3D imaging techniques, we introduced a new biometry task focused on 2D brain measurements—key clinical indicators traditionally used to assess fetal neurodevelopmental status \citep{tilea2009cerebral, Lamon2024}. One of the striking results of this first edition is that most submissions did not manage to outperform a simple model, that predicts biometry solely based on gestational age, completely ignoring image information. Most teams built biometry estimators atop segmentation outputs, potentially propagating segmentation errors, particularly in smaller and more complex structures such as vermis and corpus callosum measurements. Importantly, the biometric measurements used in this challenge were derived from clinical 2D protocols. While these can be estimated from 3D volumes, they were originally designed for manual 2D evaluation. This suggests that alternative measurements, specifically tailored to leverage the full spatial context of 3D SRR images, may offer even more informative and robust indicators of fetal neurodevelopment, though such approaches remain largely unexplored.  Nevertheless, this first competition confirms, once more, the need to have strong baseline models and validation procedures~\citep{eisenmann2023winner}, and that deep learning might not always be the optimal solution~\citep{grinsztajn2022tree}.
 
\subsection{FeTA challenge in perspective}
\paragraph{FeTA across the years} A retrospective analysis of FeTA challenge results over the years revealed no statistically significant improvements in performance metrics, with the exception of ED at two out of five sites. Similarly, no notable improvement was seen at the label level, with GM, SGM, and BS consistently remaining the most challenging structures to segment. GM is particularly difficult due to its very thin appearance in fetal brains, where partial volume effects and complex surface morphology make it especially prone to topological segmentation errors, leading to significantly higher ED values compared to other labels. Moreover, both GM and SGM have inherently low tissue contrast in MRI, making them harder to distinguish accurately \citep{PRAYER2006199}. These challenges are further illustrated in Supplementary Materials A9.1, which provide qualitative examples showing that most segmentation errors occur in regions corresponding to GM, SGM, and BS. This outcome is not entirely unexpected, as most top-performing teams relied on similar 3D architectures—primarily 3D U-Net~\citep{ciek2016} and nnU-Net~\citep{isensee2018nnu}—enhanced with extensive data augmentation and model ensembling. These findings suggest that incremental architectural modifications or model engineering alone are unlikely to yield substantial gains, aligning with trends observed in other challenges where U-Net-based approaches often outperform more complex alternatives~\citep{eisenmann2023winner}. While these techniques help mitigate certain domain shifts related to scanner differences or pathological variations, some cases remain persistently difficult across all methods. Addressing these harder cases may require deeper domain expertise and a shift toward a more data-centric approach, prioritizing data quality, annotation consistency, and dataset diversity as core components of model development~\citep{sambasivan2021everyone, zha2023data}.

\paragraph{Sources of domain shifts} Domain shifts are widely recognized as a key challenge for deep learning methods in medical imaging~\citep{Dockès2021_domshift,wiles2021fine,richiardi2025domain}, yet the specific sources of these shifts are rarely disentangled. In our analysis, though not causal, we observed that image quality had the strongest impact on generalization performance: moving from the lowest to the highest quality levels resulted in an average Dice score difference of approximately 0.10. In contrast, gestational age had a more modest effect, influencing Dice scores by about 0.05, while the scanning site contributed a difference of around 0.075 between the best- and worst-performing centers. Interestingly, pathology was the least influential factor, accounting for only about 0.008 in Dice variation. Additionally, because Dice is known to be biased toward larger structures~\citep{maier2024metrics}, we also evaluated performance using a normalized Dice metric that accounts for label volume~\citep{raina2023tackling}. As detailed in the supplementary materials section A8, the normalized Dice scores yielded rankings nearly identical to those based on standard Dice, indicating that structure size did not significantly distort the comparative performance of the reviewed algorithms, suggesting that while the size bias exists, its effect was uniform across methods.

Our results show that, despite pathological cases making up only about one-third of the training data, models were still able to generalize to pathological examples in the test set. While performance was slightly lower for pathological subjects in some datasets compared to healthy subjects, submitted models demonstrated the ability to correctly handle both healthy and pathological data. Given the rarity and wide variability of fetal pathologies \citep{attallah2019fetal}, expanding pathological datasets—whether through additional real cases or synthetic data \citep{dannecker2024cina, synthfetpath_Misha_2025}—will be crucial to narrowing this performance gap and improving overall model robustness, which is an important step toward real-world clinical deployment.

Overall, our findings suggest that technical and acquisition-related factors may play a more significant role in out-of-domain generalization than subject-level clinical variables. Still, further causal investigations~\citep{castro2020causality} are needed to confirm these patterns and to avoid misinterpretation due to confounding factors.

\subsection{Roadmap for future advancements in fetal brain MRI analysis}
While many proposed solutions appear to be reaching a performance plateau, model-centric innovations still play an important role. That said, incorporating domain-specific augmentations and auxiliary learning objectives may lead to more impactful improvements than simply refining model architectures. For example, enforcing \textit{topological consistency} within the loss function, as demonstrated by \citet{de2022multi,li2023robust, lux2024topograph}—can help maintain anatomical plausibility in the predictions. Similarly, integrating \textit{uncertainty estimation} provides a powerful way of identifying low-confidence predictions, which is particularly relevant in clinical decision-support systems. Several studies~\citep{zenk2025comparative,molchanova2025structural} have demonstrated the utility of uncertainty-aware models for quality control in medical image segmentation.

Beyond model architecture, data-driven strategies hold substantial potential for improvement. A notable limitation of current solutions is the relatively modest use of \textit{external data}, which has been largely limited to \textit{healthy} subjects from datasets like dHCP or fetal brain atlases~\citep{Gholipour2017_atlass,kcl_dhcp_atals,price2019developing}. Leveraging broader, more diverse datasets — especially those capturing rare or pathological conditions — could support more robust and clinically useful models, though curating and annotating pathological datasets is a huge endeavor. 

Manually segmenting the fetal brain is a time-consuming and tedious task, susceptible to inter-rater variability~\citep{Payette2021_naturefetadata}, and the FeTA challenge data are not exempt from this issue. When comparing model performance to inter-rater variability, we observe that top-performing teams—achieving Dice scores around 0.82, HD95 around 2.2, and VS around 0.92—are approaching the best observed human agreement levels, previously estimated on a subset of data as $0.73 \pm 0.15$ for Dice, $3.45 \pm 2.34$ for HD95, and $0.86 \pm 0.10$ for VS~\citep{payette2023feta_2021rseults}. This raises the intriguing possibility that some predictions may be more faithful to the underlying anatomy than the ground truth annotations, potentially leading to penalization of high-performing models~\citep{valabregue2024comprehensive,valabregue2024unraveling}.

A promising direction to address the limitations of data diversity and annotation availability is data synthesis \citep{Zalevskyi2024_Journal}, particularly for generating rare or pathological fetal brain appearances. Recent work~\citep{dannecker2024cina,liu2024pepsi,kaandorp2025pathological} has highlighted the potential of synthetic data to augment training and improve sensitivity to abnormal anatomy. Moreover, the strong influence of image quality on generalization performance underscores the need for better modeling of artifacts specific to fetal brain SR pipelines~\citep{sanchez2024assessing}. These efforts, in combination with foundation models and domain adaptation techniques, offer exciting prospects for enhancing model generalization across scanners, domains, and populations, ultimately helping to mitigate model drift and support the development of more trustworthy AI systems.

\section{Conclusion}
The FeTA 2024 challenge provided a valuable opportunity to evaluate the progress made in fetal brain segmentation since previous editions and to expand the scope toward new, clinically relevant tasks such as biometry. Our additional validation using the Euler difference metric showed that some existing methods can already produce topologically consistent segmentations. However, achieving this consistency more reliably, particularly through improved segmentation losses, remains an open area for further development. Likewise, the successful application of models to low-field data, with surprisingly strong performance, highlights both the advancements in recent super-resolution methods and the models’ capacity to generalize across diverse imaging settings.

In the biometry task, this first edition offered key insights, particularly on the importance of providing simple baseline models to guide participants. It also led to the emergence of a promising approach for automated biometry prediction.

As the field of fetal brain MRI analysis continues to evolve, FeTA 2024 emphasizes the need not only for more powerful and innovative models but also for building reliable and generalizable tools that can support real-world clinical adoption.

\section*{Declaration of competing interests}
The authors declare that they have no known competing financial interests or personal relationships that could have appeared to influence the work reported in this paper.

\section*{Data availability}
Part of the challenge data is available on Synapse, as described in the paper.

\section*{Acknowledgments}
This research was funded by the Swiss National Science Foundation (182602, 215641, IZKSZ3\_218590), ERA-NET Neuron MULTI-FACT project (SNSF 31NE30 203977), UKRI FLF (MR/T018119/1), the Adaptive Brain Circuits in Development and Learning Project, University Research Priority Program of the University of Zürich; by the Vontobel Foundation; by the Anna Müller Grocholski Foundation and the Prof. Max Cloetta Foundation and DFG Heisenberg funding (502024488); we acknowledge the Leenaards and Jeantet Foundations as well as CIBM Center for Biomedical Imaging, a Swiss research center of excellence founded and supported by CHUV, UNIL, EPFL, UNIGE and HUG.

Diego Fajardo-Rojas would like to acknowledge funding from the EPSRC Centre for Doctoral Training in Smart Medical Imaging (EP/S022104/1).

Lyuyang Tong, Bo Du and Jingwen Jiang would like to acknowledge funding from the National Key Research and Development Program of China under Grants 2023YFC2705700, the National Natural Science Foundation of China under Grants 62306217 and 62225113, the Postdoctoral Fellowship Program of CPSF under Grant Number GZC20231987, the China Postdoctoral Science Foundation under Grant Number 2024T170686 and 2024M752471, the Major Program (JD) of Hubei Province (2023BAA017), the Innovative Research Group Project of Hubei Province under Grants 2024AFA017.

Gerard Martí-Juan is supported by the project PCI2021-122044-2A, funded by the project ERA-NET NEURON Cofund2, by MCIN/AEI/10.13039/501100011033/ and by the European Union NextGenerationEU/PRTR.

R. Hamadache holds an IFUdG PhD grant from the University of Girona. R. Hamadache and X. Lladó received support by the PID2023-146187OB-I00 from the Ministerio de Ciencia e Innovación, Spain.

M.O. Candela-Leal, A. Gondova, and S. You received support from the National Institute of Neurological Disorders and Stroke (R01NS114087) and National Institute of Biomedical Imaging and Bioengineering (R01EB031170) of the National Institutes of Health (NIH). 

\section*{Declaration of generative AI and AI-assisted technologies in the writing process}
During the preparation of this work, the authors used Grammarly and ChatGPT (GPT 4o) to assist with spell checking, grammar refinement, and language clarity improvements. After using this tool/service, the authors reviewed and edited the content as needed and take full responsibility for the content of the publication.

\bibliographystyle{elsarticle-harv}
\bibliography{refs}



\section*{Supplementary material}

# Supplementary Materials

**Contents**

# Appendix A1. Evaluation metrics description.

**Task 1: Segmentation**

To evaluate the performance of the segmentation algorithms, we used four complementary metrics:

1. **Dice Similarity Coefficient (Dice)**: Measures voxel-wise correspondence between the predicted and ground truth (GT) segmentations. It is computed as:

$$Dice = \frac{2 \cdot |A \cap B|}{|A| + |B|}$$

   where $A$ and $B$ represent the predicted and GT segmentation sets per label, respectively. Higher values of DSC indicate better overlap.

2. **Volume Similarity (VS)**: Assesses similarity of volumes between predicted and GT segmentations, defined as:

$$VS = 1 - \frac{|V_{pred} - V_{GT}|}{|V_{pred} + V_{GT}|}$$

   where $V_{pred}$ and $V_{GT}$ are the volumes of the predicted and GT regions, respectively. A value close to 1 indicates high similarity.

3. **Hausdorff Distance (HD95)**: Quantifies contour similarity between predicted and GT segmentations using the 95th-percentile Hausdorff distance:

$$HD95 = \max \left( \max_{x \in A} \min_{y \in B} \|x - y\|, \max_{y \in B} \min_{x \in A} \|x - y\| \right)$$

   where $A$ and $B$ are boundary points of the predicted and GT segmentations, and $\|\cdot\|$ denotes the Euclidean distance. Lower HD95 values indicate better contour agreement.

4. **Euler Characteristics (ED) Difference**: Evaluates topological similarity between predicted and GT segmentations, based on Betti numbers. The Euler characteristic is:

$$EC = \text{Betti}_0 - \text{Betti}_1 + \text{Betti}_2$$

   where $\text{Betti}_0$ is the number of connected components, $\text{Betti}_1$ the number of loops, and $\text{Betti}_2$ the number of voids. The EC difference is computed as:

$$ED = |EC_{pred} - EC_{GT}|$$

   Smaller differences indicate better topological alignment. The expected GT values for Betti numbers are: $BN_1 = 0$ and $BN_2 = 0$ for all brain tissues. For the eCSF, WM, ventricles, cerebellum, dGM, and brainstem, $BN_0 = 1$; for GM, $BN_0 = 2$.

These metrics together provide a comprehensive evaluation of segmentation accuracy, considering spatial overlap, volume, shape, and topology.

**Task 2: Biometry Estimation**

The primary metric for evaluating biometry estimation algorithms is **mean average percentage error (MAPE)**, quantifying error relative to actual measurements:

$$MAPE = \frac{1}{N} \sum_{i=1}^{N} \frac{|y_i - \hat{y}_i|}{y_i} \times 100$$

where $y_i$ and $\hat{y}_i$ are the ground truth and predicted measurements, respectively, and $N$ is the total number of measurements.

This metric accounts for variable target structure sizes and assesses the accuracy of the estimated biometric measurements.

# Appendix A2. Algorithm descriptions

**Team Algorithm Descriptions**

The algorithm descriptions are presented for all participating teams in any of the tasks, in the following order:

1. CEMRG
2. falcons
3. feta_sigma
4. hilab
5. Jwcrad
6. lmrcmc
7. LIT
8. mic-dkfz
9. paramahir_2023
10. pasteurdbc
11. unipd-sum-aug
12. UPFetal24
13. vicorob
14. qd_neuroincyte
15. CeSNE-DiGAIR

# Algorithm Description Guidelines

**For a challenge submission to be considered complete, participants** in addition to submitting a docker container, **must submit the following information**.

This must be submitted by **August 12, 2024**.

1. **Team Information**

**Team Name:** CEMRG

Team Members (include names and emails of all team members, add rows as necessary):

| Name | Email |
| --- | --- |
| Abdul Qayyum | a.qayyum@imperial.ac.uk |
| Moona Mazher | moona.mazher@gmail.com |
| Steven A Niederer | s.niederer@imperial.ac.uk |
| | |
| | |

Affiliations of each Team Member:

| |
| --- |
| Imperial College London, UK |
| University College London, UK |
| Imperial College London, UK |
| |

Max three (3) team members can be included in any publication resulting from this challenge.

- Would you like to be involved in any future publications? **(Yes/No)** If yes, which three team members are to be included?
    1. Abdul Qayyum
    2. Moona Mazher
    3. Steven A Niederer

- We will create a DockerHub where the dockers submitted to the FeTA Challenge will be stored. Do we have your permission to upload your docker? (**Yes**)

- There will be a poster session as part of the FeTA Challenge in conjunction with the PIPPI workshop. Would you be interested in participating in the poster session? (**Yes**)

- MICCAI 2024 will be an in-person event. Please state if you plan to attend in person (**No).**

2. **Model Information**

1

If deep learning was used: Yes

GPU training was performed on: Yes

Software used incl. version (i.e. Tensorflow, Pytorch, etc.): Pytorch

Please attach a description of your model highlighting the main features. This description must include the following details (unless the parameter is not applicable for your model):

- Model architecture

Hybrid Cross Attention Transformer and CNN model used for feta segmentation task.

The proposed 3D model leverages an encoder-decoder architecture. On the encoder side, Swin Transformer-based blocks are integrated with a Cross Attention window-based mechanism, while the decoder side employs a conventional 3D CNN module. The Swin Transformer (Liu et al., 2021a) is well-suited for pixel-level tasks like image registration and segmentation, as it generates hierarchical feature maps at various resolutions through patch merging layers. A key feature of the Swin Transformer is its shifted window-based self-attention mechanism. Unlike the Vision Transformer (ViT) (Dosovitskiy et al., 2020), which calculates relationships between each token and all other tokens in the self-attention modules, the Swin Transformer computes self-attention within non-overlapping local windows, evenly partitioned across the original and lower resolution feature maps. In contrast to the original Swin Transformer, this work adapts the design to accommodate non-square images by using rectangular-parallelepiped windows. At each resolution, the first Swin Transformer block partitions the feature maps into non-overlapping windows, starting from the top-left voxel. The Cross Attention Transformer (CAT) is then applied locally within each window. To introduce interactions between neighboring windows, the Swin Transformer employs a shifted window design, where the windowing configuration shifts in successive Swin Transformer blocks.

The CAT block is designed to compute new feature tokens with corresponding relevance from input feature set b to feature set s through an attention mechanism. The feature sets are partitioned into two different sets of windows: the base window set and the searching window set , both of which have the same size but different window dimensions. Each base window is projected into a query set, while each search window is projected into key and value sets through linear layers. The Window-based Multi-head Cross Attention (W-MCA) then calculates cross attention between these windows, updating each base window with weighted information from the corresponding searching window. The updated output set is passed through a 2-layer MLP with GELU non-linearity to enhance learning, with LayerNorm (LN) applied before each W-MCA and MLP module to ensure layer validity. This advanced CAT block facilitates automatic correspondence finding between images, enabling efficient feature fusion within the network.

For feature extraction in the encoder, each block consists of 3D convolutional layers, batch normalization, and ReLU activation functions. A 3D max-pooling layer is used to reduce

the spatial size of the input image. The spatial resolution decreases with the increase in the number of layers in the encoder and is recovered in the decoder. To restore spatial resolution in the decoder, 3D bilinear upsampling is applied. The kernel size is set to 3x3x3 for both encoder and decoder, with feature maps set to 16, 32, 64, 128, and 256 in each encoder block. The 3D max-pooling layer in the encoder uses a 2x2x2 kernel size to downsample the spatial resolution. In the decoder, 3D transpose convolutional layers with a 2x2x2 kernel size and stride of 2 are used for upsampling. The output of each encoder block is concatenated with the corresponding decoder block, and a final 1x1 convolutional layer with a softmax function generates the output segmentation map.

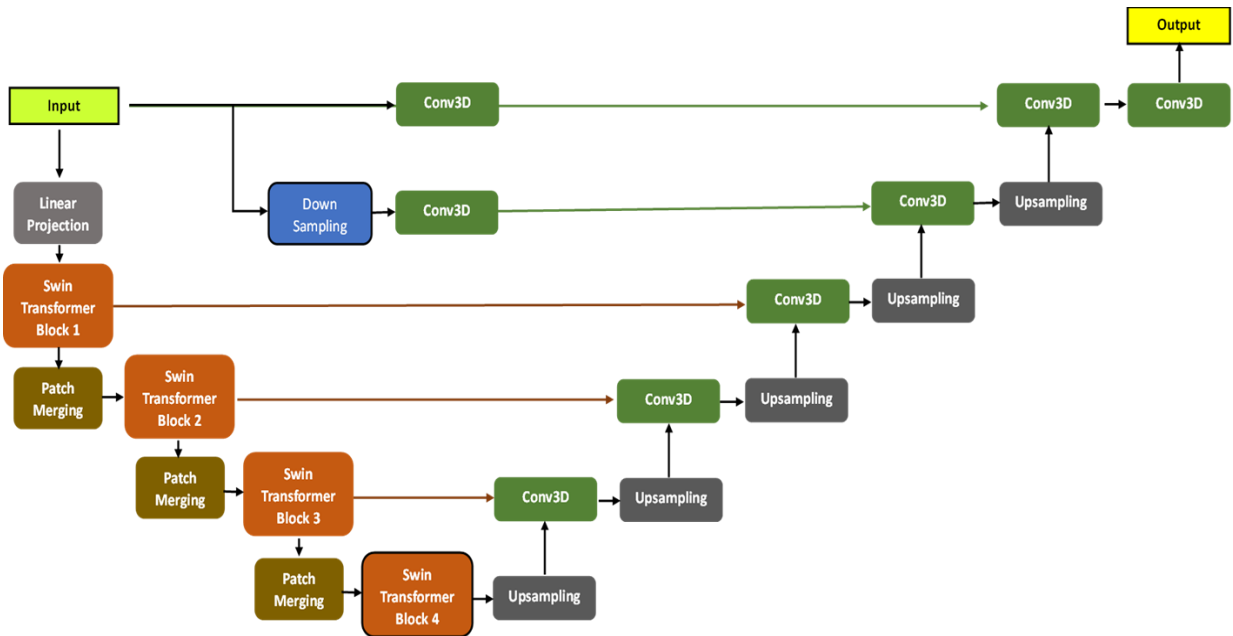

Figure.1 Proposed base on cross window attention module

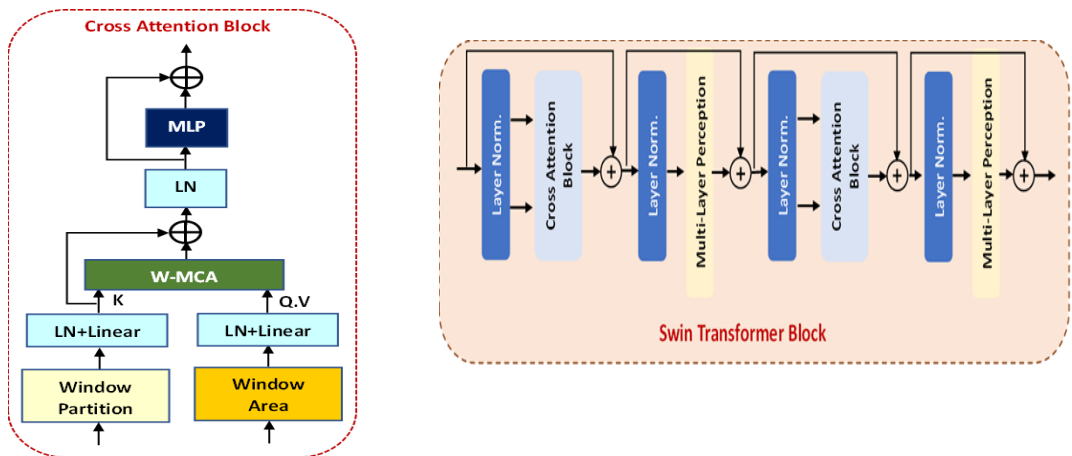

Figure2. Cross attention block and Swin transformer block used in proposed segmentation model.
- Number of layers
-

The architecture consists of five blocks in both the encoder and decoder stages. Each block includes 3D convolutional layers with a 3x3x3 kernel size, followed by ReLU activation, batch normalization (BN), and a 3D max-pooling layer. The encoder employs hybrid transformer blocks for feature extraction, while the decoder relies on convolutional layers.

In the encoder, each block uses 3D convolutional layers, batch normalization, and ReLU activation functions to extract features. A 3D max-pooling layer with a 2x2x2 kernel size is applied to progressively reduce the spatial size of the input image as the network depth increases. On the decoder side, the spatial resolution is restored using 3D bilinear upsampling. The decoder also employs 3D transpose convolutional layers with a 2x2x2 kernel size and a stride of 2 for upsampling. The feature maps in the encoder blocks are set to 16, 32, 64, 128, and 256, corresponding to the depth of the network.

Each decoder block receives inputs from the corresponding encoder block through concatenation. Finally, a 1x1 convolutional layer, followed by a softmax function, generates the output segmentation map.

- 
- Convolution kernel size
- 3x3
- Initialization
- Kaming he
- Optimizer
- Adam
- Cross-validation used?
- 5-fold
- Number of epochs
- 1500
- Number of trainable parameters
- 31079548
- Learning Rate and schedule
- Yes
- Loss Function
- BCE+Dice
- Dimensionality of input/output (ie: 2D,3D, 2D+, etc.)
- **3D**
- Batch Size
- 2
- Preprocessing steps used (ie data normalization, creation of patches, etc.),
- data normalization, created patches
- Data Augmentation steps (ie – rotation, flipping, scaling, blur, noise, etc.),
- Horizontal Flipping, Vertical flipping, scaling, normalization.
- 
- External dataset used? (allowed, but it needs to be publicly available
- **No**
- Framework (ie – MONAI, nnUNet, etc.)
- For model développement, PyTorch, for training, testing and optimization using nnUNet.
- 
- Number of models trained for final submission
- 6 models are trained and chosen the best one.

- Post-Processing Steps (ie – ensemble network, voting, label fusion)
-
- Sliding window approach used for post-processing.
-
- Clearly state which aspects are original work (if any) or already existing work
- The CAT block is designed to compute new feature tokens with corresponding relevance from input feature set b to feature set s through an attention mechanism. The feature sets are partitioned into two different sets of windows: the base window set and the searching window set , both of which have the same size but different window dimensions. Each base window is projected into a query set, while each search window is projected into key and value sets through linear layers. The Window-based Multi-head Cross Attention (W-MCA) then calculates cross attention between these windows, updating each base window with weighted information from the corresponding searching window. The updated output set is passed through a 2-layer MLP with GELU non-linearity to enhance learning, with LayerNorm (LN) applied before each W-MCA and MLP module to ensure layer validity. This advanced CAT block facilitates automatic correspondence finding between images, enabling efficient feature fusion within the network.
-
- Include relevant citations, as well as if existing code/software libraries/packages were used
-
- We will publish the code soon in the following GitHub link.
- https://github.com/RespectKnowledge
-
- Which FeTA cases were included in the training and testing (ie – all cases, only pathological, only 1 institution, etc.)
-
- We have used all cases.
-
- Training/validation/testing data splits
- 80/20
-
- Hyperparameter tuning performed
- No
-
- Training time
-
- 4 hours


Note: If a deep learning method was used, please provide the equivalent appropriate information as listed above.

# Algorithm Description Guidelines

**For a challenge submission to be considered complete, participants** in addition to submitting a docker container, **must submit the following information**.

This must be submitted by **August 12, 2024**.

1. **Team Information**

**Team Name: Fetal Automatic Landmark Computation and Optimization for Neuroimaging Segmentation (FALCONS)**

Team Members (include names and emails of all team members, add rows as necessary):

| Name | Email |
| --- | --- |
| Milton O. Candela-Leal | milton_candela@hotmail.com |
| Andrea Gondova | andrea.gondova@childrens.harvard.edu |
| Sungmin You | Sungmin.you@childrens.harvard.edu |
| Kiho Im | kiho.im@childrens.harvard.edu |
| P. Ellen Grant | Ellen.Grant@childrens.harvard.edu |

| Affiliations of each Team Member: |
| --- |
| Fetal Neonatal Neuroimaging and Developmental Science Center, Boston Children's Hospital, Harvard Medical School |
| Division of Newborn Medicine, Boston Children's Hospital, Harvard Medical School |
| Department of Pediatrics, Harvard Medical School |
| Department of Radiology, Boston Children's Hospital, Harvard Medical School |

Max three (3) team members can be included in any publication resulting from this challenge.

- Would you like to be involved in any future publications? (**Yes**/No) If yes, which three team members are to be included?
    1. Milton O. Candela-Leal
    2. Andrea Gondova
    3. Sungmin You

- We will create a DockerHub where the dockers submitted to the FeTA Challenge will be stored. Do we have your permission to upload your docker? (**Yes**/No)

- There will be a poster session as part of the FeTA Challenge in conjunction with the PIPPI workshop. Would you be interested in participating in the poster session? (**Yes**/No)

- MICCAI 2024 will be an in-person event. Please state if you plan to attend in person (**Yes**/No).

2. **Model Information**

If deep learning was used: 2D Attention Gated U-Net

GPU training was performed on: NVIDIA RTX A5000

Software used incl. version (i.e. Tensorflow, Pytorch, etc.): Tensorflow(2.10.0), FMRIB Software Library(FSL 6.0), CIVET(2.1.0), Advanced Normalization Tools(ANTs), Scikit-learn (1.5.1)

---

The FALCONS pipeline consists of three steps: preprocessing, segmentation, and biometry inference. The preprocessing steps contain brain extraction, template alignment, and intensity normalization to minimize potential variations in MRI data acquired by different sites, machines, and field strength. The procedure is based on our fetal brain MRI processing pipeline presented in previous work (Yun et al. 2022; You et al. 2024), while the brain extraction algorithm is replaced with the alternative, publicly available brain extraction tool (BET) (Smith 2002). We performed two-step registration to align the available fetal data to the 31-week gestational age (GA) public fetal brain templates (Serag et al. 2012; https://brain-development.org/brain-atlases/), with the aim improve the segmentation and biometry inference. The transformation matrix is separately stored to re-align the segmentation output to the native space. Lastly, for the intensity normalization, we corrected intensity inhomogeneity via N4 bias field correction (Tustison et al. 2010).

The employed segmentation algorithm is an ensemble of three 2D Attention Gated U-Net models, each trained separately for axial, coronal, and sagittal planes, and their outputs are combined using a multi-view aggregation with test-time augmentation (MVA-TTA) (Hong et al. 2020) to leverage the effect of iterative augmentation with majority voting to stabilize and improve quality of the final segmentation outputs. This framework effectively captures and emphasizes relevant features in fetal brain MRIs while suppressing irrelevant regions, enhancing overall segmentation performance and precision of biometry estimations that are based on the segmentation outputs.

Each 2D U-Net model consists of an encoder-decoder structure with attention gates integrated into skip connections, with the encoder and decoder containing 40 2D-convolution layers in total with pooling or up-sampling layers. The input and output data are 2D image slices with dimensions 192-by-192, and a normalized batch size of 32 is used for training. The employed hybrid loss function (Hong et al. 2020) is a combination of Dice loss and focal loss that balances overall segmentation accuracy and detailed segmentation along the borders between brain tissues. The convolution kernel size used is 3-by-3, with weights initialized using He initialization (He et al. 2015). The optimizer utilized for training is Adam (Kingma and Ba 2014), chosen for its efficiency in handling sparse gradients, with the learning rate set to 1e-5. Each 2D Unet has approximately 13M trainable parameters and was trained for 200 (training time of approx. 26 hours), 100 (training time of approx. 12 hours), and 70 epochs (training time of approx. 18 hours) for the axial, sagittal, and coronal model, respectively, using Nvidia RTX A5000 GPU. The model was implemented using the TensorFlow library (Dillon et al. 2017).

For the training of segmentation models, we included 70 images from the publicly available developing Human Connectome Project (dHCP) database (Edwards et al. 2022), in addition to the MRI data provided by the FeTA2024 challenge: 70 images from University Children's Hospital Zurich (Zurich); and 16 images from General Hospital Vienna/Medical University of Vienna (Vienna). In the case of the Vienna dataset, we included subjects with high-quality images for which the fetal brain could be successfully extracted by the BET algorithm. As an attempt to improve the generalizability of our models, we employed several augmentation approaches during training, namely: rotation (range=30°), width/height shift (range=0.2), vertical/horizontal flip, zooming (range=0.2), brightness (range=0.3), gaussian noise (std=0.2), and gaussian blurring (std=2). A hold-out test set of 10 Zurich and 10 dHCP subjects stratified on pathology and GA at scan was used to evaluate segmentation quality.

The output segmentation and T2-weighted data are used to estimate five biometry measurements within the 31-week GA template space. For the brain biparietal diameter (bBIP) and skull biparietal diameter (sBIP) in the axial plane, and maximum transverse cerebellar diameter (TCD) in the coronal plane, we use a bounding box strategy to maximize distance in the relevant structure and imaging plane. For the axial plane, we constrain the search by basal ganglia colocalization to identify similar slices for parietal eminences' location. For sBIP, we dilate the external cerebrospinal fluid (CSF) label and locate the skull boundary by identifying the largest intensity change in T2-weighted images. To measure the height of the vermis (HV) and the length of the corpus callosum (LCC), we first find centroids of cerebral tissue before and after flipping the bounding box of the cerebrum along the sagittal place. The final midsagittal slice is determined from the average of these centroids to account for hemispheric asymmetries. On the midsagittal slice, we locate HV by finding the most caudal points of the label for the cerebellum and the second point that maximizes the Euclidean distance. Similarly to the manual approach described in (Bach Cuadra, 2023), we also tested a parallel line to the brainstem strategy (estimated from the principal direction of variance within the brainstem label). However, the simpler maximization approach proved more effective for HV estimation. For LCC, we locate the corpus callosum (CC) as the overlap between white matter and dilated ventricle segmentation on the midsagittal slice, then identify the splenium as the most posterior point and measure LCC as the distance that maximizes the Euclidean distance on the CC mask. The identified landmarks are then aligned to the native space for future use.

Given the challenges we encountered, particularly with midsagittal slice localization and lowered measurement performance in pathological subjects, we utilized the strong collinearity between biometry measurements to improve accuracy. We first compare our estimates to a 'normative' model derived from manual biometry estimates from Zurich data (both typical and pathological subjects were included since their age-related slopes estimated by robust linear regression for a given metrics did not differ significantly when compared with Z-scores), and then flag values deviating from the estimated 'normative' ranges (outside α=0.99). These flagged values are then corrected using a pre-trained iterative Bayesian Ridge imputer (shape parameters α of 0.01 and λ of 0.001, and inverse scale parameters α of 0.1 and λ of 0.0001). The model's performance was validated by 10-fold cross-validation after artificially flagging 10% of true values and was implemented using scikit-learn v1.5.1. While this approach addresses some methodological difficulties with landmark localization, it may not fully account for natural biometry deviations in very pathological cases and should be further investigated.

The integration of attention gates in the UNet architecture and the ensemble approach is based on existing methodologies(Oktay et al. 2018; Hong et al. 2020), but the specific implementation details and hyperparameter tuning to the context of the FeTA2024 challenge constitute original contributions. Additionally, our framework leverages a series of preprocessing steps, including brain extraction, alignment, and non-uniform intensity correction, contributing to stable segmentation and biometry inference in diverse MRIs with potential variations.

## Algorithm Description Guidelines

**For a challenge submission to be considered complete, participants** in addition to submitting a docker container, **must submit the following information**.

This must be submitted by **August 12, 2024**.

1. **Team Information**

**Team Name:** feta_sigma

Team Members (include names and emails of all team members, add rows as necessary):

| Name | Email |
| --- | --- |
| Jiang Jingwen | williamsriver@whu.edu.cn |
| Zhang Chengsheng | chenshengzhang@whu.edu.cn |
| Wang Hanling | Hanling.Wang21@student.xjtlu.edu.cn |
| Zhang Xuezhi | zhangxz2120@mails.jlu.edu.cn |
| Cao Jiarui | CaoJiarui@stu.cqu.edu.cn |
| Tong Lyuyang | Lyuyangtong@whu.edu.cn |
| Du Bo | dubo@whu.edu.cn |

Affiliations of each Team Member:

Jiang Jingwen, Wuhan University, School of Computer Science

Zhang Chengsheng, Wuhan University, School of Computer Science

Wang Hanling, Xi'an Jiaotong-Liverpool University, School of Advanced Technology

Zhang Xuezhi, Wuhan University, School of Computer Science

Cao Jiarui, Wuhan University, School of Computer Science

Tong Lyuyang, Wuhan University, School of Computer Science

Du Bo, Wuhan University, School of Computer Science

Max three (3) team members can be included in any publication resulting from this challenge.

- Would you like to be involved in any future publications? **(Yes/No)** If yes, which three team members are to be included? **Yes**
  1. ___Jiang Jingwen_____
  2. ___Tong Lyuyang_____
  3. ___Du Bo_____

- We will create a DockerHub where the dockers submitted to the FeTA Challenge will be stored. Do we have your permission to upload your docker? (**Yes/No**)
  **Yes**
- There will be a poster session as part of the FeTA Challenge in conjunction with the PIPPI workshop. Would you be interested in participating in the poster session? (**Yes/No**) **Yes**

- MICCAI 2024 will be an in-person event. Please state if you plan to attend in person (**Yes/No**). Yes

2. **Model Information**

If deep learning was used: Yes

GPU training was performed on: NVIDIA GeForce RTX 4090

Software used incl. version (i.e. Tensorflow, Pytorch, etc.): Pytorch

Please attach a description of your model highlighting the main features. This description must include the following details (unless the parameter is not applicable for your model):

- Model architecture
- UxLSTMEnc,UNet
-
- Number of layers
- Encoder:6,Decoder:5
-
- Convolution kernel size
- 3x3x3
-
- Initialization
- Default
-
- Optimizer
- SGD
-
- Cross-validation used?
- 5-fold
-
- Number of epochs
- 200
-
- Number of trainable parameters
- UxLSTMEnc：43701120
- nnUnetForSegAll：31200424
- nnUnetForSegBg：31195594
-
- Learning Rate and schedule
- 0.001, Linear Warm-up Cosine Annealing
-
- Loss Function
- combines Dice and Cross-Entropy losses
-
- Dimensionality of input/output (ie: 2D,3D, 2D+, etc.)
- 3D
-

- Batch Size
- 1
-
- Preprocessing steps used (ie data normalization, creation of patches, etc.)
- ZScoreNormalization
- Resampling data and segmentation
- Convert data type from numpy to tensor
-
- Data Augmentation steps (ie – rotation, flipping, scaling, blur, noise, etc.)
- Spatial Transform:
- Rotation with angles for x,y,z axes;
- Random scaling in the range(0.7,1.4)
- Random cropping
- Noise and Blur:0.1probability Gaussian Noise and sigma values Gaussian Blur
- Intensity Adjustments:
- Brightness multiplicative adjustment in range (0.75, 1.25) with a probability of 0.15
- Contrast Augmentation with a probability of 0.15
- Resolution Simulation:
- Simulate Low Resolution applied with zoom range (0.5, 1) and varying probabilities
- Gamma Transformation: applied with two sets of parameters, with probabilities of 0.1 and 0.3
- Mirror Transform
- External dataset used? (allowed, but it needs to be publicly available
- no
-
- Framework (ie – MONAI, nnUNet, etc.)
- nnUNet
-
- Number of models trained for final submission
- 3
-
- Post-Processing Steps (ie – ensemble network, voting, label fusion)
- **Merging Predictions**: ensemble network(UxLSTMEnc and nnUnet)
- **Background Masking**: label fusion with foreground mask
-
- Clearly state which aspects are original work (if any) or already existing work
- nnUnet: Isensee, Fabian, et al. "nnU-Net: a self-configuring method for deep learning-based biomedical image segmentation." *Nature methods* 18.2 (2021): 203-211.
- uxLSTM: Chen, Tianrun, et al. "xLSTM-UNet can be an Effective 2D\& 3D Medical Image Segmentation Backbone with Vision-LSTM (ViL) better than its Mamba Counterpart." *arXiv preprint arXiv:2407.01530* (2024).
-
- Include relevant citations, as well as if existing code/software libraries/packages were used
- UxLSTM\nnUNet(ref + code)
-
- Which FeTA cases were included in the training and testing (ie – all cases, only pathological, only 1 institution, etc.)
- All
-
- Training/validation/testing data splits

- 5-fold cross-validation:120/120/0
-
- Hyperparameter tuning performed
- Not any
- Training time
- 5hours 160epochs 5-fold validation



Note: If a deep learning method was used, please provide the equivalent appropriate information as listed above.

# Algorithm Description Guidelines

**For a challenge submission to be considered complete, participants** in addition to submitting a docker container, **must submit the following information**.

This must be submitted by **August 12, 2024**.

1. **Team Information**

**Team Name:** hilab

Team Members (include names and emails of all team members, add rows as necessary):

| Name | Email |
| --- | --- |
| Muyang Li | 18190351554@163.com |
| Jia Fu | jia.fu@std.uestc.edu.cn |
| Guotai Wang | guotai.wang@uestc.edu.cn |
|  |  |
|  |  |

Affiliations of each Team Member:

| |
| --- |
| School of Mechanical and Electrical Engineering, University of Electronic Science and Technology of China, Chengdu, China |
| School of Mechanical and Electrical Engineering, University of Electronic Science and Technology of China, Chengdu, China |
| School of Mechanical and Electrical Engineering, University of Electronic Science and Technology of China, Chengdu, China |
| |
| |

Max three (3) team members can be included in any publication resulting from this challenge.

- Would you like to be involved in any future publications? **(Yes/No)** If yes, which three team members are to be included?
    1. ___Muyang Li_____
    2. ___Jia Fu_____
    3. ___Guotai Wang_____
- We will create a DockerHub where the dockers submitted to the FeTA Challenge will be stored. Do we have your permission to upload your docker? (**Yes/No**)
    Yes
- There will be a poster session as part of the FeTA Challenge in conjunction with the PIPPI workshop. Would you be interested in participating in the poster session? (**Yes/No**)
    Yes

- MICCAI 2024 will be an in-person event. Please state if you plan to attend in person (**Yes/No**).
  N0

2. **Model Information**

If deep learning was used: Yes

GPU training was performed on: Nvidia GeFor-ce RTX 2080 Ti

Software used incl. version (i.e. Tensorflow, Pytorch, etc.): Pytorch

Please attach a description of your model highlighting the main features. This description must include the following details (unless the parameter is not applicable for your model):

- Model architecture
- Number of layers
- Convolution kernel size
- Initialization
- Optimizer
- Cross-validation used?
- Number of epochs
- Number of trainable parameters
- Learning Rate and schedule
- Loss Function
- Dimensionality of input/output (ie: 2D,3D, 2D+, etc.)
- Batch Size
- Preprocessing steps used (ie data normalization, creation of patches, etc.)
- Data Augmentation steps (ie – rotation, flipping, scaling, blur, noise, etc.)
- External dataset used? (allowed, but it needs to be publicly available)
- Framework (ie – MONAI, nnUNet, etc.)
- Number of models trained for final submission
- Post-Processing Steps (ie – ensemble network, voting, label fusion)
- Clearly state which aspects are original work (if any) or already existing work
- Include relevant citations, as well as if existing code/software libraries/packages were used
- Which FeTA cases were included in the training and testing (ie – all cases, only pathological, only 1 institution, etc.)
- Training/validation/testing data splits
- Hyperparameter tuning performed
- Training time

Note: If a deep learning method was used, please provide the equivalent appropriate information as listed above.

**Model Description**

Our model is based on the nnU-Net architecture(https://github.com/MIC-DKFZ/nnUNet)[1], a robust and flexible framework for medical image segmentation tasks, featuring 6 descending and 6 ascending layers. It employs a 3 × 3 × 3 convolution kernel throughout the network. All parameters are initialized randomly. Stochastic Gradient Descent (SGD) is used for optimization, and the model training follows a 5-fold cross-validation strategy. The model is trained for 400 epochs, containing 31,200,424 trainable parameters. The initial learning rate is set to 0.01 with a decay strategy to ensure convergence.

The loss function is a combination of Cross-Entropy and Dice Loss to optimize both pixel-wise accuracy and segmentation overlap. The model processes 3D input and output data, with a batch size of 2 due to the high memory requirements of 3D data processing.

**Data Preprocessing and augmentation**

For data preprocessing, we first cropped the non-zero region of each reconstructed fetal brain image. Then, we used histogram equalization for intensity normalization and Z-score normalization to enhance training stability.

For data augmentation, standard data augmentation strategies were used, including random rotations, scaling, Gaussian noise, Gaussian blur, brightness and contrast adjustments, simulation of low resolution, gamma augmentation, and mirroring. Additionally, a copy-paste[2] data augmentation technique with adjusted probabilities was employed to handle rare and difficult cases. We categorized the training samples into four types based on whether they were pathological or neurotypical, and whether the gestational age was more than 25 weeks. We set different probabilities to select these four different classes of images for copy-paste, with the probability of samples with pathological and gestational age less than 25 weeks being selected set to the highest. Additionally, we duplicated the pathological cases and cases with a gestational age of less than 25 weeks twice in the training data.

**Implementation Details**

We used all provided data of the FeTA 2024 dataset and did not use any external dataset. A total of 5 models were trained for the final submission, with an ensemble network used to combine their predictions. The implementation includes original contributions, such as applying histogram equalization to 3D images, introducing differentiated probabilities for sample selection in random copy-paste augmentations, and strategically replicating challenging cases in the training data. The remaining aspects follow the existing nnU-Net framework, as referenced in the original nnU-Net paper. The data was split with a 4:1 ratio for training and validation purposes, using the default hyperparameters provided by nnU-Net. In the training process, each epoch cost about 120s and the entire training process took approximately 66.7 hours. For Post-Processing Steps, we used ensemble network.

# Algorithm Description Guidelines

**For a challenge submission to be considered complete, participants** in addition to submitting a docker container, **must submit the following information**.

This must be submitted by **August 12, 2024**.

1. **Team Information**

**Team Name:**  jwcrad

Team Members (include names and emails of all team members, add rows as necessary):

| Name | Email |
| --- | --- |
| Jae Won Choi | djc0105@gmail.com |
|  |  |
|  |  |
|  |  |
|  |  |

Affiliations of each Team Member:
Department of Radiology, Seoul National University Hospital

Max three (3) team members can be included in any publication resulting from this challenge.

- Would you like to be involved in any future publications? (**Yes/No**) If yes, which three team members are to be included?
  1. Jae Won Choi
  2. 
  3. 

- We will create a DockerHub where the dockers submitted to the FeTA Challenge will be stored. Do we have your permission to upload your docker? (**Yes/No**)

- There will be a poster session as part of the FeTA Challenge in conjunction with the PIPPI workshop. Would you be interested in participating in the poster session? (**Yes/No**)

- MICCAI 2024 will be an in-person event. Please state if you plan to attend in person (**Yes/No**).

2. **Model Information**

If deep learning was used: Yes

GPU training was performed on: 1 × NVIDIA GeForce RTX 4090

Software used incl. version (i.e. Tensorflow, Pytorch, etc.): PyTorch 2.2.2

.

Please attach a description of your model highlighting the main features. This description must include the following details (unless the parameter is not applicable for your model):

- Model architecture
- Number of layers
- Convolution kernel size
- Initialization
- Optimizer
- Cross-validation used?
- Number of epochs
- Number of trainable parameters
- Learning Rate and schedule
- Loss Function
- Dimensionality of input/output (ie: 2D,3D, 2D+, etc.)
- Batch Size
- Preprocessing steps used (ie data normalization, creation of patches, etc.)
- Data Augmentation steps (ie – rotation, flipping, scaling, blur, noise, etc.)
- External dataset used? (allowed, but it needs to be publicly available
- Framework (ie – MONAI, nnUNet, etc.)
- Number of models trained for final submission
- Post-Processing Steps (ie – ensemble network, voting, label fusion)
- Clearly state which aspects are original work (if any) or already existing work
- Include relevant citations, as well as if existing code/software libraries/packages were used
- Which FeTA cases were included in the training and testing (ie – all cases, only pathological, only 1 institution, etc.)
- Training/validation/testing data splits
- Hyperparameter tuning performed
- Training time

Note: If a deep learning method was used, please provide the equivalent appropriate information as listed above.

# Algorithm description for team jwcrad in FeTA 2024 Tasks 1 and 2

**Dataset**

For the segmentation task, all 120 cases from the training dataset were utilized. For the biometry task, 102 cases from the training dataset were selected. Out of the 110 cases in the training dataset with available ground truth biometry measurements, 8 cases with missing measurements were excluded. No external datasets were employed. Both tasks were based on 5-fold cross-validation using the same split folds. The entire training dataset of 120 cases was stratified and split by gestational age and super-resolution reconstruction methods.

**Proposed Method**

We used 3D image inputs and outputs for both tasks, without incorporating additional inputs such as gestational age. The primary loss functions were a compound loss of Dice and cross-entropy for the segmentation task and mean squared error for the biometry task. Additionally, a custom auxiliary loss function based on transformation consistency, inspired by applications in semi-supervised segmentation [1] and image-to-image translation [2], was applied to both tasks. This auxiliary loss enforced consistency between the network outputs of transformed inputs and the transformed network outputs of the original inputs, using random 90-degree rotations as the transformation. The weight of the auxiliary loss was set to 0.1.

The biometry task was approached using heatmap regression rather than direct coordinate computation. The measurement labels were converted into 10-channel heatmaps, with each channel containing a Gaussian distribution centered at the corresponding measurement point. The order of the channels was fixed as follows: anterior point of the length of the corpus callosum (LCC), posterior point of the LCC, superior point of the height of the vermis (HV), inferior point of the HV, right point of the brain biparietal diameter (bBIP), left point of the bBIP, right point of the skull biparietal diameter (sBIP), left point of the sBIP, right point of the transverse cerebellar diameter (TCD), and left point of the TCD. A dynamic standard deviation proportional to the length of the bBIP was employed in the Gaussian distribution for the heatmap, defined as the length of the bBIP divided by the pixel spacing and a scale factor of 8.

**Preprocessing**

For the segmentation task, all data were resampled to an isotropic resolution of 1 mm using trilinear interpolation, and intensities were normalized using z-score normalization. During training, random cropping was performed with a patch size of 96 × 96 × 96, which was also used as the input size during inference.

For the biometry task, masking and cropping were initially performed on the input images using the segmentation labels. The segmentation labels were merged to obtain a binary foreground label, which was then dilated to introduce padding. The intensity values outside the dilated foreground label were set to zero, and cropping was performed around the dilated foreground. The cropped image was then resampled to a fixed size of 128 × 128 × 128, followed by z-score intensity normalization.

**Network architecture**

We utilized the Residual-USE-Net [3], a 3D U-Net variant featuring an encoder with residual convolution blocks and a decoder with plain convolution blocks, integrated with residual squeeze-and-excitation (SE) blocks. This architecture was adopted from the residual variant of the nnU-Net framework [4] and USENet [5]. Each convolution block was implemented as two sets of convolutions with a kernel size of 3, batch normalization, and

Leaky ReLU activation layers. In the residual block, the residual summation occurred before the final activation. The network had a base filter size of 32 for the convolution layers and included 4 skip connections. The reduction ratio of the SE blocks was set to 8.

**Inference**

Since the images were resampled to isotropic resolution, resulting in various sizes, and the input size for the segmentation model was fixed at 96 × 96 × 96, inference was performed using MONAI's sliding window inferer with an overlap of 0.5 and Gaussian importance weighting. For the biometry task, both the resampling size and the input size of the model were set to 128 × 128 × 128, thus eliminating the need for sliding window inference.

In the test phase, an ensemble of five models from the cross-validation process was utilized for both tasks. For the segmentation task, the softmax-activated network outputs from each model were averaged and then processed with the argmax function to extract segmentation predictions. These results from the segmentation task were consequently used for the biometry task. Based on the predicted segmentations, the same foreground dilation, masking, and cropping methods used during training were applied to the input images. The predicted heatmaps from each biometry model were then averaged, and the measurement coordinates were determined as the coordinates of the point with the maximum probability in the predicted heatmap.

**Implementation details**

The overall framework was implemented using custom source code based on PyTorch Lightning [6] and MONAI [7] on a workstation with an NVIDIA GeForce RTX 4090 GPU and 128 GB RAM. The tuned hyperparameters included types of data augmentation, training epochs, learning rates, the weight of auxiliary loss, and the scale factor for the dynamic standard deviation in heatmap generation. The detailed training protocols are presented in the following table.

|  | Task 1 | Task 2 |
|---|---|---|
| Data Augmentation | Rotation, scaling, translation, intensity shift, low resolution simulation | Rotation, scaling, low resolution simulation, coarse shuffle, coarse dropout |
| Network inialization | PyTorch default | PyTorch default |
| Batch size | 2 | 2 |
| Patch size | 96 × 96 × 96 | 128 × 128 × 128 |
| Total epochs | 1000 | 300 |
| Optimizer | Adam | Adam |
| Learning rate and schedule | 0.001, constant | 0.003, constant |
| Number of trainable parameters | 11.1M | 11.1M |
| Training time | 7 hours per fold | 2 hours per fold |

# Algorithm Description

1. **Team Information**

**Team Name:** lmrcmc

Team Members (include names and emails of all team members, add rows as necessary):

| Name | Email |
|---|---|
| Li Tianhong | tianhong1.li@cn.medical.canon |
| Yang Hong | hong2.yang@cn.medical.canon |
| Zhao Longfei | longfei1.zhao@cn.medical.canon |
| | |
| | |

Affiliations of each Team Member:
Canon Medical Systems (China) Co., Ltd

Max three (3) team members can be included in any publication resulting from this challenge.

- Would you like to be involved in any future publications? **(Yes/No)** If yes, which three team members are to be included?
  **Yes**
  1. _____Li Tianhong_____
  2. _____Yang Hong_____
  3. _____Zhao Longfei_____

- We will create a DockerHub where the dockers submitted to the FeTA Challenge will be stored. Do we have your permission to upload your docker? (**Yes/No**)
  **No**

- There will be a poster session as part of the FeTA Challenge in conjunction with the PIPPI workshop. Would you be interested in participating in the poster session? (**Yes/No**)
  **Yes**
- MICCAI 2024 will be an in-person event. Please state if you plan to attend in person (**Yes/No).**
  **Yes**

2. **Model Information**

    - GPU training was performed on:
      NVIDIA A800

1

- Software used incl. version
  Pytorch (2.0.1), cudatoolkit (11.7), nnUNetV2 (2.2)
- Model architecture
  Two models are trained.
  One model is trained based on nnUNet framework[1], adopt architecture by nnUNet default plan.
  Second model is trained based on SegVol[3].
- Number of layers
  nnUNet model: 5 stages of down sampling and 5 stages of up sampling
  SegVol model: Image Encoder: ViT, Text Encoder: CLIP, Mask Decoder: transformer with self-attention and cross-attention
- Convolution kernel size:
  3 x 3 x 3
- Initialization
  nnUNet model: "he" normal initialization (default setting in nnUNet)
  SegVol model: initialization with the pre-train weight.
- Optimizer
  nnUNet model: Stochastic gradient descent (SGD) with Nesterov momentum ($\mu$ = 0.99), (default setting in nnUNet)
  SegVol model: AdamW
- Cross-validation used?
  Both of two models: No.
- Number of epochs
  nnUNet model: 1000 epoch
  SegVol model: 230 epochs.
- Number of trainable parameters:
  nnUNet model: 31.2M
  SegVol model: 117.7M
- Learning Rate and schedule:
  nnUNet model: Initial learning rate = 0.0005, polynomial learning rate schedule (default setting in nnUNet)
  SegVol model: lr is 1e-4, weight decay is 1e-5
- Loss Function
  nnUNet model: CE + Dice loss (default setting in nnUNet)
  SegVol model: BCE + Binary Dice Losee
- Dimensionality of input/output (ie: 2D,3D, 2D+, etc.)
  nnUNet model: 3D/3D
  SegVol model: 3D/3D
- Batch Size
  nnUNet model: 2
  SegVol model: 8
- Preprocessing steps used (ie data normalization, creation of patches, etc.)
  nnUNet model:
    - Image intensity is normalized to [0, 1] by min-max normalization
    - Patch size = 128 x128 x128 (by nnUNet default plan)
    - Voxel spacing = 0.5039 x 0.5039 x 0.5039 (by nnUNet default plan)

  SegVol model:

    - Image intensity is normalized to [0,1] by min-max normalization
    - Patch size = 128 x 128 x 128

- Data Augmentation steps (ie – rotation, flipping, scaling, blur, noise, etc.)
  nnUNet model:
    - Apply rotation, flipping, blur, noise and scaling augmentation by nnUNet.
    - "Location-scale Augmentation" method from SLAug augmentation strategy [2] is adopted instead of contrast/brightness/gamma transform provided by nnUNet.
      "Location-scale Augmentation" include two types of augmentation:

Global Location-scale Augmentation (GLA), which increases the source-like images through global distribution shifting, and Local Location-scale Augmentation (LLA), which conducts class-specific augmentation to explore sufficiently diverse or even extreme appearance of unseen domain [2].

Augmented image is obtained by
$$I_{aug} = s\,GLA(I_{ori}) + (1-s)\,LLA(I_{ori})$$

In SLAug paper, $s$ is estimated voxel by voxel based on a saliency map.
In our implementation, we simply set $s$ = 0.75.

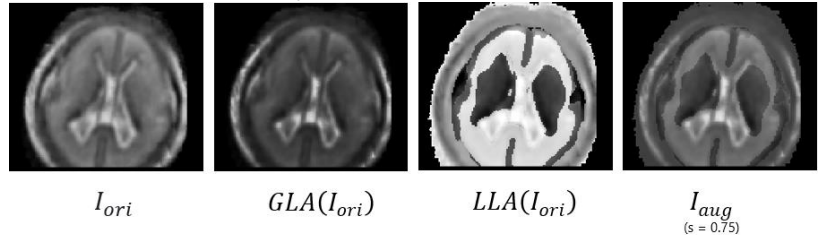

$I_{ori}$  $GLA(I_{ori})$  $LLA(I_{ori})$  $I_{aug}$ (s = 0.75)

SegVol model:
- Use flip, ScaleIntensity, ShiftIntensity, GibbsNoise, BiasField, KSpaceSpikeNoise and Affine augmentation during training.

- External dataset used? (allowed, but it needs to be publicly available
  Both of two models: No
- Framework (ie – MONAI, nnUNet, etc.)
  nnUNet and MONAI
- Number of models trained for final submission
  2
- Post-Processing Steps (ie – ensemble network, voting, label fusion)
  Average the predictions of two models

- Include relevant citations, as well as if existing code/software libraries/packages were used

[1] F. Isensee, P. F. Jaeger, S. A. A. Kohl, J. Petersen, and K. H. Maier-Hein, "nnU-Net: a self-configuring method for deep learning-based biomedical image segmentation," Nat Methods, vol. 18, no. 2, pp. 203–211, Feb. 2021. https://github.com/MIC-DKFZ/nnUNet

[2] Zixian Su, Kai Yao, Xi Yang, Qiufeng Wang, Jie Sun, Kaizhu Huang, "Rethinking data augmentation for single-source domain generalization in medical image segmentation," in Proceedings of the AAAI, vol. 37, no. 2, 2023, pp. 2366–2374. https://github.com/Kaiseem/SLAug

[3] Du Y, Bai F, Huang T, Zhao B. Segvol: Universal and interactive volumetric medical image segmentation. arXiv preprint arXiv:2311.13385. 2023 Nov 22.

- Which FeTA cases were included in the training and testing (ie – all cases, only pathological, only 1 institution, etc.)

  All cases

- Training/validation/testing data splits

  Train: 108, validation: 12

- Hyperparameter tuning performed

3

No

- Training time

    nnUNet model: 3 days

    SegVol model: 5 hr

- Abstract:

Fetal Tissue Annotation and Segmentation (FETA) towards the development of effective, domain-generalizable and reproducible methods for analyzing high resolution reconstructed MR images of the developing fetal brain from gestational week 21-36. It includes data from five different sites and magnetic fields including recent low-field systems.

The task is input 3D MRI T2w (256x256x256) image to algorithm, then the algorithm output 7 brain tissues classes. Our method bases on an U-Net model, a foundation model and ensembling the predictions of two models. The whole pipeline including pre-processing, data augmentation and post-processing(ensemble). The Dice result of ensemble result (82.3) has 1.3 ~ 2.1 improvement than only use U-Net (81.0) or foundation model (80.2), and better than ensemble result of 2 different architecture Unet model (about 81.2-81.4).

This is the first time to use foundation model in fetal brain segmentation and the result shows potential to be the complement of the U-Net.

# Algorithm Description Guidelines

**For a challenge submission to be considered complete, participants** in addition to submitting a docker container, **must submit the following information**.

This must be submitted by **August 12, 2024**.

1. **Team Information**

**Team Name:**         LIT

| Name | Email |
|---|---|
| Domen Preloznik | domen.preloznik@fe.uni-lj.si |
| Žiga Špiclin | ziga.spiclin@fe.uni-lj.si |
|  |  |
|  |  |
|  |  |

Affiliations of each Team Member:

Domen Preložnik: Faculty of Electrical Engineering, University of Ljubljana, SLO

Žiga Špiclin: Faculty of Electrical Engineering, University of Ljubljana, SLO

Max three (3) team members can be included in any publication resulting from this challenge.

- Would you like to be involved in any future publications? **(Yes/No)** If yes, which three team members are to be included?
    1. Domen Preložnik
    2. Žiga Špiclin
    3. 

- We will create a DockerHub where the dockers submitted to the FeTA Challenge will be stored. Do we have your permission to upload your docker? **Yes**

- There will be a poster session as part of the FeTA Challenge in conjunction with the PIPPI workshop. Would you be interested in participating in the poster session? **Yes**

- MICCAI 2024 will be an in-person event. Please state if you plan to attend in person **Yes.**

## 2. Model Information

If deep learning was used: Attention Unet and nnUNet

GPU training was performed on: 3x A100 (80gb)

Software used incl. version (i.e. Tensorflow, Pytorch, etc.): PyTorch, MIRTK / SVRTK

**nnUNet ResidualEncoderUNet for Brain Segmentation:**

- **Architecture:** Residual Encoder U-Net
- **Number of layers:** 6
- **Convolutional Kernel Size:** (3, 3, 3)
- **Initialization:** PyTorch default
- **Optimizer:** SGD
- **Cross-validation:** 6-fold
- **Learning rate and schedule:** initial 0.01, Poly LR Scheduler with 0.9 exponent decrease
- **Loss function:** Dice and CrossEntropy Loss
- **Dimensionality of input/output:** 3D
- **Batch Size:** 6
- **Preprocessing steps used:**
  - Image ROI extraction
  - Brain mask detection (Attention Unet)
  - Brain mask dilation and erosion
  - Image brain masking
- **Data Augmentation steps:**
  - Rotation
  - Scaling
  - Gaussian Noise
  - Gaussian Blur
  - Brightness Alteration
  - Contrast Adjustment
  - Low Resolution Simulation
  - Gamma Adjustment
  - Mirroring
- **External dataset used:** No
- **Framework :** nnUNet
- **Number of models trained for final submission:** 6
- **Post-Processing Steps:**
  - Joint prediction based on gaussian-weighted per-mirrored-patch result
  - Ensemble of 6 (per-fold) predictions
  - Transformation to patient space, with linear interpolation
  - Cropping / padding to match input image shape

- **Clearly state which aspects are original work (if any) or already existing work:**
    - Existing: MIRTK, SVRTK, nnUNet
    - Original: Preprocessing hyperparameter tuning, nnUNet configuration tuning, data split and per-fold configurations.
- **Include relevant citations, as well as if existing code/software libraries/packages were used:**
    - Uus, A. U., Hall, M., Payette, K., Hajnal, J. V., Deprez, M., Hutter, J., Rutherford, M. A., Story, L. (2023) Combined quantitative T2* map and structural T2-weighted tissue-specific analysis for fetal brain MRI: pilot automated pipeline. PIPPI MICCAI 2023 workshop (Accepted / in press)
    - Isensee, F., Jaeger, P. F., Kohl, S. A., Petersen, J., & Maier-Hein, K. H. (2021). nnU-Net: a self-configuring method for deep learning-based biomedical image segmentation. Nature methods, 18(2), 203-211.
- **Which FeTA cases were included in the training and testing (ie – all cases, only pathological, only 1 institution, etc.):**
    - All cases
- **Training/validation/testing data splits**
    - Training: 105 cases in a 6-fold split
    - Testing: 15 cases with equal distribution between pathological-neurotypical and institution
- **Hyperparameter tuning performed**
    - Epochs, batch size, patch size
- **Training time**
    - approx. 15h per fold in a multi-gpu setup
    - approx. 80-90h for all folds

# Algorithm Description Guidelines

**For a challenge submission to be considered complete, participants** in addition to submitting a docker container, **must submit the following information**.

This must be submitted by **August 12, 2024**.

### 1. Team Information

**Team Name:** mic-dkfz

Team Members (include names and emails of all team members, add rows as necessary):

| Name | Email |
|---|---|
| Maximilian Zenk | m.zenk@dkfz-heidelberg.de |
| Michael Baumgartner | m.baumgartner@dkfz-heidelberg.de |
| Klaus Maier-Hein | k.maier-hein@dkfz-heidelberg.de |
| | |
| | |

Affiliations of each Team Member:

German Cancer Research Center (DKFZ) Heidelberg, Division of Medical Image Computing, German

Medical Faculty Heidelberg, Heidelberg University, Heidelberg, Germany

Pattern Analysis and Learning Group, Department of Radiation Oncology, Heidelberg University Hospital, 69120 Heidelberg, Germany

Max three (3) team members can be included in any publication resulting from this challenge.

- Would you like to be involved in any future publications? **(Yes)** If yes, which three team members are to be included?
  1. Maximilian Zenk
  2. Michael Baumgartner
  3. Klaus Maier-Hein

- We will create a DockerHub where the dockers submitted to the FeTA Challenge will be stored. Do we have your permission to upload your docker? (**Yes**)

- There will be a poster session as part of the FeTA Challenge in conjunction with the PIPPI workshop. Would you be interested in participating in the poster session? (**No own poster**)

1

- MICCAI 2024 will be an in-person event. Please state if you plan to attend in person (**Yes/No).**

## 2. Model Information

If deep learning was used: yes

GPU training was performed on: different GPUs, including: A100, V100, RTX 2080

Software used incl. version (i.e. Tensorflow, Pytorch, etc.): Pytorch (version 2.3.1), nnunet (@commit 834b80f)

Please attach a description of your model highlighting the main features. This description must include the following details (unless the parameter is not applicable for your model):

- Model architecture
- Number of layers
- Convolution kernel size
- Initialization
- Optimizer
- Cross-validation used?
- Number of epochs
- Number of trainable parameters
- Learning Rate and schedule
- Loss Function
- Dimensionality of input/output (ie: 2D,3D, 2D+, etc.)
- Batch Size
- Preprocessing steps used (ie data normalization, creation of patches, etc.)
- Data Augmentation steps (ie – rotation, flipping, scaling, blur, noise, etc.)
- External dataset used? (allowed, but it needs to be publicly available
- Framework (ie – MONAI, nnUNet, etc.)
- Number of models trained for final submission
- Post-Processing Steps (ie – ensemble network, voting, label fusion)
- Clearly state which aspects are original work (if any) or already existing work
- Include relevant citations, as well as if existing code/software libraries/packages were used
- Which FeTA cases were included in the training and testing (ie – all cases, only pathological, only 1 institution, etc.)
- Training/validation/testing data splits
- Hyperparameter tuning performed
- Training time

Note: If a deep learning method was used, please provide the equivalent appropriate information as listed above.

# Team mic-dkfz

## Statement of original work

We tried different variations of nnUNet and selected the best-performing methods. To simulate out-of-distribution testing, we trained models only on Zurich data and evaluated on the Vienna cases. Hyperparameters we tried:

- Residual encoder for nnUNet
- Heavy data augmentation (DA5 trainer of nnUNet)
- Batch normalization
- Pretraining with MultiTalent [1] on a collection of publicly available datasets (MR and CT). A list of datasets can be found at the end of this document.

The final model was an ensemble of 3 networks, which was a tradeoff between which models performed best on in-distribution validation splits (model 1 & 2 below) and ood split (model 3):
1. Default nnUNet
2. Default nnUNet with batch size 4
3. Residual Encoder (L) nnUNet, fine-tuned from a model that was pretrained on a large set of publicly available datasets using MultiTalent

In the following, we report the most important characteristics for each model.

## Models 1/2

- Model architecture: U-Net
- Number of layers: 6 U-Net stages, each with 2 conv. layers
- Convolution kernel size: 3
- Initialization: Kaiming normal
- Optimizer: SGD + Momentum
- Cross-validation used? yes, for model selection
- Number of epochs: 250 * 1000 batches
- Number of trainable parameters: 88M
- Learning Rate and schedule: 1e-2, polynomial schedule
- Loss Function: Dice + CE
- Dimensionality of input/output (ie: 2D,3D, 2D+, etc.): 3D
- Batch Size: 2 for model 1, 4 for model 2
- Preprocessing steps used (ie data normalization, creation of patches, etc.): nnunet (z-score normalization, cropping, patches of size 128**3)
- Data Augmentation steps (ie – rotation, flipping, scaling, blur, noise, etc.): randomized; blur, gaussian noise, spatial (rotation, scaling, flipping), brightness, contrast, low-resolution simulation, gamma
- External dataset used? (allowed, but it needs to be publicly available): no

- Framework (ie – MONAI, nnUNet, etc.): nnUNet
- Post-Processing Steps (ie – ensemble network, voting, label fusion): ensembling
- Include relevant citations, as well as if existing code/software libraries/packages were used: nnUNet, pytorch, [1]
- Which FeTA cases were included in the training and testing (ie – all cases, only pathological, only 1 institution, etc.): all cases
- Training/validation/testing data splits: 5-fold CV and additional ood split (see statement of original work)
- Hyperparameter tuning performed: yes (see statement of original work)
- Training time: ~16h

## Models 3

- Model architecture: U-Net with Residual encoder
- Number of layers: 6
- Convolution kernel size: 3
- Initialization: Kaiming normal
- Optimizer: SGD + Momentum
- Cross-validation used? yes, for model selection
- Number of epochs: 250 * 1000 batches
- Number of trainable parameters: 383M
- Learning Rate and schedule: 1e-3, polynomial schedule
- Loss Function: Dice + CE
- Dimensionality of input/output (ie: 2D,3D, 2D+, etc.): 3D
- Batch Size: 2
- Preprocessing steps used (ie data normalization, creation of patches, etc.): nnunet (z-score normalization, cropping, patches of size 192**3)
- Data Augmentation steps (ie – rotation, flipping, scaling, blur, noise, etc.): randomized; blur, gaussian noise, spatial (rotation, scaling, flipping), brightness, contrast, low-resolution simulation, gamma, sharpening, blank rectangle
- External dataset used? (allowed, but it needs to be publicly available): yes, only publicly available
- Framework (ie – MONAI, nnUNet, etc.): nnUNet
- Post-Processing Steps (ie – ensemble network, voting, label fusion): ensembling
- Include relevant citations, as well as if existing code/software libraries/packages were used: nnUNet, pytorch, [1]
- Which FeTA cases were included in the training and testing (ie – all cases, only pathological, only 1 institution, etc.): all cases
- Training/validation/testing data splits: 5-fold CV and additional ood split (see statement of original work)
- Hyperparameter tuning performed: yes (see statement of original work)
- Training time: ~44h

# References

[1] Ulrich, Constantin, et al. "Multitalent: A multi-dataset approach to medical image segmentation." *International Conference on Medical Image Computing and Computer-Assisted Intervention*. Cham: Springer Nature Switzerland, 2023.

# List of datasets used during Pretraining

| Name | Images | Modality | Target | Link |
| --- | --- | --- | --- | --- |
| Decatlon Task 2 | 20 | MRI |  | http://medicaldecathlon.com/ |
| Decatlon Task 3 | 131 | CT | Liver, L. Tumor | http://medicaldecathlon.com/ |
| Decatlon Task 4 | 208 | MRI | Hippocampus | http://medicaldecathlon.com/ |
| Decatlon Task 5 | 32 | MRI | Ürpstate | http://medicaldecathlon.com/ |
| Decatlon Task 6 | 63 | CT | Lung Lesion | http://medicaldecathlon.com/ |
| Decatlon Task 7 | 281 | CT | Pancreas, P. Tumor | http://medicaldecathlon.com/ |
| Decatlon Task 8 | 303 | CT | Hepatic Vessel, H. Tumor | http://medicaldecathlon.com/ |
| Decatlon Task 9 | 41 | CT | Spleen | http://medicaldecathlon.com/ |
| Decatlon Task 10 | 126 | CT | Colon Tumor | http://medicaldecathlon.com/ |

| Name | Count | Modality | Target | URL |
|---|---|---|---|---|
| ISLES2015 | 28 | MRI | Stroke Lesion | http://www.isles-challenge.org/ISLES2015/ |
| BTCV | 30 | | 13 abdominal organs | https://www.synapse.org/Synapse:syn3193805/wiki/89480 |
| LIDC | 1010 | CT | Lung lesion | https://www.cancerimagingarchive.net/collection/lidc-idri/ |
| Promise12 | 50 | MRI | Prostate | https://zenodo.org/records/8026660 |
| ACDC | 200 | MRI | RV cavity, myocardium, LV cavity | https://www.creatis.insa-lyon.fr/Challenge/acdc/databases.html |
| ISBILesion2015 | 42 | MRI | MS Lesion | https://iacl.ece.jhu.edu/index.php/MSChallenge |
| CHAOS Task | 60 | MRI | Liver, Kidney (L&R), Spleen | https://zenodo.org/records/3431873 |
| BTCV 2 | 63 | CT | 9 abdominal organs | https://zenodo.org/records/1169361#.YiDLFnXMJFE |
| StructSeg Task1 | 50 | CT | 22 OAR Head & neck | https://structseg2019.grand-challenge.org |
| StructSeg Task2 | 50 | CT | Nasopharynx cancer | https://structseg2019.grand-challenge.org/Home/ |
| StructSeg Task3 | 50 | CT | 6 OAR Lung | https://structseg2019.grand-challenge.org/Home/ |
| StructSeg Task4 | 50 | CT | Lung Cancer | https://structseg2019.grand-challenge.org/Home/ |
| SegTHOR | 40 | CT | heart, aorta, trachea, esophagus | https://competitions.codalab.org/competitions/21145 |
| NIH-Pan | 82 | CT | Pancreas | https://wiki.cancerimagingarchive.net/display/Public/Pancreas-CT |

| Dataset | Count | Modality | Target | URL |
|---|---|---|---|---|
| VerSe2020 | 113 | CT | 28 Vertebrae | https://github.com/anjany/verse |
| M&Ms | 300 | MRI | left ventricle, right ventricle, left ventricular myocardium | https://www.ub.edu/mnms/ |
| ProstateX | 140 | MRI | Prostate lesion | https://www.aapm.org/GrandChallenge/PROSTATEx-2/ |
| RibSeg | 370 | CT | Rips | https://github.com/M3DV/RibSeg?tab=readme-ov-file |
| MSLES | 48 | MRI | MS Lesion | https://data.mendeley.com/datasets/8bctsm8jz7/1 |
| BrainMetShare | 84 | MRI | | https://aimi.stanford.edu/brainmetshare |
| CrossModa22 | 168 | MRI | vestibular schwannoma, cochlea | https://crossmoda2022.grand-challenge.org/ |
| Atlas22 | 524 | MRI | stroke lesion | https://atlas.grand-challenge.org/ |
| KiTs23 | 489 | CT | Kidneys, k. Tumors, Cysts | https://kits-challenge.org/kits23/ |
| AutoPet2 | 1014 | PET,CT | Lesions | https://autopet-ii.grand-challenge.org/ |
| AMOS | 360 | CT&MRI | 15 abdominal organs | https://amos22.grand-challenge.org/ |
| BraTs23 | 1251 | MRI | Glioblastoma | https://www.synapse.org/#!Synapse:syn25829067/wiki/610863 |
| AbdomenAtlas10 | 5195 | CT | 8 abdominal organs | https://github.com/MrGiovanni/AbdomenAtlas?tab=readme-ov-file |
| TotalSegmentator V2 | 1180 | CT | 117 classes of whole body | https://github.com/wasserth/TotalSegmentator |

| Name | Cases | Modality | Targets | Link |
|---|---|---|---|---|
| Hecktor2022 | 524 | PET,CT | nodal Gross Tumor Volumes, and nodal Gross Tumor Volumes (Head & Neck) | https://hecktor.grand-challenge.org/ |
| FLARE | 50 | CT | 13 abdominal organs | https://flare22.grand-challenge.org/ |
| SegRap | 120 | CT | 45 OARs (Head&Neck) | https://segrap2023.grand-challenge.org/ |
| SegA | 56 | CT | Aorta | https://multicenteraorta.grand-challenge.org/data/ |
| WORD | 120 | CT | 16 abdominal organs | https://github.com/HiLab-git/WORD |
| AbdomenCT1K | 996 | CT | Liver, Kidney, Spleen, pancreas | https://github.com/JunMa11/AbdomenCT-1K |
| DAP-ATLAS | 533 | CT | 142 classes of whole body | https://github.com/alexanderjaus/AtlasDataset |
| CTORG | 140 | CT | lung, brain, bones, liver, kidneys and bladder | https://www.cancerimagingarchive.net/collection/ct-org/ |
| HanSeg | 42 | CT | OAR (Head&Neck) | https://han-seg2023.grand-challenge.org/ |
| TopCow | 200 | CT&MRI | vessel components of CoW | https://topcow23.grand-challenge.org/ |

# Algorithm Description Guidelines

1. **Team Information**

   **Team Name:** paramahir_2023

   Team Members (include names and emails of all team members, add rows as necessary):

   | Name | Email |
   | --- | --- |
   | Ms. Param Ahir | ahirparam@gmail.com |
   | Dr. Mehul Parikh | mehulcparikh@ldce.ac.in |

   Affiliations of each Team Member:
   Gujrat Technological University, Gujarat, India

   L.D. College of Engineering, Gujarat, India

   Max three (3) team members can be included in any publication resulting from this challenge.

   - Would you like to be involved in any future publications? **(Yes/No)** If yes, which three team members are to be included? - Yes
     1. Ms. Param Ahir
     2. Dr. Mehul Parikh

   - We will create a DockerHub where the dockers submitted to the FeTA Challenge will be stored. Do we have your permission to upload your docker? (**Yes/No) - Yes**

   - There will be a poster session as part of the FeTA Challenge in conjunction with the PIPPI workshop. Would you be interested in participating in the poster session? (**Yes/No) - Yes**

   - MICCAI 2024 will be an in-person event. Please state if you plan to attend in person (**Yes/No). - No**

2. **Model Information**

   If deep learning was used: Yes

   GPU training was performed on: RTX 4090 (24 GB VRAM, 30 GB RAM)

   Software used incl. version (i.e. Tensorflow, Pytorch, etc.):

   - PyTorch 2.3 - PyTorch is used for defining and training the deep learning models, including the 3D UNet and custom BiometryModel.
   - MONAI: Version 1.3.2 - It's used for data handling, transformations, and model architecture like UNet.
   - SimpleITK: Version 2.3.1 SimpleITK is used for reading, processing, and writing medical images

### 3. Description of the Model

**1. Model Architecture**

The model consists of two components:

- A 3D UNet for segmentation tasks.
- A custom Biometry Model that incorporates a 3D UNet for feature extraction, followed by a fully connected regression head for predicting biometry values.

2. Number of Layers

- 3D UNet for Segmentation: The UNet has 5 levels of depth, with each level containing convolutional blocks with residual connections.
- Biometry Model: The UNet used in the biometry model has 3 levels of depth. The regression head consists of two fully connected layers.

3. Convolution Kernel Size: 3x3x3.

4. Initialization:

- Convolutional layers in the UNet models are initialized using Kaiming Normal initialization.
- Batch normalization layers, if used, have their weights initialized to 1 and biases to 0.

5. Optimizer: Adam Optimizer with a learning rate of 1e-4 was used for both segmentation and biometry models.

6. Cross-Validation: NA

7. Number of Epochs

- The segmentation model was trained for 1500 epochs.
- The biometry model was trained for 100 epochs.

8. Number of Trainable Parameters

Segmentation Model Trainable Parameters: 4941726

Biometry Model Trainable Parameters: 143963

9. Learning Rate and Schedule: A constant learning rate of 1e-4 was used, with no learning rate schedule.

10. Loss Function

- Segmentation Model: Dice Loss
- Biometry Model: Mean Squared Error (MSE) Loss

11. Dimensionality of Input/Output

- Both the input and output of the models are 3D (volumetric data).
- Input images are resized to a spatial dimension of 128x128x128 during preprocessing.

12. Batch Size: 1

13. Preprocessing Steps

    - Data Normalization: Intensity scaling was applied to the images.
    - Resampling: Images were resampled to have a uniform voxel spacing of 1.5x1.5x1.5 mm.
    - Reorientation: Images were reoriented to the RAS coordinate system.
    - Resizing: Images were resized to 128x128x128.

14. Data Augmentation Steps

    - Random Flipping: 50% probability of flipping along one axis.
    - Random Rotation: 50% probability of a 90-degree rotation.
    - Random Intensity Shifts: Intensity values were randomly shifted by an offset of 0.1 with a 50% probability.

15. External Dataset Used: No external datasets were used. Only the data provided in the FeTA challenge was used for training and testing.

16. Framework: MONAI framework

17. Number of Models Trained for Final Submission: 2

    - A segmentation model (3D UNet).
    - A biometry prediction model (custom UNet-based regression model).

18. Post-Processing Steps

    - Segmentation Output: Resized to match the original input image size using simple interpolation.
    - Biometry Output: Direct prediction without additional post-processing.
    - Landmarks: Not Implemented, only dummy code as a placeholder

19. Original Work vs. Existing Work

    - The core UNet architecture and certain transformations are based on existing methods and implementations, notably from the MONAI framework.
    - The combination of segmentation and biometry prediction in a unified pipeline, as well as data preprocessing and augmentation, are original work.

20. Citations

    - MONAI framework: Cardoso, M. J., Li, W., Brown, R., Ma, N., Kerfoot, E., Wang, Y., ... & Feng, A. (2022). Monai: An open-source framework for deep learning in healthcare. arXiv preprint arXiv:2211.02701.
    - UNet architecture: Ronneberger, O., Fischer, P., Brox, T. (2015). U-Net: Convolutional Networks for Biomedical Image Segmentation. In: Navab, N., Hornegger, J., Wells, W., Frangi, A. (eds) Medical Image Computing and Computer-Assisted Intervention – MICCAI 2015. MICCAI 2015. Lecture Notes in Computer Science(), vol 9351. Springer, Cham. https://doi.org/10.1007/978-3-319-24574-4_28 21.

21. FeTA Cases that were included:

    - All available cases from the FeTA dataset were included in segmentation training and testing, irrespective of pathology or institution.
    - Those subjects whose biometry data was missing were not considered during biometry model training.

22. Training/Validation/Testing Data Splits: 80/20

23. Hyperparameter Tuning: NA

24. Training Time:

    - Segmentation Model: 10 Hours
    - Biometry Model: 2 Hours

# Algorithm Description Guidelines

**For a challenge submission to be considered complete, participants** in addition to submitting a docker container, **must submit the following information**.

This must be submitted by **August 12, 2024**.

1. **Team Information**

**Team Name:** pasteurdbc

Team Members (include names and emails of all team members, add rows as necessary):

| Name | Email |
| --- | --- |
| Robin CREMESE[1,2] | Robin.cremese@pasteur.fr |
| Kein SAM[1,2] | kein.sam@etu.u-paris.fr |
| Fleur GAUDFERNAU[3] | fleur.gaudfernau@orange.fr |
| Jean-Baptiste MASSON[1,2] | jbmasson@pasteur.fr |
| Iwan Quemada[1,2] | iwan.quemada@pasteur.fr |

Affiliations of each Team Member:
(1) Institut Pasteur, Université Paris Cité, CNRS UMR 3571, Decision and Bayesian Computation, Paris, France
(2) Inria, Épiméthée, Paris, France
(3) Inria, HeKA, PariSantéCampus, Paris, France

Max three (3) team members can be included in any publication resulting from this challenge.

- Would you like to be involved in any future publications? **(Yes/No)** If yes, which three team members are to be included? Yes
  1. Robin CREMESE
  2. Kein SAM
  3. Fleur GAUDFERNAU

- We will create a DockerHub where the dockers submitted to the FeTA Challenge will be stored. Do we have your permission to upload your docker? **Yes**

- There will be a poster session as part of the FeTA Challenge in conjunction with the PIPPI workshop. Would you be interested in participating in the poster session? **Yes**

- MICCAI 2024 will be an in-person event. Please state if you plan to attend in person. **Yes, I will attend it in person**

1

## 2. Model Information

If deep learning was used: Yes, of course

GPU training was performed on: 5 NVIDIA A40 with 48Go VRAM each

Software used incl. version (i.e. Tensorflow, Pytorch, etc.): Pytorch (2.3.1)

Please attach a description of your model highlighting the main features. This description must include the following details (unless the parameter is not applicable for your model):

- Model architecture MedNeXt_L
- Number of layers: 162
- Convolution kernel size: 3
- Initialization: None
- Optimizer: AdamW(amsgrad=False,betas=(0.9,0.999),lr=0.001,weight_decay=3e-05)
- Cross-validation used? Yes, 5-cross validation
- Number of epochs: 200
- Number of trainable parameters: 61.8 M
- Learning Rate and schedule: 1e-3, no scheduler
- Loss Function: DiceCE
- Dimensionality of input/output (ie: 2D,3D, 2D+, etc.): 3D
- Batch Size: 6
- Preprocessing steps used (ie data normalization, creation of patches, etc.): cf. nnUnetV1
- Data Augmentation steps (ie – rotation, flipping, scaling, blur, noise, etc.):
  - RandomScaling(min=0.7, max=1.4)
  - RandomRoatation(min=-0.5, max=0.5)
  - RandomAdjustContrast(gamma_min=0.7, gamma_min=1.5))
  - RandFlip(all_axis)
- External dataset used? (allowed, but it needs to be publicly available) Yes
- Framework (ie – MONAI, nnUNet, etc.) : nnUNet
- Number of models trained for final submission: 5
- Post-Processing Steps (ie- ensemble network, voting, label fusion): ensemble voting
- Clearly state which aspects are original work (if any) or already existing work: Introduction of a new dataset + automatic biometry measurement
- Include relevant citations, as well as if existing code/software libraries/packages were used:

  Roy, S., Koehler, G., Ulrich, C., Baumgartner, M., Petersen, J., Isensee, F., Jaeger, P.F. & Maier-Hein, K. (2023). MedNeXt: Transformer-driven Scaling of ConvNets for Medical Image Segmentation. International Conference on Medical Image Computing and Computer-Assisted Intervention (MICCAI), 2023.

- Which FeTA cases were included in the training and testing (ie – all cases, only pathological, only 1 institution, etc.): all cases
- Training/validation/testing data splits: there were no testing dataset and training / validation split where made randomly across the two datasets on a 5-cross validation basis

- Hyperparameter tuning performed: No
- Training time: 3 days

Note: If a deep learning method was used, please provide the equivalent appropriate information as listed above.

# Algorithm Description Guidelines

**For a challenge submission to be considered complete, participants** in addition to submitting a docker container, **must submit the following information**.

This must be submitted by **August 12, 2024**.

1. **Team Information**

**Team Name**: unipd-sum-aug

Team Members (include names and emails of all team members, add rows as necessary):

| Name | Email |
| --- | --- |
| Marco Castellaro[1] | marco.castellaro@unipd.it, |
| Daniele Barbiero[1,2] | daniele.barbiero.3@phd.unipd.it |
| Tommaso Ciceri[1,3] | tommaso.ciceri@studenti.unipd.it |
| Alice Giubergia[1,3] | alice.giubergia@studenti.unipd.it |
| Simone Perra[1] | simone.perra@unipd.it |
| Marco Pinamonti[1] | marco.pinamonti.1@phd.unipd.it |
| Mario Severino[1] | mario.severino@phd.unipd.it |
| Valentina Visani[1] | valentina.visani@studenti.unipd.it |

Affiliations of each Team Member:
[1] Department of Information Engineering, University of Padova, Padova, Italy
[2] Institute for Photonics and Nanotechnologies, National Research Council, Padova, Italy
[3] Neuroimaging Unit, Scientific Institute, IRCCS Eugenio Medea, Bosisio Parini, Italy

Max three (3) team members can be included in any publication resulting from this challenge.

- Would you like to be involved in any future publications? **(Yes)** If yes, which three team members are to be included?
    1. Marco Castellaro
    2. Marco Pinamonti
    3. Valentina Visani

- We will create a DockerHub where the dockers submitted to the FeTA Challenge will be stored. Do we have your permission to upload your docker? (**Yes**)

- There will be a poster session as part of the FeTA Challenge in conjunction with the PIPPI workshop. Would you be interested in participating in the poster session? (**Yes**)

- MICCAI 2024 will be an in-person event. Please state if you plan to attend in person (**Yes**).

2. **Model Information**

We adopted the 2D Swin-UMamba architecture[1], exploiting the implementation of data augmentation strategies (TorchIO transforms and GIN techniques) to enhance the model's robustness. Pytorch, NNuNetv2 platform (https://github.com/MIC-DKFZ/nnUNet/) and Monai (https://monai.io/) were used to implement preprocessing, training and the Swin-UMamba model. We started modifiying the extisting Swin-UMamba pretrained on imageNet repository (https://github.com/JiarunLiu/Swin-UMamba).

Swin-UMamba combines features from two key architectures:

- Mamba: a newer state space model architecture showing promising performance on information-dense data such as language modeling[2].
- UNet: a widely used architecture for biomedical image segmentation tasks, known for its encoder-decoder structure that captures spatial information effectively[3].

The model utilizes pretraining on the ImageNet dataset. ImageNet is a large-scale dataset with millions of labeled images across thousands of categories[4].

Swin-UMamba is structured with an encoder-decoder architecture. The encoder consists of 5 layers arranged as follows:

- a *convolutional* layer with a 7x7 kernel, padding size of 3 and a stride size of 2
- a *patch embedding* layer with a 2x2 patch size
- three *patch merging* layers

All layers from the second to the fifth are augmented with VSS blocks[5], with the number of VSS blocks per layer being {2, 2, 9, 2}, respectively. The VSS block (Visual State Space block) is a key component of Swin-UMamba designed specifically for handling 2D image data. Unlike the traditional use of Mamba in 1D sequence modeling for natural language processing, the VSS block adapts Mamba's capabilities to effectively manage spatial information essential for vision tasks. The VSS blocks and patch merging layers are initialized using pretrained weights from VMamba-Tiny trained on ImageNet[6]. However, the patch embedding layer does not use pretrained weights due to differences in patch size and input channels.

The Swin-UMamba decoder adopts the same decoder architecture as U-Net by introducing two modifications: 1) an extra convolution block with a residual connection to process skip connection features, and 2) an additional segmentation head at each scale for deep supervision.

The Adam optimizer with weight decay = 0.05 (AdamW) optimizer is used for training the model.

Each fold of was trained for 1000 epochs and contains approximately 60 million trainable parameters. The initial learning rate is set to $1\times10^{-4}$, with a cosine annealing schedule to gradually reduce the learning rate. The loss function used is a combination of Dice loss and Cross-Entropy loss.

The input and output are both 3D images with one channel. A batch size of 130 was used during training that was conducted parallelly for each fold on a Nvidia A40 equipped with 48GB or VRAM.

Our contribution stands in the augmentation of the dataset and its integration in the Swin-UMamba pipeline. Briefly, data from different scanners were pair-wise co-registered, both intra-scanner and inter-scanner to augment the dataset. Images, and their segmentations, were linearly co-registered with affine and rigid transforms using the Advanced Nomalization Tools[7]. To reach the

best overlap between the brains, particularly when they belong to different datasets, the brain masks of the two subjects were used as initial moving transform. After performing image co-registration, a significant number of images are generated (more then 14000). A 5-fold cross-validation was employed to ensure the robustness and generalizability of the model. Briefly, for each scanner, subjects were ordered by gestational age and divided into four gestational age blocks: < 23 weeks, 23.1-27 weeks, 27.1-31 weeks, and > 31.1 weeks. From each block, 20% of subjects were selected for validation of each fold, ensuring a balance between healthy and pathological cases. The remaining 80% was used for training.

Various data augmentation techniques were employed, including:

- spatial transforms: rotations and scaling
- gaussian noise addition
- gaussian blur addition
- brightness alteration
- contrast changes
- simulate low resolution by down sampling and then up sampling
- gamma correction
- mirroring images

We then integrated TorchIO transforms into the Swin-Umamba pretrained data augmentation pipeline, enabling the simulation of MRI acquisition artifacts and variations from other scanner types. TorchIO is an open-source Python framework dedicated to the manipulation and transformation of medical images in deep learning applications. It provides a wide range of built-in transformations for data manipulation and augmentation, particularly useful in medical image segmentation tasks[8]. Among these transforms, we have included:

- *RandomMotion*: this transform simulates motion artifacts in MRI scans by applying random displacements to the image, mimicking patient movement during the scan.
- *RandomGhosting*: this transform introduces ghosting artifacts, which are common in MRI scans due to the signal instability between pulse cycle repetitions.
- *RandomBiasField*: this transform models intensity variations across the image, known as bias fields, which occur due to imperfections in the MRI scanner's magnetic field.

Moreover, GIN (Global intensity non-linear augmentation) technique for data augmentation was used. Inspired by the work of Ouyang et al.[9], GIN efficiently transforms image intensities and textures using shallow convolutional networks with random kernels and Leaky ReLU, producing diverse transformations and enhancing model robustness.

A total of 5 models were trained, corresponding to the 5 folds of cross-validation. The final prediction is an ensemble of these models by weighted mean of the probability map to determine the final label for each pixel. The training duration lasted approximately 10 days for each fold. Hyperparameter were optimized using the nnUnetv2 framework in which swin-umamba is implemented.

# Algorithm Description Guidelines

**For a challenge submission to be considered complete, participants** in addition to submitting a docker container, **must submit the following information**.

This must be submitted by **August 12, 2024**.

## 1. Team Information

**Team Name:** UPFetal24

Team Members (include names and emails of all team members, add rows as necessary):

| Name | Email |
|---|---|
| Simone Chiarella | simone.chiarella@studio.unibo.it |
| Gerard Martí-Juan | gerard.marti@upf.edu |
| Gemma Piella | gemma.piella@upf.edu |
| Oscar Càmara | oscar.camara@upf.edu |
| Miguel Angel González Ballester | ma.gonzalez@upf.edu |

Affiliations of each Team Member:
1: Università di Bologna, Bologna, Italy.
2: BCN MedTech, Department of Information and Communication Technologies, Universitat Pompeu Fabra, Barcelona, Spain
3: ICREA, Barcelona, Spain

Max three (3) team members can be included in any publication resulting from this challenge.

- Would you like to be involved in any future publications? **Yes** If yes, which three team members are to be included?
  1. Simone Chiarella
  2. Gerard Martí-Juan
  3. Miguel Ángel González Ballester

- We will create a DockerHub where the dockers submitted to the FeTA Challenge will be stored. Do we have your permission to upload your docker? **Yes**

- There will be a poster session as part of the FeTA Challenge in conjunction with the PIPPI workshop. Would you be interested in participating in the poster session? (**Yes/No).**

- MICCAI 2024 will be an in-person event. Please state if you plan to attend in person (**Yes/No).**

1

**2. Model Information**

If deep learning was used: yes, it was for the brain segmentation task

GPU training was performed on: NVIDIA QUADRO RTX 6000

Software used incl. version (i.e. Tensorflow, Pytorch, etc.): nnunetv2 = 2.5, pytorch = 2.3.0,

pytorch-cuda = 12.1, tqdm = 4.66.4, nibabel = 5.2.1, numpy = 1.26.4, scipy = 1.13.1,

batchgeneratorsv2 = 0.2, scikit-learn = 1.5.0, scikit-image = 0.22.0, pandas = 2.2.2,

tifffile = 2024.5.22, seaborn = 0.13.2, matplotlib = 3.9.0, imagecodecs = 2024.1.1,

requests = 2.31.0, python = 3.11.9

Please attach a description of your model highlighting the main features. This description must include the following details (unless the parameter is not applicable for your model):

- Model architecture
- Number of layers
- Initialization
- Cross-validation used?
- Number of epochs
- Number of trainable parameters
- Learning Rate and schedule
- Loss Function
- Dimensionality of input/output (ie: 2D,3D, 2D+, etc.)
- Batch Size
- Preprocessing steps used (ie data normalization, creation of patches, etc.)
- Data Augmentation steps (ie – rotation, flipping, scaling, blur, noise, etc.)
- External dataset used? (allowed, but it needs to be publicly available)
- Framework (ie – MONAI, nnUNet, etc.)
- Number of models trained for final submission
- Post-Processing Steps (ie – ensemble network, voting, label fusion)
- Clearly state which aspects are original work (if any) or already existing work
- Include relevant citations, as well as if existing code/software libraries/packages were used
- Which FeTA cases were included in the training and testing (ie – all cases, only pathological, only 1 institution, etc.)
- Training/validation/testing data splits
- Hyperparameter tuning performed
- Training time

**Generalizable fetal brain segmentation:**

The method we propose for brain segmentation is based on the open source nnU-Net by Isensee F. et al.. The version we applied is nnU-Net ResEncL, which makes use of a residual

encoder and which architecture can be found at this [link](link). We trained three models having the same architecture, and we ensembled them together to get the predictions.

The architecture involves 6 resolution stages for the encoder and the decoder, with a number of computational blocks of [1, 3, 4, 6, 6, 6] per stage, respectively. Three layers are present in each computational block:

- One convolution layer ([torch.nn.modules.conv.Conv3d](torch.nn.modules.conv.Conv3d)) with a kernel size of [3, 3, 3].
- One normalization layer ([torch.nn.InstanceNorm3d](torch.nn.InstanceNorm3d)).
- One non-linear activation function ([torch.nn.LeakyReLU](torch.nn.LeakyReLU)).

Between each encoder stage there is a downsampling layer, and between each decoder stage there is an upsampling layer. One convolution is performed between the output of each decoder stage and the skip connection from the corresponding stage in the encoder. The number of trainable parameters is 102354600.

For initialization, Kaiming uniform distribution ([torch.nn.init.kaiming_uniform](torch.nn.init.kaiming_uniform)) was used. Stochastic Gradient Descent ([torch.optim.SGD](torch.optim.SGD)) was used as an optimizer. The initial learning rate was set to 0.01, and it followed the Polynomial Learning Rate ([torch.optim.lr_scheduler.PolynomialLR](torch.optim.lr_scheduler.PolynomialLR)) schedule, ensuring an almost linear decrease to 0. Each model was trained on 1000 epochs; each epoch is defined as 250 training iterations, using a batch size of 2. As a loss function, [nnunetv2.training.loss.compound_losses.DC_and_CE_loss](nnunetv2.training.loss.compound_losses.DC_and_CE_loss) was used. It is a weighted sum of a Dice loss and [torch.nn.CrossEntropyLoss](torch.nn.CrossEntropyLoss). Dice loss optimizes the evaluation metric directly, but due to the patch based training, in practice merely approximates it. Combining the Dice loss with a cross-entropy loss improves training stability and segmentation accuracy.

The dimensionality of both input and output is 3. Train-test data split was 100/0 during the final training. No cross-validation was used for the final model. 5-fold CV was used when deciding which configurations to include in the final ensemble and which external datasets to include. To train our model, beside the datasets from Zurich and Vienna, three publicly available external datasets of fetal T2w images were used:

- 40 cases from dHCP atlas

    ([https://gin.g-node.org/kcl_cdb/fetal_brain_mri_atlas/src/master/cnn_cortex_probability](https://gin.g-node.org/kcl_cdb/fetal_brain_mri_atlas/src/master/cnn_cortex_probability))

- 11 cases from KCL 055T Fetal MRI atlas

    ([https://gin.g-node.org/kcl_cdb/055t_fetal_mri_atlases](https://gin.g-node.org/kcl_cdb/055t_fetal_mri_atlases))

- 15 cases from Spina Bifida Aperta Spatio-temporal Brain MRI Atlas

    ([https://www.synapse.org/Synapse:syn25887675/wiki/611424](https://www.synapse.org/Synapse:syn25887675/wiki/611424))

Pre-processing steps used are the ones implemented by default in nnU-Net:

- Cropping
- Usage of patches having sizes [192, 192, 192]

- Z-scoring normalization (subtract mean and standard deviation) separately for each train case

Data augmentation is where the three models we trained differentiate from each other.

- ❖ First model: default nnU-Net DA (training time ~84 h)
    - ➢ Rotation and scaling applied with a probability of 0.2 each
    - ➢ Gaussian noise added to each voxel in the sample independently, with a probability of 0.15
    - ➢ Gaussian blurring applied with a probability of 0.2 per sample
    - ➢ Brightness adjustment applied to each voxel intensities with a probability of 0.15
    - ➢ Contrast adjustment applied to each voxel intensities with a probability of 0.15
    - ➢ Simulation of low resolution applied with a probability of 0.25 per sample and 0.5 per associated modality
    - ➢ Gamma augmentation applied with a probability of 0.15
    - ➢ Gamma augmentation with image inversion applied with a probability of 0.15
    - ➢ Mirroring applied to all patches with a probability of 0.5 along all axes
- ❖ Second model: high-probability nnU-Net DA (training time ~95 h)
    - ➢ same as the first model but all probabilities are 0.8
    - ➢ Because of a mistake that was found too late during the training process, this second model was trained with 10 cases less from the Zurich center
- ❖ Third model: GIN-IPA + default nnU-Net DA (training time ~84 h)
    - ➢ GIN: apply Global Intensity Non-linear transformation (GIN) to the input image using a convolutional network with randomly sampled weights. This transformation introduces random variations in the textures and intensities of the input data.
    - ➢ IPA: the Hadamard products of two randomly GIN-augmented images with pseudo-correlation maps are combined, resulting in what is termed "Interventional Pseudo-correlation Augmentation" (IPA). This process creates an image that eliminates domain-specific spurious correlations.
    - ➢ Acknowledgments: github.com/cheng-01037/Causality-Medical-Image-Domain-Generalization

The predictions from the three models are ensembled together to get the final prediction, by making use of the dedicated nnU-Net script.

**Biometric estimation:**

We propose a multi-step process combining image registration, geometric calculations and various knowledge-based heuristics. First, the T2-weighted MRI images and their corresponding segmentations, obtained by the methodology described above, are registered to a standardized atlas space using the Medical Image Registration Toolkit (MIRTK) framework (https://mirtk.github.io/). The registration step includes rigid alignment and scaling. The atlas used is the atlas based on the subjects of the developing Human Connectome Project (https://gin.g-node.org/kcl_cdb/fetal_brain_mri_atlas/). The appropriate atlas was selected based on each subject's gestational age.

Following registration, we make the assumption that the brains are perfectly centered, in an orientation that tries to follow the one used by the clinicians when doing those measurements. Then, two key points were found for each biometric measurement:

- Length of Corpus Callosum (LCC): The points were assumed to be in the mid-sagittal plane. As we don't have information about the CC segmentation, we use the ventricle as a surrogate for the CC, assuming that the CC is situated alongside it. We focus on the higher half of the image to avoid including 3th and 4th ventricles into the measurement. Then, we identify the most anterior and posterior points of the ventricle, and we select those coordinates as the keypoints.
- Height of Vermis (HV): The Vermis is measured parallel to the brainstem in the mid-sagittal plane. To find it, what we do is to determine the angle of the brainstem with respect to the vertical by fitting a line to the region. We then rotate the cerebellum mask according to the found angle. Then, we just wind the maximum vertical extent of the rotated mask, and we transform the found coordinates back to the original image space.
- Brain Biparietal Diameter (bBIP): bBIP was measured in the axial plane. We first identified the slice with maximum brain width, and we measured the outer edge of the brain tissue from one side to the other, making sure that both keypoints were on the same horizontal line.
- Skull Biparietal Diameter (sBIP): Analogous to bBIP, but including both brain tissue and cerebrospinal fluid segmentation.
- Transverse Cerebellar Diameter (TCD): Measured in the coronal plane. We first identified the slice with maximum cerebellar width, and we measured the longest horizontal length on the cerebellum.

After getting the 10 keypoints, we do the inverse transformation back to the original space. We do two extra steps to avoid issues that arose while doing this inverse transformation:

- In order to make sure that the keypoints don't disappear when transforming to the original space, where normally the brain is smaller, we first do a dilation of the 10 keypoints. Then, we do the inverse transformation to the original space, and finally, we remove any extra voxels remaining, leaving only the center of each voxel "blob".
- Sometimes, the keypoints of bBIP and sBIP overlap (when there is low cerebrospinal fluid or the segmentation is not good). We account for this happening as an edge case and use the same keypoint for both measurements.

Finally, when the keypoints are defined in the original space, we compute the measurements using the provided get_dist() function and output both the measurements and the keypoints.

# Algorithm Description Guidelines

**For a challenge submission to be considered complete, participants** in addition to submitting a docker container, **must submit the following information**.

This must be submitted by **August 12, 2024**.

1. **Team Information**

**Team Name:** ViCOROB

Team Members (include names and emails of all team members, add rows as necessary):

| Name | Email |
| --- | --- |
| Rachika Elhassna Hamadache[1] | rachikaelhassna.hamadache@udg.edu |
| Amina Bouzid[1] | aminabouzid01@gmail.com |
| Ricardo Montoya del Ángel[1] | ricardo.montoya@udg.edu |
| Marawan Elbatel[2] | mkfmelbatel@ust.hk |
| Cansu Yalçın[1] | cansu.yalcin@udg.edu |
| Hadeel Awwad[1] | rg.hadil87@gmail.com |
| Dr. Adrià Casamitjana[1] | adria.casamitjana@udg.edu |
| Dr. Arnau Oliver[1] | arnau.oliver@udg.edu |
| Dr. Robert Martí Marly[1] | robert.marti@udg.edu |
| Dr. Xavier Lladó Bardera[1] | xavier.llado@udg.edu |
| | |

Affiliations of each Team Member:

[1] Research Institute of Computer Vision and Robotics (ViCOROB), Universitat de Girona, Spain

[2] Hong Kong University of Science and Technology (HKUST)

Max three (3) team members can be included in any publication resulting from this challenge.

- Would you like to be involved in any future publications? **(Yes)** If yes, which three team members are to be included?
    1. Rachika Elhassna Hamadache
    2. Amina Bouzid
    3. Dr. Xavier Lladó Bardera

- We will create a DockerHub where the dockers submitted to the FeTA Challenge will be stored. Do we have your permission to upload your docker? (**Yes)**

- There will be a poster session as part of the FeTA Challenge in conjunction with the PIPPI workshop. Would you be interested in participating in the poster session? (**Yes**)

- MICCAI 2024 will be an in-person event. Please state if you plan to attend in person (**Yes**).

## 2. Model Information

If deep learning was used: **Yes, the proposed solution is based on a Deep Learning approach.**

GPU training was performed on: **Three NVIDIA A30 GPUs of 24G each.**

Software used incl. version (i.e. Tensorflow, Pytorch, etc.): **Pytorch-based implementation (2.2.2+cu121).**

Please attach a description of your model highlighting the main features. This description must include the following details (unless the parameter is not applicable for your model):

- Model architecture
- Number of layers
- Convolution kernel size
- Initialization
- Optimizer
- Cross-validation used?
- Number of epochs
- Number of trainable parameters
- Learning Rate and schedule
- Loss Function
- Dimensionality of input/output (ie: 2D,3D, 2D+, etc.)
- Batch Size
- Preprocessing steps used (ie data normalization, creation of patches, etc.)
- Data Augmentation steps (ie – rotation, flipping, scaling, blur, noise, etc.)
- External dataset used? (allowed, but it needs to be publicly available
- Framework (ie – MONAI, nnUNet, etc.)
- Number of models trained for final submission
- Post-Processing Steps (ie – ensemble network, voting, label fusion)
- Clearly state which aspects are original work (if any) or already existing work
- Include relevant citations, as well as if existing code/software libraries/packages were used
- Which FeTA cases were included in the training and testing (ie – all cases, only pathological, only 1 institution, etc.)
- Training/validation/testing data splits
- Hyperparameter tuning performed
- Training time

Note: If a deep learning method was used, please provide the equivalent appropriate information as listed above.

2

# Algorithm description of the segmentation task of the Fetal Tissue Annotation and Segmentation challenge - FeTA 2024

Rachika E. Hamadache[1], Amina Bouzid[1], Ricardo Montoya[1], Marawan Elbatel[2], Adria Casamitjana[1], Arnau Oliver[1], Robert Marti[1], and Xavier Lladó[1]

[1]Research Institute of Computer Vision and Robotics (ViCOROB), Universitat de Girona, Catalonia, Spain
[2]Hong Kong University of Science and Technology (HKUST)

August 2024

This document provides a brief overview of the strategy used during our participation in the segmentation task of the FeTA 2024 challenge. Our approach is based on the nnUNet Deep Learning framework [Ise+21], which is adapted here for the multiclass segmentation of fetal brain tissues.

This work builds on existing research in medical image segmentation, with a focus on enhancing the domain generalization of the trained models:

- The nnUNet [Ise+21] framework, known for its state-of-the-art performance in segmentation tasks, including multiclass segmentation,

- SynthSeg-inspired [Bil+23] T2w image synthesizer, where the brain tissue label maps are used to generate T2w-looking images by sampling Gaussian distributions for each label based on the dataset's mean and standard deviation values, with added conditions to maintain realistic tissue contrast.

- The Sharpness-Aware Minimization (SAM) optimizer [For+20], which has recently been used on top of other optimizers to simultaneously minimize the loss' value and sharpness, thereby improving model generalization,

- Additional augmentations, including random bias field and motion artifacts implemented within the TorchIO library, and low-resolution simulation from the MONAI library.

These enhancements demonstrated promising results in our cross-validation experiments. In the following sections, further details on the models' architecture and training procedure are presented.

# 1  Model description

The proposed solution uses an ensemble of 3-fold cross-validation 3D full-resolution nnUNet models, each based on the PlainConvUNet architecture. These models feature six stages with $3 \times 3 \times 3$ convolution kernels and are initialized using He initialization, the default in nnUNet. The input to the models consists of 3D patches sized $128 \times 128 \times 128$, making a total of 31.2M trainable parameters.

# 2  Initial training

For training the models, the dataset (all cases) [Pay+21] was divided into three folds for cross-validation. Each time, 2 folds (80 original images) were used for training together with 600 synthetic images. Then, the remaining 40 original images were used for validation. The created images, inspired by SynthSeg's method and conditioned to resemble T2w images as previously described, were always the same across all folds, with five synthetic images generated for each original image (120×5). No external datasets were used in the proposed solution.

The standard nnUNet preprocessing steps were applied, including non-zero region cropping, z-score intensity normalization, voxel resampling (forced to $0.5 \times 0.5 \times 0.5$), and patch extraction with a size of $128 \times 128 \times 128$.

For data augmentation, the default nnUNet techniques were employed: rotation, scaling, Gaussian noise, Gaussian blur, brightness and contrast adjustments, low-resolution simulation, gamma correction, and mirroring. Additionally, online random bias field and motion artifacts were introduced using the TorchIO library.

The models were trained for 1000 epochs with a batch size of 2, using an initial learning rate of $1e^{-2}$ managed by the PolyLRScheduler scheduler. The default nnUNet loss function was used, which combines Dice and CrossEntropy losses.

As for the optimizer, the SAM optimizer built on top of SGD was used to enhance generalization during training. Besides that, no particular hyperparameter tuning was performed.

# 3  Fine-tuning

To better handle the low field MRI data in the hidden test set, the three trained cross-validation models were fine-tuned using the same data with additional low-resolution (LR) images. These images were carefully selected to prevent data leakage, with 10 used for training and 5 for validation in each fold. The LR images were created using the RandSimulateLowResolution function from MONAI and then resampled to a $0.8 \times 0.8 \times 0.8$ spacing to match the test set resolution.

The models were fine-tuned on this new dataset for 200 epochs, starting with an initial learning rate of $1e^{-4}$.

# 4  Post-processing

After an average of two days of training per fold for the original models, and half a day for the fine-tuned models, the final submitted solution was an ensemble of the three fine-tuned folds.

In the post-processing step, the predictions were first masked using the non-zero regions of the original image. Then, the largest and most centered component from the binary version of the prediction was retained and applied as a final mask to the multiclass prediction, yielding the final tissue segmentation result.

# 5 Available codes

- [nnUNet framework's github](#)
- [SAM optimizer](#)
- [SynthSeg-inspired image synthesizer - Pytorch implementation](#)
- [TorchIO augmentations](#)
- [MONAI low resolution simulation](#)

# Algorithm Description Guidelines

**For a challenge submission to be considered complete, participants** in addition to submitting a docker container, **must submit the following information**.

This must be submitted by **August 12, 2024**.

1. **Team Information**

**Team Name:** qd_neuroincyte

Team Members (include names and emails of all team members, add rows as necessary):

| Name | Email |
| --- | --- |
| Qi Zeng | qi.zeng@childrens.harvard.edu |
| Davood Karimi | davood.karimi@childrens.harvard.edu |
| | |
| | |
| | |

Affiliations of each Team Member:
Boston Children's Hospital, Harvard Medical School

Max three (3) team members can be included in any publication resulting from this challenge.

- Would you like to be involved in any future publications? **(Yes/No)** If yes, which three team members are to be included?
  1. Qi Zeng
  2. Davood Karimir
  3. 

- We will create a DockerHub where the dockers submitted to the FeTA Challenge will be stored. Do we have your permission to upload your docker? **(Yes/No)**
  **Yes**

- There will be a poster session as part of the FeTA Challenge in conjunction with the PIPPI workshop. Would you be interested in participating in the poster session? **(Yes/No)**
  **No**

- MICCAI 2024 will be an in-person event. Please state if you plan to attend in person (**Yes/No**).
  **Yes**

2. **Model Information**

If deep learning was used: Yes SwinUnetr

GPU training was performed on: A6000 Ada, RTX 4090

Software used incl. version (i.e. Tensorflow, Pytorch, etc.): Pytorch 2.2.2

Please attach a description of your model highlighting the main features. This description must include the following details (unless the parameter is not applicable for your model):

- Model architecture
- Number of layers
- Convolution kernel size
- Initialization
- Optimizer
- Cross-validation used?
- Number of epochs
- Number of trainable parameters
- Learning Rate and schedule
- Loss Function
- Dimensionality of input/output (ie: 2D,3D, 2D+, etc.)
- Batch Size
- Preprocessing steps used (ie data normalization, creation of patches, etc.)
- Data Augmentation steps (ie – rotation, flipping, scaling, blur, noise, etc.)
- External dataset used? (allowed, but it needs to be publicly available
- Framework (ie – MONAI, nnUNet, etc.)
- Number of models trained for final submission
- Post-Processing Steps (ie – ensemble network, voting, label fusion)
- Clearly state which aspects are original work (if any) or already existing work
- Include relevant citations, as well as if existing code/software libraries/packages were used
- Which FeTA cases were included in the training and testing (ie – all cases, only pathological, only 1 institution, etc.)
- Training/validation/testing data splits
- Hyperparameter tuning performed
- Training time

Note: If a deep learning method was used, please provide the equivalent appropriate information as listed above.

Method Description:

For this submission, our team mainly wants to validate the performance of the relatively new Swin transformer backbones in the tasks related to fetal brain MR image segmentation. All models that we developed were based on Swin UNETR framework [1] and the implementation was primarily based on the MONAI API [2].

For the first image segmentation task, we trained a single Swin UNETR model with the standard 4-stage multi-head attention encoder and 5-stage fully convolutional decoder network (FCN) with 48 base channels outputting seven plus one background channels as the segmentation label map. We trained this model with image patches of 128x128x128 pixels at the resolution of 0.5x0.5x0.5 mm³. For the output predictor, we used the SoftMax function, and the training back propagation was supervised with Dice as the loss function. At the training time, we first threshold the background with intensity >10 then normalize the image with the intensity standard deviation. For each training case, we use a sliding window to randomly crop 128x128x128 patches, with additional argumentation of flipping, rotation and translation. The model was first trained with the Zürich dataset for 2000 epochs and then tuned with the Vienna dataset for 1000 epochs. At test time, we argument the sliding window patches with a 3D shift grid of [0,32,64,128] with tracked flipping. All augmented patches were loaded to the model to predict label tissue maps. For post processing, we use label fusion to fuse all prediction results to generate the final tissue map.

In our testing, we noticed that our model had conservative performance for the Vienna dataset, as the images kept the surrounding tissue. To further improve the performance, we trained a smaller brain mask network for the Vienna dataset to clean up the segmentation results. The model setting is very similar to our main segmentation network with a reduced base channel size to 24, outputting a single binary map of brain tissue regions including the external cerebrospinal fluid. This model was trained with Vienna dataset for 1000 epochs, and it is only used to clean up the segmentation results of Vienna dataset.

In the biometry estimation task, our technique was inspired by key point detection method previously presented in [3], where a deep FCN was trained to predict potential heat maps of measurement landmarks for distance metric estimation. Originally, we had attempted to train a Swin-UNETR to predict heat maps of all 10 landmarks for the five brain biometry measurements, with both MR image and the segmentation map as the inputs. However, our initial validation results showed that this approach was hard to generalize for the Zürich dataset. Thus, we converted out models to predict landmark heat maps only using the segmentation maps, this is to ensure that the models can correctly respond to the variation of the brain tissue structure while assuming we can obtain reliable segmentation maps from the input images.

For this submission, we train four separate models to carry out the five biometric measurements. Each model is based on a Swin-UNETR with 24 base channels taking 128x128x128 binary label map as the input, and the number of the output channels are set based on the required landmarks for each or grouped measurements: 1. For the length of the corpus callosum (LCC) the segmentation map of ventricles are used as input, and the mode outputted two channels of the two landmarks; 2. For the height of the vermis (HV) and the transverse cerebellar diameter (TCD_cor), the label map of cerebellum was used as input for estimating the four landmarks for completing the measurements; and 3. For the brain biparietal diameter (bBIP_ax) and the skull biparietal diameter (sBIP_ax), two networks are train separately to generate two landmarks for each measurements, where the segmentation map input of bBIP_ax network include all brain tissue classes except the external cerebrospinal fluid, and the input of sBIP_ax network takes the

3

full brain label mask as the input. At the training stage, the ground truth of each case was prepared by converting the transformation corrected landmarks into a 10-channel heat map, where each landmark takes one channel. To ensure the gradient could be properly initialized for effective network training, the landmark points were enlarged into a spherical disc with scaled Gaussian kernel with $\sigma = 3$ mm. For prediction, the Sigmoid function was used, and the Dice score was used as the training loss function. For all cases, we preprocess the label map with center cropping and resampling to make sure we have unified input size of 128x128x128 with a spatial resolution of 1x1x1 mm³. For post processing, outputs were first threshold and converted into a binary label map. Then point coordinate of the activated pixels were extracted for center of mass estimation. K-nearest-neighbor clustering algorithm [4] was utilized to reject further distanced sparse outliers. Then the estimated landmark center point pairs were utilized to compute the Euclidean distances as the results. All models were first developed and trained with the Zürich dataset for 1000 epochs. As we only have limited time (lease than a week) with the Vienna dataset, we only had time to tune the model with Vienna data for 500 epochs. Thus, we do not expect to have good performance with the Vienna dataset.

For more details of our methods, please refer to the following listed report based off the method description questionnaire:

- Model architecture
  Swin UNETR [1]

- Number of layers
  Default Swin UNETR settings were utilized. All models have a 4-stage transformer encoder and 5-stage FCN decoder. For all encoder stages, the number of base Swin transformer layer is [2,2,2,2] and the number of heads for the transformer were [3,6,12,24] respectively. For the decoder, the standard res-net block with Unet base FCN layers was used to enrich features. For up-sampling, x2 stride convolution was added. The segmentation model has a base feature channel size of 48. The biometry networks and the Vienna dataset image mask network all have a base feature channel size of 24.

- Convolution kernel size
  In all convolution layers, the base 3x3x3 kernel was used

- Initialization
  Default normal distribution initialization was used for all layers

- Optimizer
  Adam

- Cross-validation used
  For this submission, cross-validation was only used for initial segmentation model parameter tunning, where the Zurich dataset was used with a training vs validation ratio of 70/10.

- Number of epochs
  Segmentation:
  Data from University Children's Hospital Zürich was trained with 2000 epoch.
  Data from General Hospital Vienna/Medical University of Vienna was trained for 1000 epoch.

Biometry:
All models were initially trained with data from University Children's Hospital Zürich for 1000 epoch, data from General Hospital Vienna/Medical University of Vienna was used to further fine tune the models for 500 epochs.

Mask network of Vienna data:
All samples of the Vienna data were used to training model for 1000 epoch

- Number of trainable parameters
Segmentation network:
62187002 (1 input channel with 8 SoftMax output channels)

Biometry Networks:
length of the corpus callosum (LCC) – 15703004
(1 input channel with 2 Sigmoid output channels)
Height of the vermis (HV) + transverse cerebellar diameter (TCD_cor) – 15703054
(1 input channel with 4 Sigmoid output channels)
Brain biparietal diameter (bBIP_ax) – 15703004
(1 input channel with 2 Sigmoid output channels)
Skull biparietal diameter (sBIP_ax) – 15703004
(1 input channel with 2 Sigmoid output channels)

Image mask network for Vienna dataset – 15703004
(1 input channel with 2 SoftMax output channels)

- Learning Rate and schedule
Fixed learning rate 1e-4 with the Adam optimizer

- Loss Function
Dice

- Dimensionality of input/output (ie: 2D,3D, 2D+, etc.)
Segmentation/masking: 3D 128x128x128 center cropped image
Biometry: 3D 128x128x128 center cropped segmentation map

- Batch Size
Batch size stays as 1

- Preprocessing steps used (ie data normalization, creation of patches, etc.)
Intensity thresholding and standard deviation normalization
Segmentation training & testing:
128x128x128 patches with random sliding window on centered images

Biometry training & testing:
Center cropped 128x128x128 segmentation label maps:
- length of the corpus callosum (LCC) – Ventricles (4)
- Height of the vermis (HV) + transverse cerebellar diameter (TCD_cor) – Cerebellum (5)

- Brain biparietal diameter (bBIP_ax) – All segmentation label classes except the external cerebrospinal fluid
- Skull biparietal diameter (sBIP_ax)
  - All tissue label fused into one brain mask
- In training, landmark labels in the image space was converted into a spherical disc Gaussian heatmap with $\sigma$ = 3 mm.
- To obtain the biometry measurement, each out channel of the biometry network outputs was first threshold and converted into a binary label map. Then point coordinate of the activated pixels were extracted for center of mass estimation. K-nearest-neighbor clustering algorithm was utilized to reject further distanced sparse outliers [4]. Then the estimated landmark center point pairs were utilized to compute the biometry distances.

- Data Augmentation steps (ie – rotation, flipping, scaling, blur, noise, etc.)
  Random sliding window, flipping, 1% gaussian noise, rigid rotation of ± 25° around all axes, random shifting ± 5 mm along all axes

- External dataset used?
  No additional data was used

- Framework (ie – MONAI, nnUNet, etc.)
  Our Swin UNETR implementation was base on the MONAI Framework [2]

- Number of models trained for final submission
  For this submission we trained five models in total, which included 1 for segmentation, 4 for biometry and 1 for Vienna data image masking

- Post-Processing Steps (ie – ensemble network, voting, label fusion)
  For segmentation, we use label fusion to combine segmentation generated from all 128x128x128 patches in each case
  Results for the Vienna dataset is further cleaned with the mask network

- Clearly state which aspects are original work (if any) or already existing work
  Our submission has limited novelty on methodology. However, in existing literature, we only found few methods on end-to-end learning based 3D biometry estimation with fetal brain data [3].

- Which FeTA cases were included in the training and testing (ie – all cases, only pathological, only 1 institution, etc.)
  Data with from both centers were used. However, we had extremely limited time (less than a week) with the Vienna dataset due to late data transfer approval.

- Training/validation/testing data splits
  For the final submission, all training data is used to train the model.

- Hyperparameter tuning performed
  Learning rate was tuned with initial cross-validation

- Training time
  Overall, we spent around a week training the model.

## Algorithm Description Guidelines

**For a challenge submission to be considered complete, participants** in addition to submitting a docker container, **must submit the following information**.

This must be submitted by **August 12, 2024**.

1. **Team Information**

**Team Name:   CeSNE-DiGAIR**

Team Members (include names and emails of all team members, add rows as necessary):

| Name | Email |
|---|---|
| Ciceri Tommaso[1] | tommaso.ciceri@lanostrafamiglia.it |
| Di Stefano Marina[1] | marina.distefano@lanostrafamiglia.it |
| Frigerio Giulia[2] | giulia.frigerio@lanostrafamiglia.it |
| Longari Giorgio[3] | giorgio.longari@unimib.it |
| Maccarone Francesca[1,3] | francesca.maccarone@lanostrafamiglia.it |
| Melzi Simone[3] | simone.melzi@unimib.it |
| Peruzzo Denis[1] | denis.peruzzo@lanostrafamiglia.it |
| Prudentino Rocco[1] | rocco.prudentino@lanostrafamiglia.it |
| Rizzato Gloria[2] | gloria.rizzato@lanostrafamiglia.it |

Affiliations of each Team Member:

1. Neuroimaging Unit, Scientific Institute IRCCS E. Medea, Bosisio Parini, Italy

2. Neuroradiology Unit, Scientific Institute IRCCS E. Medea, Bosisio Parini, Italy

3. Department of Informatics, Systems and Communication. University of Milano Bicocca, Milan, Italy

Max three (3) team members can be included in any publication resulting from this challenge.

- Would you like to be involved in any future publications? **(Yes)** If yes, which three team members are to be included?
    1. ___Denis Peruzzo___
    2. ___Tommaso Ciceri___
    3. ___Giorgio Longari___

- We will create a DockerHub where the dockers submitted to the FeTA Challenge will be stored. Do we have your permission to upload your docker? (**Yes**)

1

- There will be a poster session as part of the FeTA Challenge in conjunction with the PIPPI workshop. Would you be interested in participating in the poster session? (**No**)

- MICCAI 2024 will be an in-person event. Please state if you plan to attend in person (**No**).

2. **Model Information**

If deep learning was used: Yes

GPU training was performed on: NVIDIA GeForce GTX 1080 Ti (12Gb RAM)

Software used incl. version (i.e. Tensorflow, Pytorch, etc.): based on the contained template provided for the FeTA Challenge, we included:

- PyTorch Version 2.4.0

- Python 3.10.12

- Python libraries: SimpleITK 2.3.1; pandas 2.1.4; antspyx 0.4.2; monai 1.3.2; nibabel 5.2.1; nilearn 0.10.4; numpy 1.24.4; lightning 2.3.0; lightning-utilities 0.11.6; pytorch-lightning 2.3.0; fslpy 3.21.0;

- Other software: ANTs (ver 2.5.3); miniconda3; BOUNTI (from the public available docker)

Model description:

*Task 1: tissue segmentation*

- Data preprocessing: T2w images were skull stripped by applying a binary brain mask derived from BOUNTI [1]. Each masked T2w image was registered to a publicly available atlas [2] through affine transformation reaching a pre-defined resolution of 0.5x0.5x0.5 mm$^3$; the same transform matrix was applied to the label maps of the training set using a nearest neighborhood interpolation. Final images (T2w/labels) were cropped to [176, 224, 176] matrix size and T2w intensity was normalized.

- Data augmentation: Deformable (SyN) registrations were performed between couples of scans from the preprocessed training dataset. The couples were selected combining neurotypical and pathological samples, until a sufficient number of registered samples was obtained (i.e. 600 T2w/label images).

- Segmentation model description: a 3DUNet-based architecture from MONAI framework [3] was used to perform the segmentation task. The adopted UNet comprises 5 layers (channels: 16, 32, 64, 128, 256) each having an encode and decode path with a skipping connection between them. In the encoding phase, data are downsampled by strided convolutions (2 strides for each of middle layers), while in the decoding phase, data are upsampled using strided transpose convolutions. A default kernel size of 3x3x3 was utilized. To train the model, 3D patches of dimension 96x96x96 were given as input. This leads to 4815745 trainable parameters.

- Segmentation model training strategy: model training was performed using Adam optimizer, starting from randomly initialized weights and with learning rate of 1e-4 and batch size of 2. No cross-validation strategy was applied, but we considered 10% samples (65 samples) for validation during the hyper-parameter tuning stage (loss function: Dice Loss). We trained the model until it reached a performance plateau (i.e. 600 epochs) and we selected the best one. Within training data augmentation was also performed using both spatial and intensity transforms [4]. Spatial transformations included dropout, rotation, mirroring and zoom, while intensity transformations involved Gaussian noise, Gaussian blur, random bias field, low resolution simulation, contrast adjustment and scaling. The total training time was 75 hours.

- Segmentation post-processing: we implemented a two steps procedure:

  1. segmentation accuracy enhancement was performedby removing incorrect predictions through a denoising autoencoder. We utilized the default 3D AutoEncoder network from the MONAI framework with 6 layers (3 encoding + 3 deconding, with 16, 32, 64 channels), a convolutional kernel size of 3x3x3, providing a model with 145 325 trainable parameters. The Adam optimizer (learing rate = 1e-4; loss function: Dice Lss, batch size 1; patch size 96x96x96) was used to train the model for 500 epochs. The training set was built using as input the model segmentation and the manual segmentation as gound truth. Data augmentation was also included by artificially corrupting the manual segmentations to simulate different scenarios (e.g. region dilatation, dropout, label swapping along the region boundaries, etc). The total training time was 6h.

  2. Finally, the inverse transform was applied to the predicted labels to project them to to the subject's original space.

*Task 2: Biometric measure extraction* (based on the output from task 1)

- Data preprocessing: data for biometric measure extraction were derived from the tissue segmentation performed in the previous task. In particular, input images were the label maps at the resolution of 0.5x0.5x0.5 mm$^3$; (matrix size: 176x224x176). Images (label maps) were resized to a shape of 128x128x128 to be used as input for the model.

- Data augmentation: No further data augmentation has been applied with respect to task 1.

- Model description: A 3D Convolutional Neural Network (CNN) model was implemented to predict the coordinates of the keypoints. The adopted CNN comprises four convolutional layers to extract relevant patterns from the segmentation mask in input, two max pooling layers were used in the last two layers to downsample the feature maps and summarize information, and a fully connected head to perform the regression on the keypoints. A default kernel of size (3x3x3) was utilized.

- Training strategy: model training was performed using Adam optimizer, starting from randomly initialized weights and with learning rate of 5e-5 and batch size of 16. A scheduler was employed to reduce the learning rate by a factor of 10 when the training

loss did not decrease once in 5 epochs. No cross-validation strategy was applied, but we considered 10% samples (10 samples) for validation during the hyper-parameter tuning stage (loss function: L1). We trained the model until it reached a performance plateau (i.e. 30 epochs) and we selected the best one. The total training time was 20 minutes.

- <u>Segmentation post-processing:</u> Following the same approach utilized in [5], a local optimization approach was applied to each pair of points predicted by the model. Exploiting the segmentation, we maximized the distance of the two predicted points in a region, along the line passing between them, and selecting as final position just voxels that belong to the prediction mask. Finally, the inverse transform was applied to the predicted keypoint to project them to the subject's original space.

# Appendix A3. FeTA 2024 segmentation results by site

Table A3: Segmentation results for FeTA 2024 by site, presented as mean ± standard deviation.

| Team | Site | Dice | HD95 | Volume Similarity | Euler diff. |
|---|---|---|---|---|---|
| **cemrg_feta** | CHUV | 0.819 +- 0.079 | 2.239 +- 1.641 | 0.888 +- 0.091 | 73.271 +- 157.842 |
| | KCL | 0.868 +- 0.048 | 1.482 +- 0.575 | 0.944 +- 0.050 | 8.729 +- 17.511 |
| | KISPI | 0.775 +- 0.178 | 3.756 +- 12.339 | 0.875 +- 0.175 | 17.150 +- 47.150 |
| | UCSF | 0.835 +- 0.064 | 4.180 +- 10.689 | 0.941 +- 0.056 | 21.732 +- 48.451 |
| | VIEN | 0.834 +- 0.079 | 1.846 +- 1.174 | 0.947 +- 0.056 | 38.200 +- 76.002 |
| **cesne-digair** | CHUV | 0.830 +- 0.060 | 2.234 +- 1.397 | 0.927 +- 0.058 | 29.104 +- 51.562 |
| | KCL | 0.856 +- 0.052 | 1.694 +- 0.524 | 0.950 +- 0.041 | 6.257 +- 13.210 |
| | KISPI | 0.776 +- 0.154 | 2.954 +- 2.863 | 0.890 +- 0.138 | 9.211 +- 17.841 |
| | UCSF | 0.823 +- 0.069 | 2.126 +- 1.307 | 0.942 +- 0.050 | 14.571 +- 25.900 |
| | VIEN | 0.814 +- 0.087 | 2.266 +- 1.687 | 0.947 +- 0.050 | 38.132 +- 93.476 |
| **falcons** | CHUV | 0.828 +- 0.063 | 2.054 +- 1.271 | 0.922 +- 0.056 | 82.214 +- 166.732 |
| | KCL | 0.832 +- 0.061 | 2.076 +- 1.080 | 0.906 +- 0.065 | 17.450 +- 33.388 |
| | KISPI | 0.763 +- 0.151 | 2.890 +- 2.490 | 0.887 +- 0.139 | 35.557 +- 80.026 |
| | UCSF | 0.430 +- 0.253 | 21.320 +- 14.077 | 0.602 +- 0.281 | 192.339 +- 318.110 |
| | VIEN | 0.389 +- 0.275 | 22.377 +- 17.046 | 0.580 +- 0.296 | 134.443 +- 158.795 |
| **feta_sigma** | CHUV | 0.823 +- 0.075 | 2.248 +- 1.659 | 0.892 +- 0.085 | 68.571 +- 143.222 |
| | KCL | 0.871 +- 0.049 | 1.469 +- 0.576 | 0.944 +- 0.051 | 10.521 +- 23.896 |
| | KISPI | 0.772 +- 0.178 | 3.311 +- 4.349 | 0.867 +- 0.177 | 14.814 +- 36.054 |
| | UCSF | 0.835 +- 0.064 | 2.427 +- 3.941 | 0.937 +- 0.059 | 21.279 +- 47.773 |
| | VIEN | 0.835 +- 0.084 | 2.216 +- 3.005 | 0.943 +- 0.062 | 32.771 +- 65.733 |
| **hilab** | CHUV | 0.813 +- 0.079 | 2.311 +- 1.464 | 0.878 +- 0.083 | 66.004 +- 147.774 |
| | KCL | 0.851 +- 0.058 | 1.765 +- 0.823 | 0.926 +- 0.066 | 5.807 +- 11.417 |
| | KISPI | 0.775 +- 0.176 | 2.937 +- 3.457 | 0.878 +- 0.169 | 13.029 +- 30.945 |
| | UCSF | 0.828 +- 0.068 | 2.677 +- 3.871 | 0.939 +- 0.053 | 24.786 +- 62.808 |
| | VIEN | 0.831 +- 0.081 | 2.145 +- 2.654 | 0.941 +- 0.062 | 28.832 +- 61.805 |
| **jwcrad** | CHUV | 0.741 +- 0.126 | 4.495 +- 4.241 | 0.840 +- 0.156 | 60.668 +- 119.543 |
| | KCL | 0.845 +- 0.060 | 2.101 +- 2.275 | 0.942 +- 0.059 | 10.857 +- 25.192 |
| | KISPI | 0.780 +- 0.154 | 3.001 +- 3.122 | 0.892 +- 0.142 | 15.054 +- 63.731 |
| | UCSF | 0.775 +- 0.129 | 4.139 +- 12.272 | 0.908 +- 0.132 | 22.382 +- 72.242 |
| | VIEN | 0.743 +- 0.173 | 3.375 +- 3.568 | 0.876 +- 0.174 | 30.314 +- 48.812 |
| **lit** | CHUV | 0.817 +- 0.074 | 2.397 +- 1.632 | 0.892 +- 0.078 | 73.400 +- 150.808 |
| | KCL | 0.867 +- 0.049 | 1.478 +- 0.511 | 0.950 +- 0.041 | 7.957 +- 16.179 |
| | KISPI | 0.761 +- 0.180 | 3.400 +- 4.158 | 0.866 +- 0.176 | 17.854 +- 49.090 |
| | UCSF | 0.812 +- 0.075 | 2.139 +- 1.852 | 0.931 +- 0.062 | 33.982 +- 59.647 |
| | VIEN | 0.810 +- 0.093 | 2.086 +- 2.610 | 0.937 +- 0.059 | 51.168 +- 93.347 |
| **lmrcmc** | CHUV | 0.814 +- 0.080 | 2.406 +- 1.409 | 0.881 +- 0.084 | 58.054 +- 126.355 |
| | KCL | 0.860 +- 0.052 | 1.611 +- 0.728 | 0.936 +- 0.058 | 9.429 +- 21.316 |
| | KISPI | 0.774 +- 0.171 | 3.013 +- 3.697 | 0.881 +- 0.168 | 14.675 +- 29.453 |
| | UCSF | 0.798 +- 0.084 | 4.399 +- 6.528 | 0.933 +- 0.066 | 38.914 +- 66.956 |
| | VIEN | 0.805 +- 0.093 | 3.682 +- 8.311 | 0.945 +- 0.057 | 31.021 +- 50.330 |
| **mic-dkfz-feta24** | CHUV | 0.824 +- 0.074 | 2.125 +- 1.520 | 0.889 +- 0.083 | 83.104 +- 190.117 |
| | KCL | 0.872 +- 0.046 | 1.456 +- 0.540 | 0.950 +- 0.048 | 8.471 +- 17.095 |
| | KISPI | 0.778 +- 0.176 | 3.667 +- 12.334 | 0.878 +- 0.175 | 24.968 +- 98.618 |
| | UCSF | 0.849 +- 0.060 | 1.648 +- 0.929 | 0.943 +- 0.059 | 20.679 +- 48.929 |
| | VIEN | 0.839 +- 0.077 | 1.842 +- 1.361 | 0.947 +- 0.055 | 34.443 +- 70.139 |
| **paramahir_2023** | CHUV | 0.019 +- 0.016 | 67.984 +- 9.712 | 0.257 +- 0.191 | 836.771 +- 514.176 |
| | KCL | 0.105 +- 0.114 | 40.203 +- 38.827 | 0.665 +- 0.278 | 1053.286 +- 1194.848 |
| | KISPI | 0.036 +- 0.069 | 134.320 +- 80.816 | 0.220 +- 0.303 | 3243.929 +- 2176.151 |
| | UCSF | 0.042 +- 0.044 | 69.115 +- 33.709 | 0.392 +- 0.246 | 891.479 +- 1040.067 |
| | VIEN | 0.029 +- 0.040 | 71.887 +- 14.464 | 0.317 +- 0.263 | 875.496 +- 803.573 |
| **pasteurdbc** | CHUV | 0.816 +- 0.080 | 2.272 +- 1.574 | 0.879 +- 0.087 | 98.582 +- 221.114 |
| | KCL | 0.870 +- 0.048 | 1.494 +- 0.557 | 0.953 +- 0.043 | 9.714 +- 19.600 |
| | KISPI | 0.770 +- 0.178 | 3.247 +- 4.094 | 0.868 +- 0.177 | 24.864 +- 82.402 |
| | UCSF | 0.820 +- 0.076 | 2.826 +- 3.592 | 0.920 +- 0.075 | 21.893 +- 49.927 |
| | VIEN | 0.835 +- 0.081 | 2.040 +- 1.813 | 0.946 +- 0.060 | 36.650 +- 71.768 |
| **qd_neuroincyte** | CHUV | 0.762 +- 0.127 | 3.973 +- 3.116 | 0.877 +- 0.125 | 27.993 +- 45.480 |
| | KCL | 0.800 +- 0.071 | 15.895 +- 15.467 | 0.896 +- 0.073 | 22.821 +- 24.303 |
| | KISPI | 0.769 +- 0.174 | 3.436 +- 4.185 | 0.878 +- 0.170 | 19.061 +- 41.108 |
| | UCSF | 0.444 +- 0.244 | 22.082 +- 26.679 | 0.648 +- 0.288 | 62.807 +- 133.521 |
| | VIEN | 0.689 +- 0.194 | 9.546 +- 9.233 | 0.869 +- 0.132 | 33.057 +- 49.988 |
| **unipd-sum-aug** | CHUV | 0.802 +- 0.084 | 2.517 +- 1.798 | 0.871 +- 0.093 | 85.850 +- 191.883 |
| | KCL | 0.863 +- 0.047 | 1.553 +- 0.587 | 0.950 +- 0.041 | 13.136 +- 22.792 |
| | KISPI | 0.762 +- 0.184 | 3.169 +- 3.723 | 0.868 +- 0.180 | 33.868 +- 108.254 |
| | UCSF | 0.827 +- 0.069 | 2.199 +- 2.283 | 0.934 +- 0.059 | 29.396 +- 62.257 |
| | VIEN | 0.826 +- 0.077 | 1.835 +- 0.987 | 0.945 +- 0.054 | 54.325 +- 108.289 |
| **upfetal24** | CHUV | 0.816 +- 0.080 | 2.296 +- 1.703 | 0.882 +- 0.091 | 89.425 +- 204.043 |
| | KCL | 0.844 +- 0.061 | 2.391 +- 1.490 | 0.931 +- 0.072 | 31.729 +- 45.318 |
| | KISPI | 0.776 +- 0.175 | 3.055 +- 3.537 | 0.876 +- 0.171 | 19.861 +- 44.504 |
| | UCSF | 0.840 +- 0.060 | 2.452 +- 5.282 | 0.940 +- 0.056 | 20.836 +- 51.070 |
| | VIEN | 0.837 +- 0.078 | 1.855 +- 1.405 | 0.945 +- 0.056 | 33.868 +- 68.319 |
| **vicorob** | CHUV | 0.831 +- 0.070 | 1.969 +- 1.266 | 0.899 +- 0.077 | 93.789 +- 206.532 |
| | KCL | 0.869 +- 0.051 | 1.499 +- 0.659 | 0.947 +- 0.051 | 9.986 +- 19.663 |
| | KISPI | 0.782 +- 0.170 | 2.982 +- 3.709 | 0.881 +- 0.169 | 19.579 +- 71.933 |
| | UCSF | 0.830 +- 0.067 | 2.276 +- 1.965 | 0.937 +- 0.056 | 29.343 +- 68.239 |

# Appendix A4. FeTA 2024 segmentation results by label

Table A4: Segmentation results for FeTA 2024 by label, presented as mean ± standard deviation.

| Team | Label | Dice | HD95 | Volume Similarity | Euler diff. |
|---|---|---|---|---|---|
| **cemrg_feta** | BS | 0.773 ± 0.109 | 4.066 ± 4.480 | 0.864 ± 0.128 | 0.561 ± 0.749 |
| | CBM | 0.869 ± 0.121 | 3.097 ± 15.680 | 0.933 ± 0.124 | 0.178 ± 0.718 |
| | CSF | 0.805 ± 0.126 | 3.090 ± 6.242 | 0.922 ± 0.126 | 79.172 ± 77.460 |
| | GM | 0.748 ± 0.075 | 1.885 ± 4.882 | 0.921 ± 0.062 | 144.156 ± 183.114 |
| | SGM | 0.801 ± 0.109 | 3.417 ± 4.473 | 0.863 ± 0.124 | 1.056 ± 1.166 |
| | VM | 0.864 ± 0.061 | 2.028 ± 6.517 | 0.951 ± 0.046 | 2.389 ± 2.444 |
| | WM | 0.892 ± 0.038 | 2.269 ± 5.881 | 0.961 ± 0.032 | 13.161 ± 14.716 |
| **cesne-digair** | BS | 0.790 ± 0.091 | 3.665 ± 2.187 | 0.899 ± 0.091 | 0.206 ± 0.525 |
| | CBM | 0.868 ± 0.088 | 1.466 ± 0.716 | 0.945 ± 0.083 | 0.367 ± 1.624 |
| | CSF | 0.794 ± 0.108 | 2.726 ± 2.860 | 0.937 ± 0.080 | 49.006 ± 35.876 |
| | GM | 0.730 ± 0.077 | 1.557 ± 0.786 | 0.935 ± 0.043 | 87.644 ± 109.380 |
| | SGM | 0.798 ± 0.105 | 3.061 ± 1.881 | 0.879 ± 0.125 | 0.833 ± 0.647 |
| | VM | 0.848 ± 0.066 | 1.713 ± 1.228 | 0.946 ± 0.045 | 1.578 ± 1.468 |
| | WM | 0.883 ± 0.043 | 2.030 ± 0.835 | 0.964 ± 0.030 | 6.817 ± 9.445 |
| **falcons** | BS | 0.519 ± 0.310 | 11.287 ± 10.191 | 0.688 ± 0.288 | 28.778 ± 41.150 |
| | CBM | 0.570 ± 0.361 | 19.103 ± 22.806 | 0.721 ± 0.288 | 42.172 ± 94.277 |
| | CSF | 0.608 ± 0.244 | 12.110 ± 11.915 | 0.719 ± 0.265 | 175.044 ± 316.827 |
| | GM | 0.604 ± 0.177 | 10.279 ± 11.330 | 0.845 ± 0.145 | 289.128 ± 272.968 |
| | SGM | 0.542 ± 0.324 | 9.231 ± 16.580 | 0.617 ± 0.334 | 17.350 ± 25.692 |
| | VM | 0.762 ± 0.158 | 5.269 ± 5.874 | 0.863 ± 0.144 | 44.506 ± 74.144 |
| | WM | 0.791 ± 0.145 | 10.000 ± 11.203 | 0.903 ± 0.116 | 108.122 ± 164.206 |
| **feta_sigma** | BS | 0.776 ± 0.104 | 4.263 ± 4.436 | 0.865 ± 0.118 | 0.483 ± 0.751 |
| | CBM | 0.867 ± 0.124 | 2.042 ± 3.452 | 0.925 ± 0.128 | 0.244 ± 0.759 |
| | CSF | 0.802 ± 0.132 | 3.058 ± 5.381 | 0.911 ± 0.136 | 83.772 ± 73.078 |
| | GM | 0.750 ± 0.076 | 1.438 ± 0.766 | 0.922 ± 0.063 | 121.089 ± 165.583 |
| | SGM | 0.800 ± 0.112 | 3.196 ± 2.267 | 0.860 ± 0.128 | 1.083 ± 1.143 |
| | VM | 0.869 ± 0.057 | 1.293 ± 0.696 | 0.951 ± 0.040 | 2.261 ± 2.175 |
| | WM | 0.893 ± 0.038 | 1.722 ± 0.536 | 0.962 ± 0.032 | 13.039 ± 12.755 |
| **hilab** | BS | 0.772 ± 0.098 | 4.155 ± 3.437 | 0.874 ± 0.108 | 0.350 ± 0.523 |
| | CBM | 0.866 ± 0.121 | 1.931 ± 3.170 | 0.924 ± 0.123 | 0.067 ± 0.310 |
| | CSF | 0.802 ± 0.121 | 2.506 ± 3.554 | 0.916 ± 0.125 | 79.889 ± 89.962 |
| | GM | 0.740 ± 0.086 | 1.492 ± 0.743 | 0.917 ± 0.068 | 114.972 ± 170.136 |
| | SGM | 0.790 ± 0.107 | 3.513 ± 3.317 | 0.860 ± 0.124 | 0.750 ± 0.838 |
| | VM | 0.853 ± 0.071 | 1.672 ± 2.266 | 0.931 ± 0.067 | 2.439 ± 2.077 |
| | WM | 0.889 ± 0.039 | 1.768 ± 0.550 | 0.955 ± 0.039 | 12.394 ± 13.574 |
| **jwcrad** | BS | 0.676 ± 0.175 | 5.253 ± 3.670 | 0.779 ± 0.212 | 7.417 ± 74.544 |
| | CBM | 0.799 ± 0.153 | 3.189 ± 3.409 | 0.878 ± 0.164 | 2.989 ± 9.602 |
| | CSF | 0.774 ± 0.121 | 2.693 ± 3.000 | 0.926 ± 0.093 | 62.778 ± 55.944 |
| | GM | 0.705 ± 0.088 | 1.931 ± 1.834 | 0.908 ± 0.085 | 105.406 ± 138.418 |
| | SGM | 0.737 ± 0.155 | 6.690 ± 15.354 | 0.827 ± 0.176 | 9.600 ± 74.649 |
| | VM | 0.829 ± 0.098 | 2.808 ± 3.244 | 0.931 ± 0.088 | 4.700 ± 6.119 |
| | WM | 0.866 ± 0.081 | 2.420 ± 2.004 | 0.950 ± 0.064 | 15.317 ± 16.491 |
| **lit** | BS | 0.770 ± 0.103 | 4.159 ± 2.815 | 0.872 ± 0.119 | 0.511 ± 0.639 |
| | CBM | 0.866 ± 0.120 | 1.787 ± 3.629 | 0.935 ± 0.118 | 0.083 ± 0.394 |
| | CSF | 0.782 ± 0.131 | 2.615 ± 3.683 | 0.917 ± 0.123 | 78.561 ± 69.369 |
| | GM | 0.718 ± 0.081 | 1.652 ± 1.190 | 0.916 ± 0.067 | 184.161 ± 166.182 |

|   |     |                   |                      |                   |                        |
|---|-----|-------------------|----------------------|-------------------|------------------------|
|   | SGM | $0.793 \pm 0.116$ | $3.250 \pm 2.395$    | $0.862 \pm 0.136$ | $0.933 \pm 1.122$      |
|   | VM  | $0.848 \pm 0.063$ | $1.487 \pm 1.074$    | $0.924 \pm 0.060$ | $2.361 \pm 2.044$      |
|   | WM  | $0.878 \pm 0.054$ | $1.790 \pm 0.867$    | $0.951 \pm 0.036$ | $13.983 \pm 21.532$    |
|   | BS  | $0.761 \pm 0.112$ | $4.559 \pm 5.464$    | $0.871 \pm 0.116$ | $1.333 \pm 2.422$      |
|   | CBM | $0.855 \pm 0.108$ | $3.499 \pm 7.590$    | $0.927 \pm 0.111$ | $3.850 \pm 11.014$     |
|   | CSF | $0.799 \pm 0.120$ | $2.689 \pm 5.592$    | $0.932 \pm 0.115$ | $52.617 \pm 48.115$    |
| **lmrcmc** | GM  | $0.713 \pm 0.076$ | $2.053 \pm 2.993$    | $0.910 \pm 0.077$ | $124.883 \pm 138.935$  |
|   | SGM | $0.782 \pm 0.116$ | $4.384 \pm 3.510$    | $0.852 \pm 0.132$ | $3.722 \pm 8.246$      |
|   | VM  | $0.846 \pm 0.077$ | $2.597 \pm 6.345$    | $0.945 \pm 0.053$ | $5.006 \pm 7.943$      |
|   | WM  | $0.876 \pm 0.044$ | $2.474 \pm 4.290$    | $0.953 \pm 0.036$ | $37.844 \pm 72.378$    |
|   | BS  | $0.792 \pm 0.102$ | $3.430 \pm 2.265$    | $0.877 \pm 0.116$ | $6.250 \pm 74.563$     |
|   | CBM | $0.875 \pm 0.121$ | $2.452 \pm 14.827$   | $0.936 \pm 0.124$ | $5.589 \pm 74.609$     |
|   | CSF | $0.811 \pm 0.123$ | $2.492 \pm 3.960$    | $0.921 \pm 0.124$ | $76.667 \pm 79.789$    |
| **mic-dkfz-feta24** | GM  | $0.754 \pm 0.078$ | $1.342 \pm 0.664$    | $0.919 \pm 0.068$ | $153.933 \pm 222.976$  |
|   | SGM | $0.803 \pm 0.112$ | $2.981 \pm 1.915$    | $0.859 \pm 0.128$ | $1.278 \pm 1.303$      |
|   | VM  | $0.867 \pm 0.060$ | $1.275 \pm 0.655$    | $0.953 \pm 0.046$ | $2.144 \pm 2.387$      |
|   | WM  | $0.895 \pm 0.036$ | $1.598 \pm 0.464$    | $0.961 \pm 0.032$ | $14.583 \pm 15.262$    |
|   | BS  | $0.002 \pm 0.004$ | $80.114 \pm 52.805$  | $0.439 \pm 0.300$ | $1315.700 \pm 1572.594$ |
|   | CBM | $0.003 \pm 0.011$ | $98.476 \pm 46.129$  | $0.471 \pm 0.319$ | $1182.661 \pm 1613.381$ |
|   | CSF | $0.057 \pm 0.068$ | $75.774 \pm 55.679$  | $0.364 \pm 0.283$ | $1668.756 \pm 1510.524$ |
| **paramahir_2023** | GM  | $0.040 \pm 0.038$ | $77.447 \pm 54.473$  | $0.284 \pm 0.275$ | $1239.539 \pm 1637.509$ |
|   | SGM | $0.034 \pm 0.039$ | $81.118 \pm 51.993$  | $0.238 \pm 0.230$ | $1424.194 \pm 1523.328$ |
|   | VM  | $0.049 \pm 0.056$ | $74.136 \pm 54.399$  | $0.328 \pm 0.273$ | $2102.550 \pm 1503.443$ |
|   | WM  | $0.092 \pm 0.105$ | $78.237 \pm 53.946$  | $0.238 \pm 0.238$ | $982.206 \pm 1700.717$  |
|   | BS  | $0.771 \pm 0.110$ | $4.197 \pm 2.852$    | $0.863 \pm 0.124$ | $5.939 \pm 74.584$     |
|   | CBM | $0.867 \pm 0.109$ | $1.809 \pm 1.441$    | $0.925 \pm 0.112$ | $0.217 \pm 1.135$      |
|   | CSF | $0.804 \pm 0.132$ | $2.797 \pm 4.533$    | $0.918 \pm 0.136$ | $75.700 \pm 75.225$    |
| **pasteurdbc** | GM  | $0.735 \pm 0.082$ | $1.517 \pm 0.790$    | $0.909 \pm 0.079$ | $184.989 \pm 256.428$  |
|   | SGM | $0.795 \pm 0.116$ | $3.224 \pm 2.146$    | $0.850 \pm 0.132$ | $1.072 \pm 1.237$      |
|   | VM  | $0.862 \pm 0.064$ | $1.333 \pm 0.902$    | $0.945 \pm 0.055$ | $2.056 \pm 1.902$      |
|   | WM  | $0.884 \pm 0.050$ | $2.439 \pm 3.815$    | $0.951 \pm 0.048$ | $20.678 \pm 21.555$    |
|   | BS  | $0.564 \pm 0.322$ | $16.358 \pm 33.573$  | $0.679 \pm 0.343$ | $31.933 \pm 164.379$   |
|   | CBM | $0.699 \pm 0.287$ | $12.927 \pm 18.047$  | $0.813 \pm 0.238$ | $6.317 \pm 11.709$     |
|   | CSF | $0.642 \pm 0.164$ | $7.771 \pm 4.291$    | $0.826 \pm 0.140$ | $50.983 \pm 42.466$    |
| **qd_neuroincyte** | GM  | $0.604 \pm 0.156$ | $6.012 \pm 5.081$    | $0.882 \pm 0.104$ | $85.433 \pm 61.912$    |
|   | SGM | $0.693 \pm 0.171$ | $10.169 \pm 9.490$   | $0.791 \pm 0.174$ | $13.939 \pm 25.437$    |
|   | VM  | $0.765 \pm 0.161$ | $11.839 \pm 11.728$  | $0.894 \pm 0.104$ | $26.578 \pm 37.313$    |
|   | WM  | $0.797 \pm 0.120$ | $8.013 \pm 6.939$    | $0.902 \pm 0.106$ | $24.883 \pm 26.451$    |
|   | BS  | $0.762 \pm 0.119$ | $4.033 \pm 2.908$    | $0.857 \pm 0.133$ | $6.178 \pm 74.569$     |
|   | CBM | $0.857 \pm 0.102$ | $2.000 \pm 1.808$    | $0.929 \pm 0.107$ | $0.278 \pm 0.702$      |
|   | CSF | $0.798 \pm 0.138$ | $2.282 \pm 3.628$    | $0.917 \pm 0.139$ | $85.600 \pm 90.364$    |
| **unipd-sum-aug** | GM  | $0.726 \pm 0.086$ | $1.499 \pm 0.758$    | $0.910 \pm 0.078$ | $215.856 \pm 226.509$  |
|   | SGM | $0.794 \pm 0.111$ | $3.363 \pm 2.449$    | $0.863 \pm 0.132$ | $1.222 \pm 1.194$      |
|   | VM  | $0.854 \pm 0.067$ | $1.443 \pm 0.958$    | $0.937 \pm 0.055$ | $2.400 \pm 2.590$      |
|   | WM  | $0.885 \pm 0.038$ | $1.708 \pm 0.486$    | $0.954 \pm 0.038$ | $15.144 \pm 16.221$    |
|   | BS  | $0.780 \pm 0.113$ | $4.028 \pm 2.620$    | $0.871 \pm 0.132$ | $1.611 \pm 3.007$      |
|   | CBM | $0.872 \pm 0.108$ | $1.529 \pm 1.116$    | $0.935 \pm 0.109$ | $0.350 \pm 1.339$      |
|   | CSF | $0.810 \pm 0.121$ | $2.556 \pm 4.235$    | $0.921 \pm 0.122$ | $84.133 \pm 80.372$    |
| **upfetal24** | GM  | $0.742 \pm 0.077$ | $1.401 \pm 0.708$    | $0.917 \pm 0.074$ | $164.483 \pm 233.666$  |
|   | SGM | $0.790 \pm 0.108$ | $3.652 \pm 4.141$    | $0.850 \pm 0.128$ | $1.611 \pm 3.941$      |

|  |     |                   |                   |                   |                       |
| --- | --- | --- | --- | --- | --- |
|  | VM  | 0.861 ± 0.064     | 1.636 ± 3.494     | 0.942 ± 0.051     | 3.061 ± 4.046         |
|  | WM  | 0.889 ± 0.035     | 2.082 ± 3.293     | 0.954 ± 0.036     | 24.522 ± 35.431       |
| **vicorob** | BS  | 0.788 ± 0.104     | 3.744 ± 2.639     | 0.886 ± 0.109     | 6.372 ± 74.563        |
|  | CBM | 0.873 ± 0.101     | 1.616 ± 1.192     | 0.932 ± 0.106     | 0.167 ± 0.780         |
|  | CSF | 0.807 ± 0.124     | 2.370 ± 3.839     | 0.919 ± 0.127     | 97.094 ± 91.240       |
|  | GM  | 0.745 ± 0.076     | 1.415 ± 0.774     | 0.919 ± 0.070     | 169.622 ± 238.677     |
|  | SGM | 0.801 ± 0.112     | 3.079 ± 2.062     | 0.868 ± 0.131     | 0.928 ± 1.041         |
|  | VM  | 0.868 ± 0.059     | 1.405 ± 1.703     | 0.956 ± 0.042     | 2.706 ± 2.315         |
|  | WM  | 0.891 ± 0.039     | 1.677 ± 0.504     | 0.957 ± 0.035     | 12.161 ± 12.048       |

# Appendix A5. FeTA 2024 segmentation results by pathology

Table A5: Segmentation results for FeTA 2024 by pathology, presented as mean ± standard deviation.

| Team | Pathology | Dice | HD95 | Volume Similarity | Euler diff. |
|---|---|---|---|---|---|
| **cemrg_feta** | Neurotypical | 0.838 ± 0.078 | 1.949 ± 1.685 | 0.923 ± 0.083 | 36.990 ± 104.012 |
| | Pathological | 0.808 ± 0.127 | 3.596 ± 10.477 | 0.910 ± 0.122 | 32.150 ± 79.136 |
| **cesne-digair** | Neurotypical | 0.834 ± 0.072 | 2.037 ± 1.401 | 0.939 ± 0.063 | 17.255 ± 36.003 |
| | Pathological | 0.801 ± 0.114 | 2.557 ± 2.138 | 0.921 ± 0.095 | 24.059 ± 65.344 |
| **falcons** | Neurotypical | 0.695 ± 0.233 | 9.431 ± 15.061 | 0.819 ± 0.212 | 89.800 ± 155.566 |
| | Pathological | 0.571 ± 0.297 | 12.417 ± 13.435 | 0.720 ± 0.285 | 110.080 ± 229.547 |
| **feta_sigma** | Neurotypical | 0.838 ± 0.079 | 1.944 ± 1.509 | 0.922 ± 0.082 | 35.382 ± 95.466 |
| | Pathological | 0.809 ± 0.128 | 2.846 ± 4.160 | 0.907 ± 0.125 | 28.568 ± 69.351 |
| **hilab** | Neurotypical | 0.832 ± 0.079 | 2.029 ± 1.465 | 0.917 ± 0.080 | 33.253 ± 97.439 |
| | Pathological | 0.802 ± 0.126 | 2.781 ± 3.629 | 0.906 ± 0.120 | 27.445 ± 72.466 |
| **jwcrad** | Neurotypical | 0.796 ± 0.110 | 3.051 ± 3.409 | 0.901 ± 0.121 | 33.539 ± 88.625 |
| | Pathological | 0.746 ± 0.163 | 4.013 ± 8.390 | 0.872 ± 0.166 | 26.496 ± 68.179 |
| **lit** | Neurotypical | 0.830 ± 0.080 | 1.971 ± 1.471 | 0.922 ± 0.079 | 40.859 ± 102.670 |
| | Pathological | 0.789 ± 0.132 | 2.751 ± 3.317 | 0.902 ± 0.122 | 39.423 ± 85.891 |
| **lmrcmc** | Neurotypical | 0.822 ± 0.086 | 2.884 ± 5.872 | 0.921 ± 0.082 | 31.589 ± 78.737 |
| | Pathological | 0.790 ± 0.126 | 3.432 ± 4.936 | 0.906 ± 0.118 | 33.745 ± 71.625 |
| **mic-dkfz-feta24** | Neurotypical | 0.842 ± 0.076 | 1.844 ± 1.368 | 0.925 ± 0.080 | 41.797 ± 130.019 |
| | Pathological | 0.817 ± 0.126 | 2.550 ± 8.005 | 0.912 ± 0.121 | 33.278 ± 93.104 |
| **paramahir_2023** | Neurotypical | 0.041 ± 0.066 | 75.381 ± 49.828 | 0.365 ± 0.293 | 1316.442 ± 1491.284 |
| | Pathological | 0.038 ± 0.060 | 85.358 ± 55.674 | 0.314 ± 0.282 | 1502.144 ± 1709.552 |
| **pasteurdbc** | Neurotypical | 0.834 ± 0.082 | 2.000 ± 1.522 | 0.918 ± 0.085 | 46.558 ± 146.162 |
| | Pathological | 0.802 ± 0.129 | 2.879 ± 3.591 | 0.901 ± 0.126 | 37.212 ± 98.278 |
| **qd_neuroincyte** | Neurotypical | 0.744 ± 0.178 | 9.321 ± 15.672 | 0.867 ± 0.156 | 31.234 ± 71.161 |
| | Pathological | 0.626 ± 0.242 | 11.399 ± 16.616 | 0.792 ± 0.233 | 36.915 ± 78.492 |
| **unipd-sum-aug** | Neurotypical | 0.827 ± 0.083 | 2.041 ± 1.614 | 0.917 ± 0.087 | 49.062 ± 132.939 |
| | Pathological | 0.797 ± 0.131 | 2.582 ± 2.797 | 0.903 ± 0.125 | 44.620 ± 111.600 |
| **upfetal24** | Neurotypical | 0.832 ± 0.080 | 2.096 ± 1.579 | 0.918 ± 0.087 | 45.657 ± 131.983 |
| | Pathological | 0.811 ± 0.124 | 2.682 ± 4.136 | 0.909 ± 0.120 | 35.099 ± 88.130 |
| **vicorob** | Neurotypical | 0.841 ± 0.077 | 1.870 ± 1.600 | 0.929 ± 0.076 | 46.210 ± 138.108 |
| | Pathological | 0.811 ± 0.122 | 2.457 ± 2.685 | 0.911 ± 0.117 | 37.085 ± 97.032 |

# Appendix A6. FeTA 2024 biometry results by site, label and pathology

**Biometry Estimation**  In this section of the supplementary materials, we present detailed results for the second task of the FeTA 2024 challenge, which focuses on fetal brain biometry estimation. To complement the primary evaluation metrics, we also report the **Mean Absolute Error (MAE)**, which offers an intuitive interpretation of the prediction errors in physical units (millimeters).

The MAE is defined as:

$$\text{MAE} = \frac{1}{N} \sum_{i=1}^{N} |\hat{y}_i - y_i|$$

where $N$ is the number of samples, $\hat{y}_i$ is the predicted measurement for the $i$-th subject, and $y_i$ is the corresponding ground truth measurement.

All MAE values are reported in millimeters (mm), which facilitates a direct understanding of the magnitude of the errors made by the algorithms in the context of fetal brain structure dimensions.

Table A6.1. Biometry results for all teams participating in the FeTA 2024 biometry challenge together with the baseline model and inter-rater variability stratified by label. The values for each team are sorted by MAPE in the increasing order.

| Team | Pathology | MAE (mm) | MAPE |
|---|---|---|---|
| **GA** | Pathological | 3.468±4.504 | 0.090±0.120 |
| | Neurotypical | 2.948±3.057 | 0.103±0.137 |
| **inter-rater** | Neurotypical | 1.333±1.781 | 0.041±0.057 |
| | Pathological | 1.625±2.040 | 0.065±0.105 |
| **cesne-digair** | Neurotypical | 3.283±6.208 | 0.092±0.126 |
| | Pathological | 2.896±4.831 | 0.099±0.139 |
| **falcons** | Neurotypical | 10.262±14.883 | 0.262±0.281 |
| | Pathological | 13.525±14.425 | 0.411±0.421 |
| **feta_sigma** | Neurotypical | 2.447±2.567 | 0.069±0.074 |
| | Pathological | 3.499±4.260 | 0.123±0.168 |
| **jwcrad** | Neurotypical | 1.948±1.779 | 0.057±0.057 |
| | Pathological | 3.123±6.939 | 0.095±0.143 |
| **paramahir_2023** | Pathological | 9.564±11.359 | 0.257±0.235 |
| | Neurotypical | 13.438±15.592 | 0.306±0.262 |
| **pasteurdbc** | Neurotypical | 3.704±3.551 | 0.139±0.177 |
| | Pathological | 3.852±3.698 | 0.176±0.254 |
| **qd_neuroincyte** | Pathological | 14.208±19.129 | 0.384±0.452 |
| | Neurotypical | 19.497±26.977 | 0.419±0.433 |

Table A6.2. Biometry results for all teams participating in the FeTA 2024 biometry challenge together with the baseline model and inter-rater variability stratified by label

| Team | Team | MAE (mm) | MAPE |
|---|---|---|---|
| [GA] | HV | 1.409±1.388 | 0.113±0.131 |
| | LCC | 3.406±3.302 | 0.127±0.161 |
| | TCD | 2.787±3.415 | 0.108±0.167 |
| | bBIP | 4.156±4.864 | 0.068±0.074 |
| | sBIP | 4.363±4.691 | 0.065±0.064 |
| [inter-rater] | HV | 0.993±0.885 | 0.080±0.088 |
| | LCC | 2.542±3.541 | 0.096±0.142 |
| | TCD | 1.189±1.233 | 0.049±0.072 |
| | bBIP | 1.852±1.388 | 0.033±0.028 |
| | sBIP | 0.951±0.950 | 0.015±0.016 |
| cesne-digair | HV | 1.197±1.199 | 0.098±0.120 |
| | LCC | 5.704±3.329 | 0.177±0.102 |
| | TCD | 3.238±3.914 | 0.123±0.162 |
| | bBIP | 2.381±3.250 | 0.040±0.055 |
| | sBIP | 3.079±10.121 | 0.047±0.147 |
| falcons | HV | 5.747±4.488 | 0.463±0.507 |
| | LCC | 9.804±8.722 | 0.349±0.349 |
| | TCD | 10.294±10.779 | 0.367±0.342 |
| | bBIP | 14.906±17.945 | 0.246±0.274 |
| | sBIP | 18.980±20.905 | 0.281±0.293 |
| feta_sigma | HV | 1.429±1.273 | 0.116±0.121 |
| | LCC | 3.540±3.073 | 0.126±0.145 |
| | TCD | 3.213±3.489 | 0.137±0.203 |
| | bBIP | 3.369±3.453 | 0.057±0.060 |
| | sBIP | 3.507±5.144 | 0.055±0.075 |
| jwcrad | HV | 1.377±1.087 | 0.103±0.083 |
| | LCC | 3.241±3.047 | 0.112±0.117 |
| | TCD | 1.983±1.735 | 0.072±0.072 |
| | bBIP | 3.285±7.850 | 0.054±0.131 |
| | sBIP | 3.013±7.579 | 0.048±0.132 |
| paramahir_2023 | HV | 4.148±3.911 | 0.294±0.243 |
| | LCC | 8.936±8.877 | 0.285±0.244 |
| | TCD | 9.459±9.225 | 0.308±0.270 |
| | bBIP | 16.397±16.698 | 0.261±0.242 |
| | sBIP | 17.728±18.415 | 0.255±0.244 |
| pasteurdbc | HV | 5.694±2.572 | 0.435±0.262 |
| | LCC | 5.833±5.816 | 0.205±0.230 |
| | TCD | 1.261±1.381 | 0.054±0.099 |
| | bBIP | 3.820±2.663 | 0.065±0.046 |
| | sBIP | 2.470±1.637 | 0.037±0.027 |
| qd_neuroincyte | HV | 5.762±5.638 | 0.428±0.528 |
| | LCC | 9.946±10.714 | 0.328±0.379 |
| | TCD | 15.352±15.694 | 0.479±0.413 |
| | bBIP | 24.821±29.450 | 0.384±0.432 |
| | sBIP | 26.958±32.977 | 0.378±0.436 |

Table A6.3. Biometry results for all teams participating in the FeTA 2024 biometry challenge together with the baseline model and inter-rater variability stratifies by site

| Team | Site | MAE (mm) | MAPE |
|---|---|---|---|
| [GA] | KCL | 2.365±2.297 | 0.055±0.045 |
| | CHUV | 3.027±2.790 | 0.083±0.121 |
| | UCSF | 3.280±4.375 | 0.094±0.121 |
| | VIEN | 3.788±5.361 | 0.106±0.112 |
| | KISPI | 3.262±3.219 | 0.121±0.172 |
| [inter-rater] | KCL | 0.747±0.645 | 0.020±0.024 |
| | CHUV | 1.019±0.924 | 0.029±0.029 |
| | UCSF | 1.421±1.513 | 0.050±0.059 |
| | KISPI | 1.753±1.814 | 0.069±0.095 |
| | VIEN | 2.168±3.070 | 0.086±0.135 |
| cesne-digair | UCSF | 2.487±3.767 | 0.079±0.094 |
| | KCL | 3.109±3.544 | 0.085±0.116 |
| | KISPI | 2.610±4.410 | 0.096±0.147 |
| | VIEN | 3.665±9.285 | 0.106±0.157 |
| | CHUV | 3.551±3.629 | 0.109±0.133 |
| falcons | KCL | 4.885±6.207 | 0.109±0.108 |
| | CHUV | 11.379±22.126 | 0.262±0.427 |
| | KISPI | 7.156±5.066 | 0.282±0.311 |
| | VIEN | 14.906±13.615 | 0.435±0.457 |
| | UCSF | 18.260±12.581 | 0.508±0.230 |
| feta_sigma | KCL | 1.779±1.603 | 0.046±0.047 |
| | CHUV | 2.164±2.126 | 0.069±0.116 |
| | KISPI | 2.567±2.734 | 0.111±0.189 |
| | VIEN | 3.540±3.598 | 0.117±0.122 |
| | UCSF | 4.402±5.357 | 0.121±0.119 |
| jwcrad | KCL | 1.305±1.025 | 0.035±0.035 |
| | CHUV | 2.392±1.896 | 0.071±0.079 |
| | UCSF | 2.499±5.267 | 0.074±0.096 |
| | KISPI | 2.047±1.704 | 0.078±0.085 |
| | VIEN | 4.051±9.407 | 0.108±0.182 |
| paramahir_2023 | KCL | 4.153±4.023 | 0.107±0.105 |
| | KISPI | 4.483±3.553 | 0.161±0.184 |
| | VIEN | 7.354±6.228 | 0.177±0.102 |
| | UCSF | 6.523±5.287 | 0.185±0.137 |
| | CHUV | 30.396±16.278 | 0.677±0.044 |
| pasteurdbc | KCL | 3.515±2.607 | 0.119±0.139 |
| | CHUV | 3.129±2.686 | 0.127±0.175 |
| | UCSF | 3.663±3.287 | 0.167±0.224 |
| | KISPI | 3.749±2.977 | 0.179±0.283 |
| | VIEN | 4.774±5.341 | 0.182±0.221 |
| qd_neuroincyte | KCL | 5.122±8.736 | 0.094±0.132 |
| | KISPI | 3.080±3.803 | 0.122±0.208 |
| | VIEN | 7.525±10.180 | 0.234±0.461 |
| | UCSF | 15.855±13.145 | 0.429±0.262 |
| | CHUV | 45.380±30.413 | 0.958±0.296 |

# Appendix A7. Correlation of quality and challenge metrics

**Correlation plots of visual quality scores and metrics across different sites and super-resolution methods for the best teams in FeTA 2024** Each dot represents an average metric value( of the top 3 teams in FeTA2024 `cesne-digair`, `mic-dkfz-feta24`, `vicorob`) for a given subject, with blue indicating quality scores (left axis) and red indicating a given metric (right axis). Sites and methods are grouped on the x-axis. Dashed lines connect data points for individual subjects across metrics. Pearson correlation coefficients (r) between quality and Dice are shown above each group

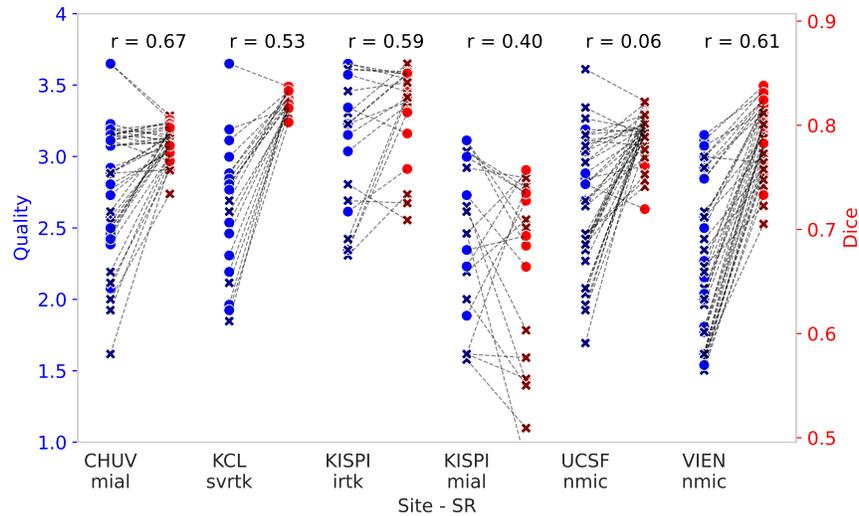

Figure A7.1: Correlation between visual quality scores and Dice across different sites and super-resolution methods.

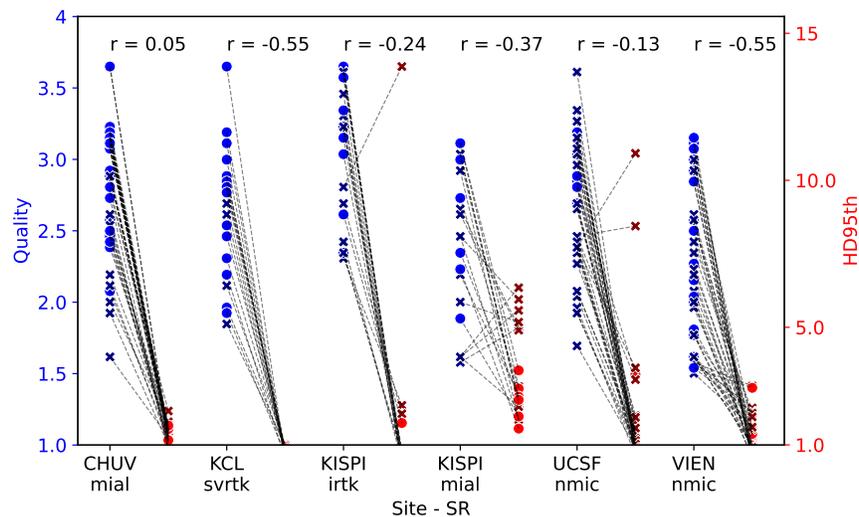

Figure A7.2: Correlation between visual quality scores and HD95 across different sites and super-resolution methods.

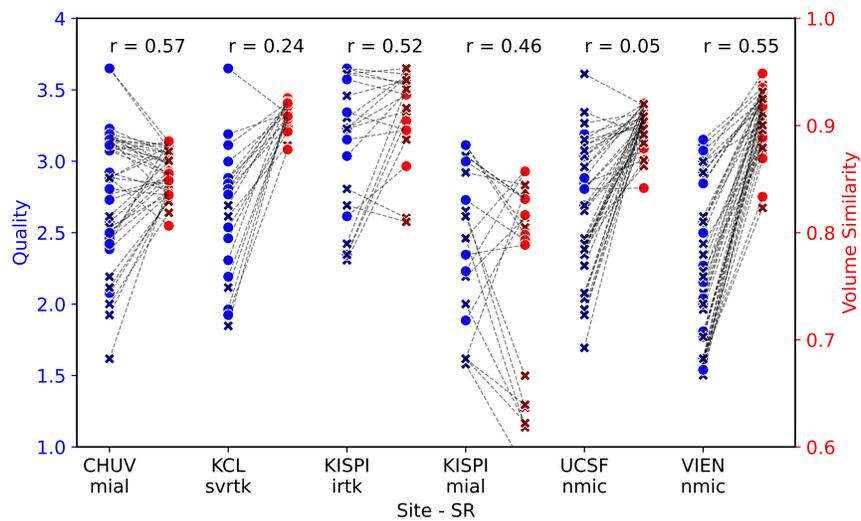

Figure A7.3: Correlation between visual quality scores and VS across different sites and super-resolution methods.

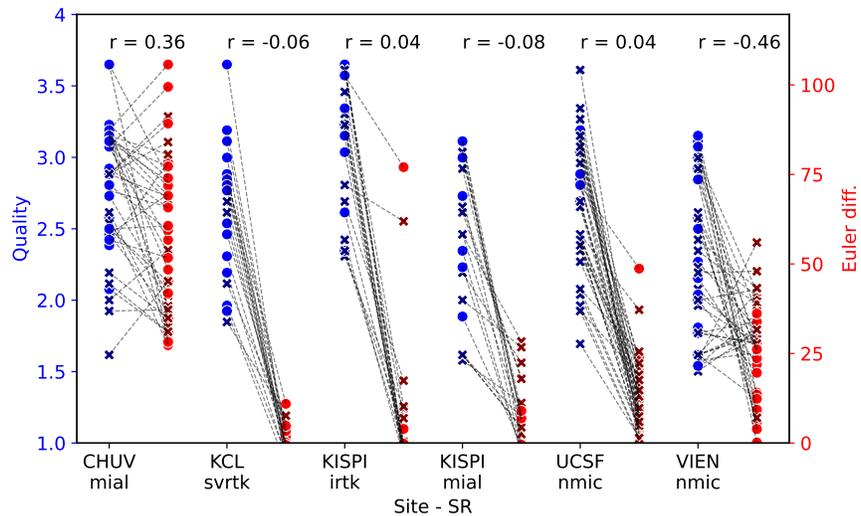

Figure A7.4: Correlation between visual quality scores and ED across different sites and super-resolution methods.

## Appendix A8. Exploring normalized Dice coefficient

**Correlation between Dice, Normalized Dice, Volume, and GA**  In Figure A8.1 (a), we present the distribution of Pearson correlation coefficients between each team's Dice score and the volume of the segmented label, averaged across all subjects for the 15 participating teams in the FeTA 2024 challenge. We observe that using the normalized Dice coefficient mitigates the known bias of the standard Dice metric toward larger volumes. However, normalized Dice does not eliminate the correlation with gestational age (GA). This residual association may be due to GA acting as a complex confounding factor, potentially interacting with other variables such as acquisition site and pathology, rather than directly influencing the metric.

**Impact of Using Normalized Dice on Team Rankings**  In Figure A8.1 (b), we assess the stability of team rankings when using normalized Dice instead of the standard Dice score. Although normalized Dice reduces the volume-related bias, the correlation structure across teams appears relatively stable. Rankings remain largely consistent, with only two pairs of teams swapping positions. Notably, the top six teams maintain their original ranking when switching from Dice to normalized Dice, suggesting that the relative performance differences are robust to this normalization.

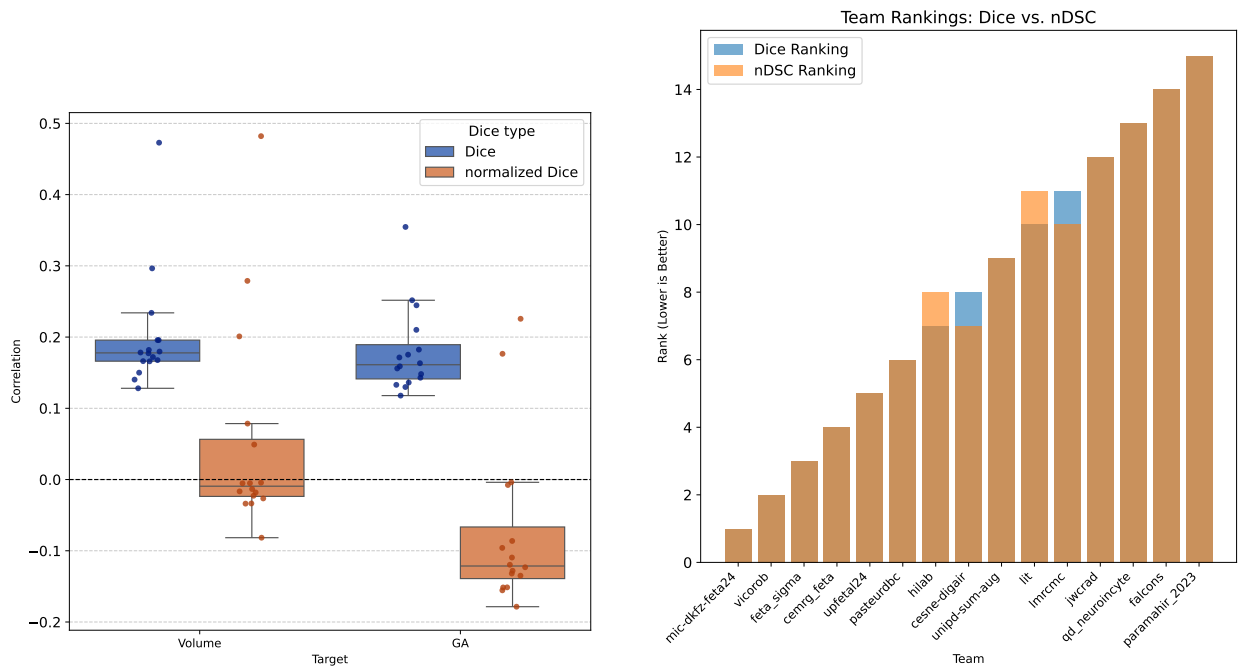

(a) Correlation coefficients between Dice/normalized Dice with volume and GA.

(b) Team-ranking stability using Dice vs. normalized Dice.

Figure A8.1: (a) Per-team Dice vs. volume/GA correlations. (b) Impact of normalization on team rankings.

# Appendix A9. Qualitative examples of predictions

**Qualitative Results**  This section of the supplementary materials presents qualitative comparisons of the predictions made by the top four teams in the FeTA 2024 challenge.

Figure A9.1 shows representative segmentation outputs for five subjects from the testing set, while Figure A9.2 highlights the corresponding segmentation errors. Overall, the visual differences between the top-performing models are minimal, indicating that all four algorithms deliver highly consistent and accurate segmentations in well-defined cases.

Figures A9.3 and A9.4 illustrate cases with the lowest Dice scores across the evaluated models. These difficult cases are typically associated with poor image quality (e.g., Subjects 2 and 3), or signs of abnormal fetal brain development (e.g., Subjects 0, 1, and 4). Such challenges underscore the limitations of current methods when dealing with low-quality inputs or atypical anatomy.

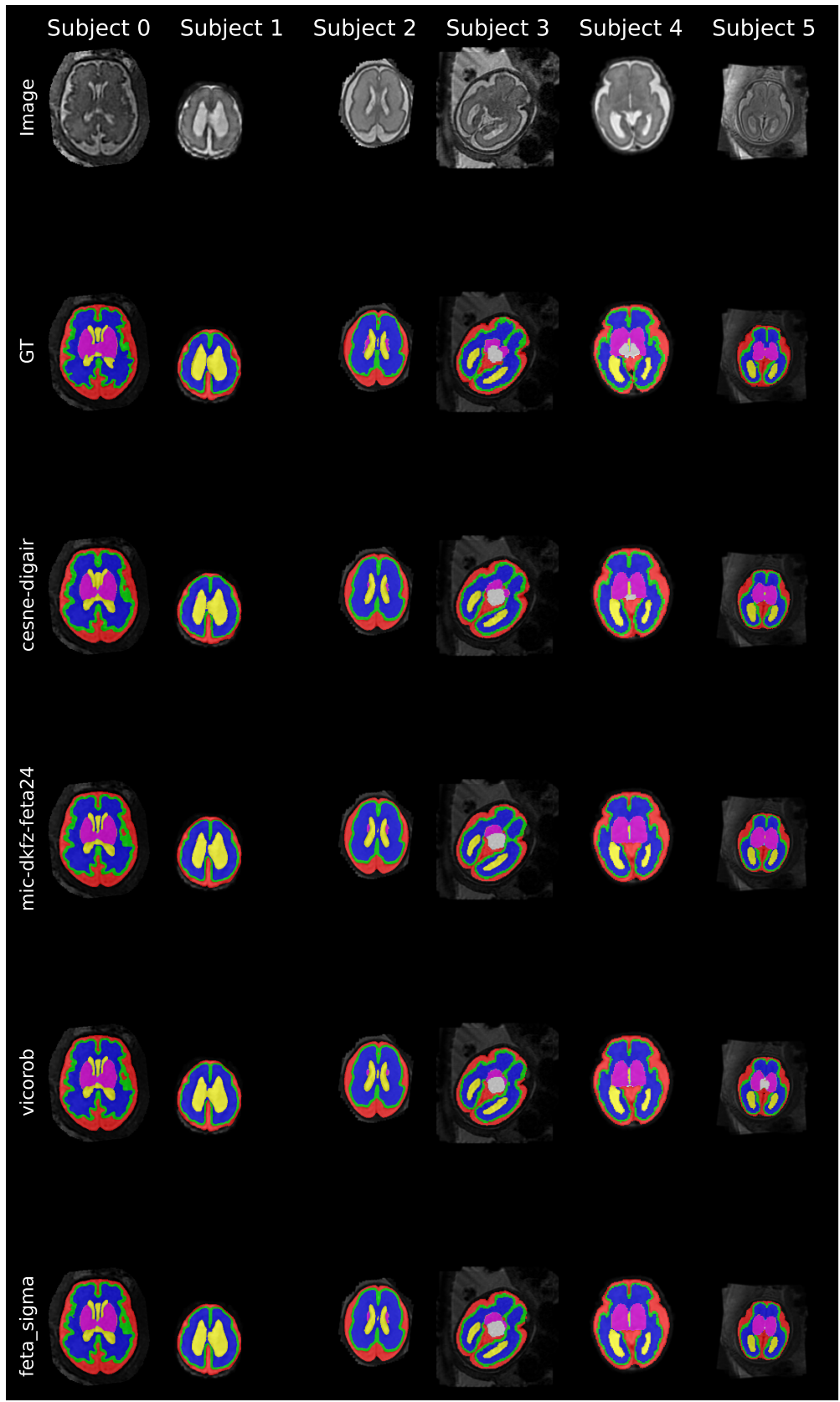

Figure A9.1: Segmentation results for five testing subjects produced by the top four teams in the FeTA 2024 challenge.

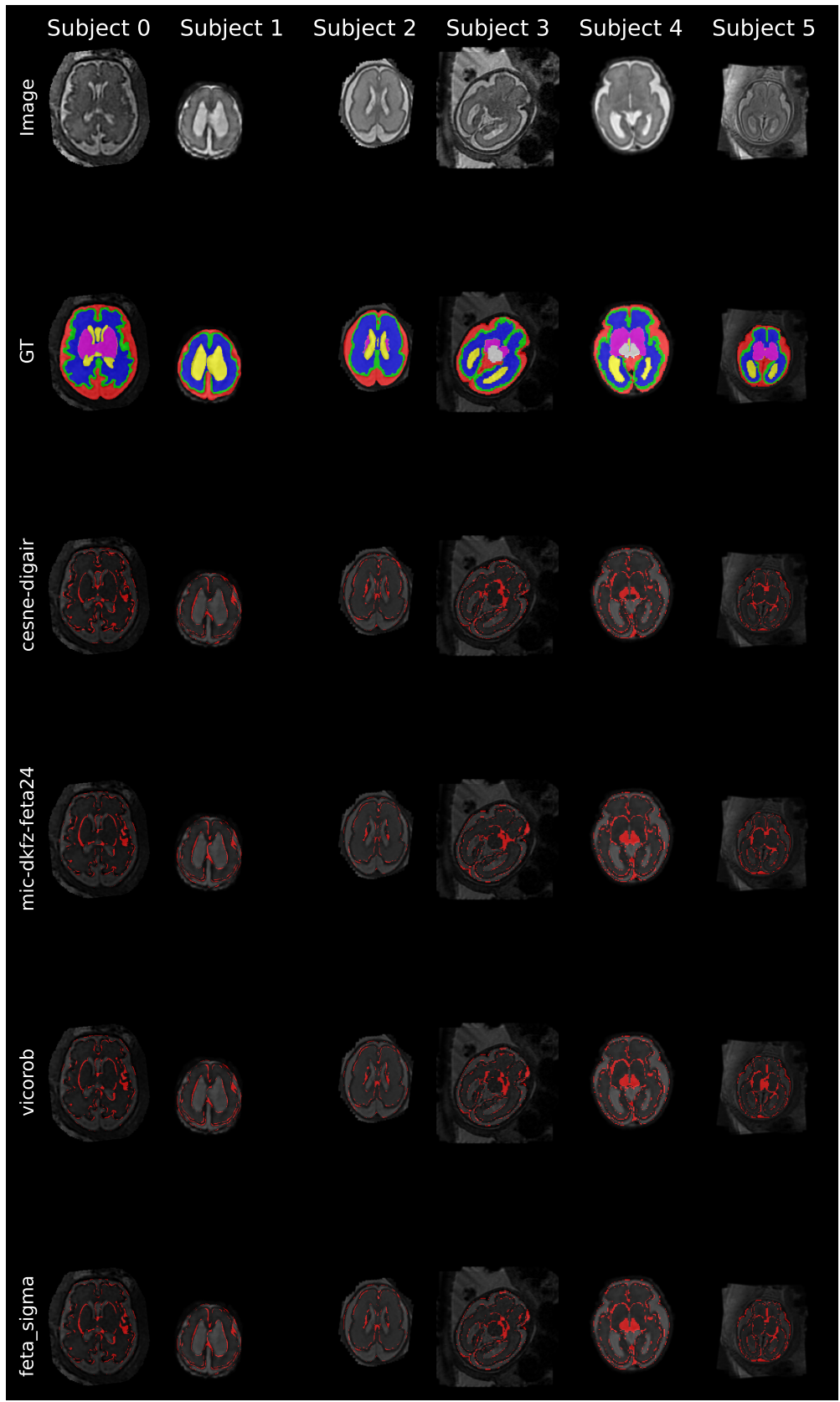

Figure A9.2: Segmentation errors for five testing subjects produced by the top four teams in the FeTA 2024 challenge. Red regions indicate voxels where predicted labels differ from the ground truth.

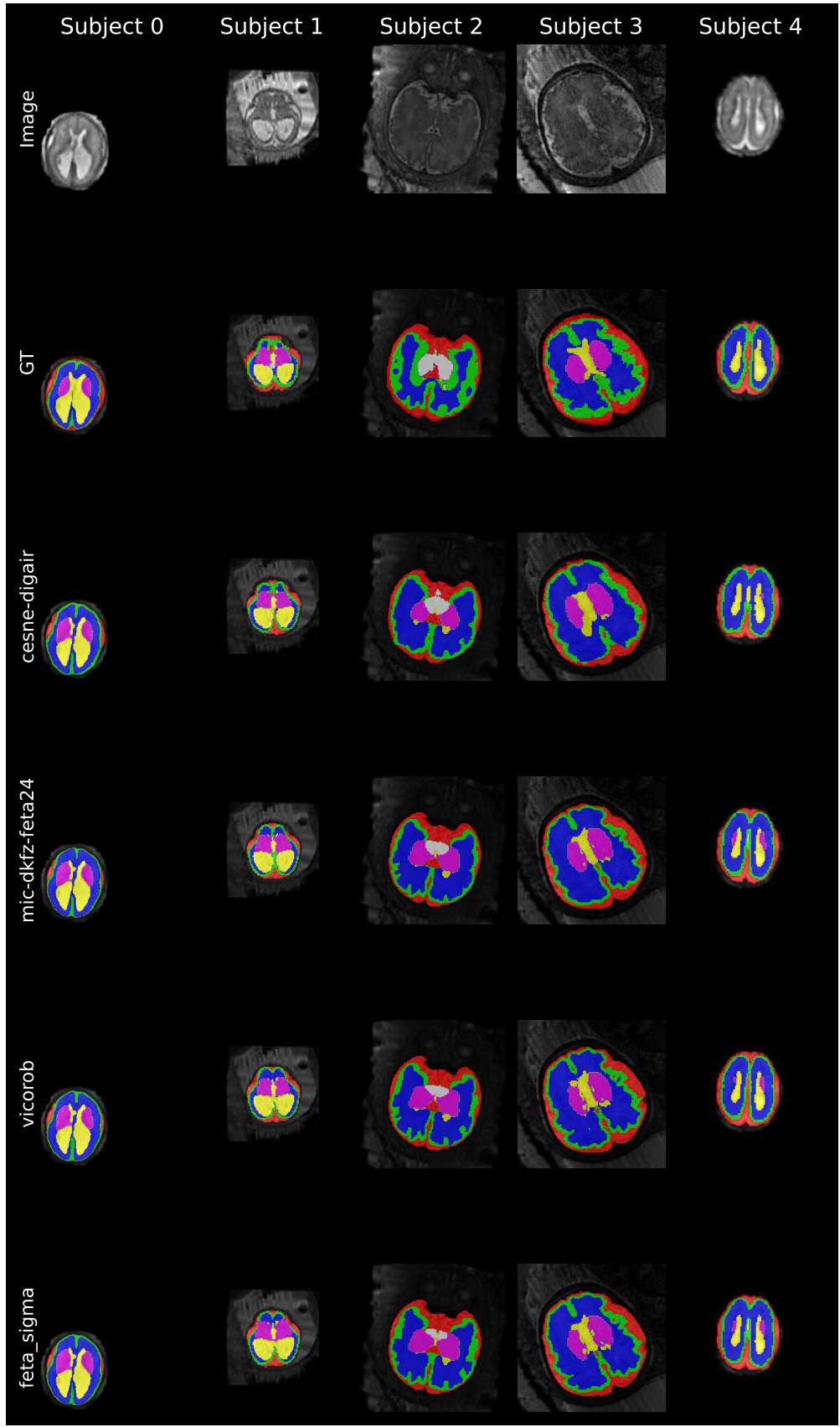

Figure A9.3: Segmentation results for five challenging testing subjects with the lowest Dice scores, produced by the top four teams in the FeTA 2024 challenge.

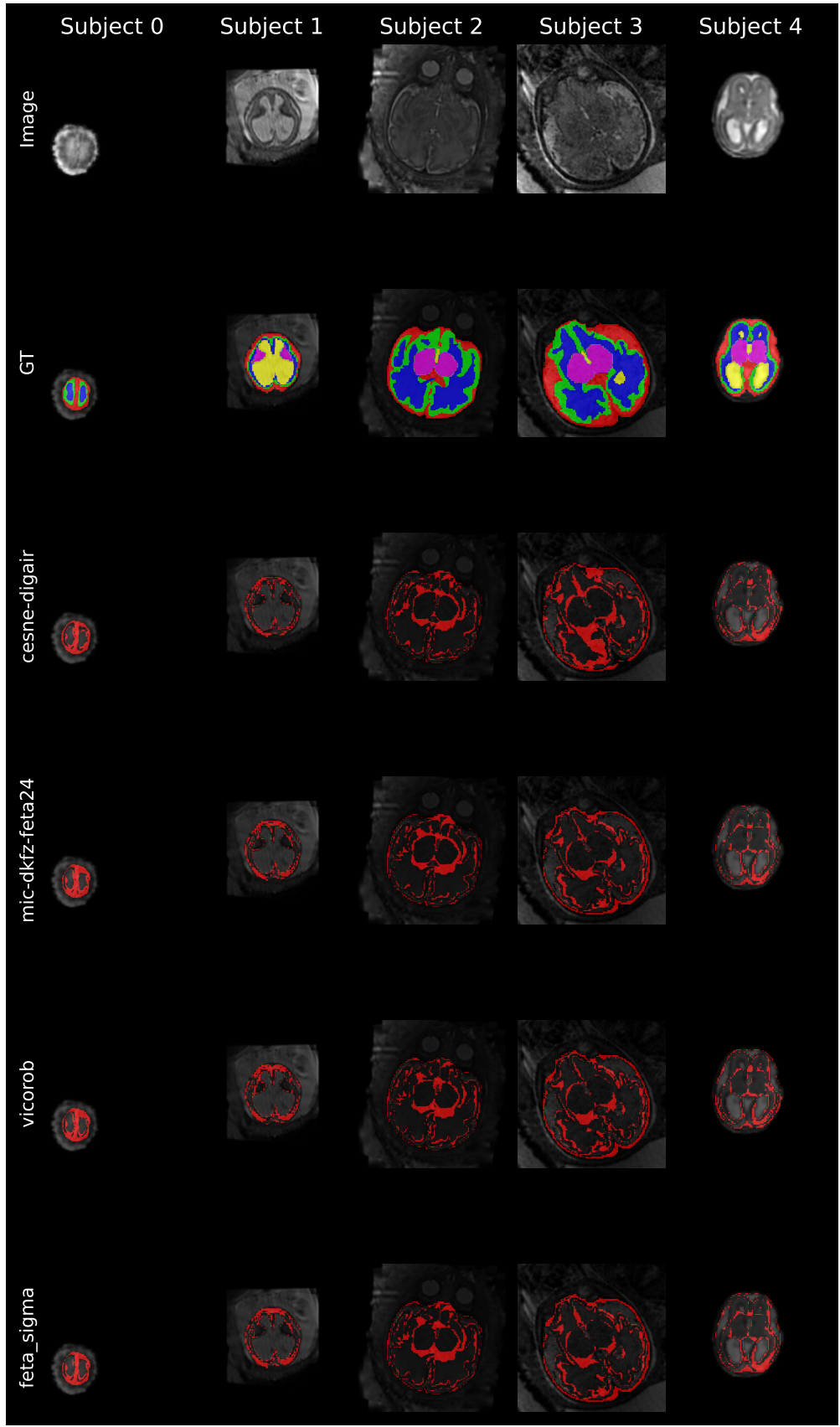

Figure A9.4: Segmentation errors for the five challenging testing subjects shown in Figure A9.3. Red voxels indicate mismatches between the predicted labels and ground truth.



\end{document}